\documentclass[lettersize,journal]{IEEEtran}
\usepackage{amsmath,amsfonts}
\usepackage{algorithmic}
\usepackage{algorithm}
\usepackage{array}
\usepackage[caption=false,font=normalsize,labelfont=sf,textfont=sf]{subfig}
\usepackage{textcomp}
\usepackage{stfloats}
\usepackage{url}
\usepackage{verbatim}
\usepackage{graphicx}
\usepackage{epstopdf}
\usepackage{cite}
\usepackage{bm} 
\begin{document}

\title{Hybrid Causal Identification and Causal Mechanism Clustering}

\author{Saixiong Liu, Yuhua Qian,~\IEEEmembership{Member,~IEEE,} Jue Li, \\Honghong Cheng,~\IEEEmembership{Member,~IEEE,}and Feijiang Li
        % <-this % stops a space
\thanks{This work was supported by the Science and Technology Innovation 2030-“New Generation of Artificial Intelligence” Major Program under Grant (No.2021ZD0112400), the National Natural Science Foundation of China (Nos. 62136005), the Science and Technology Major Project of Shanxi (No.202201020101006), the Key R\&D Program of Shanxi Province, China (Grant no. 202202020101004).}% <-this % stops a space
\thanks{Saixiong Liu, Yuhua Qian, Jue Li and Feijiang Li are with the Key with Institute of Big Data Science and Industry, Shanxi University, Taiyuan, Shanxi 030006, China (E-mail: \{liu\_saixiong, jinchengqyh\}@126.com; lijue688@163.com; fjli@sxu.edu.cn).}
\thanks{Honghong Cheng is with the School of Information, Shanxi University of Finance and Economics, Taiyuan, Shanxi 030012, China, and also with Institute of Big Data Science and Industry, Shanxi University, Taiyuan, Shanxi 030006, China. (E-mail: chhsxdx@163.com).}
\thanks{(Corresponding author: Yuhua Qian.)}
\thanks{Manuscript received xx, 2023; revised xx, xx, 202X.}
}

\markboth{Journal of \LaTeX\ Class Files,~Vol.~14, No.~8, August~2021}%
{Shell \MakeLowercase{\textit{et al.}}: A Sample Article Using IEEEtran.cls for IEEE Journals}

\IEEEpubid{xx/xx.xx~\copyright~2023 IEEE}

\maketitle

\begin{abstract}
Bivariate causal direction identification is a fundamental and vital problem in the causal inference field. Among binary causal methods, most methods based on additive noise only use one single causal mechanism to construct a causal model. In the real world, observations are always collected in different environments with heterogeneous causal relationships. Therefore, on observation data, this paper proposes a Mixture Conditional Variational Causal Inference model (MCVCI) to infer heterogeneous causality. Specifically, according to the identifiability of the Hybrid Additive Noise Model (HANM), MCVCI combines the superior fitting capabilities of the Gaussian mixture model and the neural network and elegantly uses the likelihoods obtained from the probabilistic bounds of the mixture conditional variational auto-encoder as causal decision criteria. Moreover, we model the casual heterogeneity into cluster numbers and propose the Mixture Conditional Variational Causal Clustering (MCVCC) method, which can reveal causal mechanism expression. Compared with state-of-the-art methods, the comprehensive best performance demonstrates the effectiveness of the methods proposed in this paper on several simulated and real data.
\end{abstract}


\section{Introduction}
\IEEEPARstart{T}{he} causal method can help AI fields, such as machine learning and deep learning, with an essential modeling method to establish a more stable and generalized model. Scholars have made lots of efforts in this area, and many use causal methods to construct a robust and interpretable artificial intelligence world \cite{ref2,ref34} like solving the domain generalization problem through causality\cite{ref1}. The key to these studies is to identify whether the treatment variable is the cause of the target variable, that is, to infer the causal relationship between the bivariate.\\

Many binary causal methods have appeared, and the function-based causal method mainly follows the assumptions as characteristics of the noise are non-Gaussian, the causal variables and the noise are independent of each other, the distribution of the causal variables and the gradient of the causal function are independent, to infer the direction of causality between variables. Similar to Additive Noise Model (ANM) \cite{ref6}, most of the functional models built on the Structural Causal Model (SCM), such as the linear non-Gaussian acyclic model (LiNGAM)\cite{ref32}, post-nonlinear model (PNL) \cite{ref5}, Information-Geometric Causal Inference (IGCI) \cite{ref7} method, and the Regression Error based Causal Inference (RECI) \cite{ref8} method, all use the asymmetry to distinguish causality. In addition, in recent years, algorithms as the adversarial network against orthogonal regression based on the principle of additive noise \cite{ref9}, the non-parametric method quantile copula causal discovery (QCCD) \cite{ref10}, Heteroscedastic Noise Models for causal inference (HEC) \cite{ref11}, and ANM Mixture Model (ANM-MM) \cite{ref12} applying GPPOM measurement, can be summarized as the model for causal inference due to the characteristics of the data.

Most of the above methods are for single data or one causal model, and only a few works\cite{ref3,ref11,ref12} are for mixed causal models and heterogeneous noise. Causal research under mixed data still faces significant challenges. Deep neural networks are proven to have strong fitting capabilities over the last decade. From the perspective of the causal noise model under the generation mechanism of multi-source data, this paper combines the variational inference ability in the conditional variational encoder (CVAE)  \cite{ref15}, and a mixture variational causal rule is established to realize the causal identification.

In terms of causal clustering, in recent years, due to ushering in the big data revolution, researchers tend to pay more attention to correlation relationships rather than causality, especially in clustering methods. We found that most causal feature extraction methods use the Markov blankets  \cite{ref14} to extract features, which often appear in high-dimensional data methods and are used to reduce the dimension. This paper aims to identify causality by modeling different causal mechanisms according to their essence. Furthermore, data clustering is realized through the proposed causal mechanism to divide the data. This paper considers the problem of hybrid modeling of binary causal inference under observational data. We propose our hybrid causal model and construct a clustering model for\linebreak \\ 

\noindent hybrid causal mechanisms. Our contributions consist of the following points:

1. We refined the hybrid additive noise model and supplied its identifiability proof. A mixture conditional variational auto-encoder is constructed, which is a general regressor, and a hybrid causal recognition rule based on ELBO bounds is established by addressing the probability model of the mixture conditional variational auto-encoder.

2. A new clustering solution is proposed, which can use causality for high-level representation of features. Here we tightly couple causal heterogeneity with causal mechanism clustering.

3. The comprehensive best performance of our algorithms has been obtained on both the simulated and real datasets.

\section{RELATED WORK}
In the binary causal methods part, we only discuss the function-based binary causal methods in the structure causal methods \cite{ref4,ref13}. On observed data, Shimizu et al. proposed the Lingam algorithm\cite{ref32}, which can identify acyclic causal graphs of non-Gaussian exogenous noise. Hoyer et al. \cite{ref6} suggested one non-liner model ANM. Following the ANM method, Zhang Kun introduced the PNL method\cite{ref5}, expressed as $Y=f_2{(f}_1\left(X\right)+\varepsilon)$. Subsequently, the additive noise model based on discrete data was provided by Peters \cite{ref16}, while the above methods mainly deal with continuous data. The classic binary causality identification IGCI method \cite{ref7} uses information entropy and information geometry theory to infer causality by measuring the information relationship between two variables. Besides the IGCI method, the information theory casual method as Slope\cite{ref17} uses the MDL length in accordance with on Kolmogorov complexity to distinguish causality. At the same time, Sloppy \cite{ref18} uses the regression error by Kolmogorov’s rules for causal distinguish.

With the advent of big data, binary causal methods based on neural networks have also appeared in the field of vision. For example, through the regression error, the RECI \cite{ref8} method uses a relatively simple way to identify the causal relationship. The neural network shows good recognition accuracy among the four regression methods. In view of regression error methods, there are NNCL\cite{ref19}, QCCD \cite{ref10}, HEC\cite{ref11}, etc. The paper \cite{ref9} introduces the AdOR and Adose methods under the idea of GAN and trains the GAN network to identify cause and effect through the principle of small mutual information between the noise and cause. Although the CANM\cite{ref23} method is a method for hidden variables, it uses the log-likelihood rule and the variational reasoning ability of VAE\cite{ref33} to determine causality. The ANM-MM method is a recent approach to infer causality using mixed data in continuous variables in binary causality. In summary, causal inference research on heterogeneous data still needs to be improved. This paper mainly conducts causal modeling for mixed data and identifies the causality of ultimately observed data.

Most existing clustering methods use similarity measurement and other principles, then optimize the distance between all points and the cluster center to obtain the cluster division, like k-means\cite{ref21,ref27} and spectral clustering method\cite{ref28}, which ignore the data generation mechanism. The existing causal clustering methods generally use Markov blankets to extract causal structural features, mainly for multi-variable data, and perform clustering. Few clustering studies on the causal mechanism, and only one research ANM-MM reflected it. Therefore, this paper proposes a causal mechanism clustering method based on our mixture causal model and data generation mechanism.

\section{METHODOLOGY}
\subsection{Additive noise model}
Hoyer et al. \cite{ref6} proposed the additive noise model, which derives the casual asymmetric essence that only holds in the causal direction. ANM expresses the effect as a function of the cause with independent additive noise as in formula (1).
\begin{equation} \label{(1)}
    Y=f(X)+\varepsilon
\end{equation}
where $\varepsilon $ is noise, it is shown that under most circumstances, there is an ANM model in the forward direction. Still, there is no model that conforms to the ANM form in the reverse direction as formula (2), and the causal direction can be regarded as  $X\rightarrow Y$. That is, $X$ is the cause, and $Y$ is the effect.
\begin{equation} \label{(2)}
    X=g\left(Y\right)+\hat{\varepsilon}
\end{equation}
\subsection{Hybrid Additive noise model definition}
In the conditional variable causality inference model MCVCI, we first start the construction of our hybrid model. Assume that the observation data consists of finite components, and each component is an ANM. We set the mixed number as $K$, sample number as $m$, $X={{\{x_{i,k}\}}}_{i=1,k=1}^{m,K}$, and $Y={\{y_{i,k}\}}_{i=1,k=1}^{m,K}$. We have the following HANM,
\begin{equation} \label{(3)}
 Y= \sum_{k=1}^{K}{w_k(f_k\left(x_k\right)+\epsilon_k)}
\end{equation}
where $w_k$ is the weight of the $kth$ component, $\sum_{k=1}^{K}w_k=1$, $\epsilon$ is the additive noise term, and $x_k\perp \!\!\! \perp{\epsilon}_k$.

The problem of label shifts in the domain generalization field and Gaussian mixture models inspires the definition of our approach. The change in $Y$ is due to an offset in $X$, in function $f$ or in noise. Furthermore, unlike the ANM-MM method which considers the first factor, we consider the latter two. Here, we directly use the mixture function and noise to represent the offset part, which is similar to the modeling idea of the Gaussian mixture model.

\subsection{The identifiability of the HANM method}
Similar to the ANM-MM approach, we present an identifiability-proof version of our constructed hybrid model.For formula (1), assume $\varepsilon$ and $X$  have the strict positive density $p_\varepsilon$ and $p_X$, $p_\varepsilon$, $p_X$ and $f$ are strictly third-order differentiable. The joint distribution $p\left(X,Y\right)$ of ANM has formula (4).
\begin{equation}
\left\{
\begin{aligned}
    & p\left(X,Y\right)=p_\varepsilon\left(Y-f\left(X\right)\right)p_X\left(X\right) \\
    & p\left(X,Y\right)=p_{\widetilde{\varepsilon}}\left(X-g\left(Y\right)\right)p_Y\left(Y\right)
\end{aligned}
\right.
\label{(4)}
\end{equation}

Therefore, combining formula (3) and formula (4), the same joint distribution of HANM as formula (5).
\begin{equation}
\left\{
\begin{aligned}
    & p\left(X,Y\right)=p\left(Y\right)\sum_{k=1}^{K}{w_kp_{{\widetilde{\epsilon}}_k}\left(x_k-g_k\left(y_k\right)|y_k\right)} \\
    & p\left(X,Y\right)=p\left(X\right)\sum_{k=1}^{K}{w_kp_{\epsilon_k}\left(y_k-f_k\left(x_k\right)|x_k\right)}
\end{aligned}
\right.
\label{(5)}
\end{equation}

\textbf{Lemma 1} When $X\rightarrow Y$ and conform to a HANM, there is a HANM in the anti-causal direction, i.e.
\begin{equation} \label{(6)}
   X=\sum_{k=1}^{K}{w_k\left(g_k\left(y_k\right)+{\widetilde{\epsilon}}_k\right)}
\end{equation}

The casual distribution$\ p(X)$, noise distribution $p_\epsilon$, and nonlinear function $f$, parameters distribution should satisfy the following ordinary differential equation (7).
\begin{equation} \label{(7)}
   \xi^{\prime\prime\prime}=\frac{G\left(X,Y\right)}{M\left(X,Y\right)}\xi^{\prime\prime}+\frac{(U\left(X,Y\right)G(X,Y)}{M\left(X,Y\right)}-H\left(X,Y\right)
\end{equation}
where $\xi:=logp\left(X\right)$, the details of$\ G(X,Y)$,$\ M\left(X,Y\right)$, $U\left(X,Y\right)$, and $H\left(X,Y\right)$ are as follows.

\noindent \textbf{\emph{Proof of Lemma 1.}}

Suppose there is a hybrid additive noise model in the inverse direction, we assume that the mixing number is $K$, sample number is $m$, $X={\{x_{i,k}\}}_{i=1,k=1}^{m,K}$, and $Y={\{y_{i,k}\}}_{k=1}^{m,K}$, we have
$$X=w_k(g_k\left(Y\right)+{\widetilde{\epsilon}}_k).$$

The joint probability for backward modeling is
\begin{align*}
    p\left(X,Y\right)&= p\left(Y\right)p\left(X|Y\right) \\
    &= p\left(Y\right)\sum_{k=1}^{K}{w_kp_{{\widetilde{\epsilon}}_k}\left(x_k-g_k\left(y_k\right)|y_k\right)}
\end{align*}
where $\sum_{k=1}^{K}w_k=1$. When $y\perp \!\!\! \perp\widetilde{\epsilon}$,
$$p\left(X,Y\right)=\ p\left(Y\right)\sum_{k=1}^{K}{w_kp_{{\widetilde{\epsilon}}_k}\left(x_k-g_k\left(y_k\right)\right)}$$

We seek the likelihood function for the joint density of the forward model, then can get equation (8).
\begin{equation} \label{(8)}
  \pi\left(X,Y\right)=logp\left(X,Y\right)=\sum_{k=1}^{K}{w_kv_k\left(y_k-f_k\left(x_k\right)\right)}+\xi\left(X\right)
\end{equation}
where $\xi\left(X\right)=logp\left(X\right)=\sum_{k=1}^{K}{w_k\xi_k\left(x_k\right)}$. We simplify it as $\widetilde{v}\left(X-g\left(Y\right)\right)=\sum_{k=1}^{K}{w_klogp_{{\widetilde{\epsilon}}_k}}=\sum_{k=1}^{K}{w_k{\widetilde{v}}_k\left(x_k-g_k\left(y_k\right)\right)}$, $\eta(Y)=logp\left(Y\right)$.

If formula (8) holds, $\pi\left(X,Y\right)=\ \widetilde{v}\left(X-g\left(Y\right)\right)+\eta(Y)$.

As the ANM method derivation, we also use $\pi$ for the partial derivation of $X$.
\begin{equation} \label{(9)}
   \frac{\partial\pi}{\partial X}={\widetilde{v}}^\prime\left(X-g\left(Y\right)\right)
\end{equation}

Take partial derivatives of $Y$ using $\frac{\partial\pi}{\partial X}$ ,  we can get equation (10).
\begin{equation} \label{(10)}
   \frac{\partial^2\pi}{\partial X\partial Y}=\sum_{k=1}^{K}{w_k{\widetilde{v}_k}^{\prime\prime}\left(x_k-g_k\left(y_k\right)\right)g_k\prime\left(y_k\right)}={\widetilde{v}}^{\prime\prime}\left(X-g(Y)\right)
\end{equation}

And take partial derivatives of $X$ using $\frac{\partial\pi}{\partial X}$ , obtain the second order derivative of $\pi$ with respect to $X$.
\begin{equation} \label{(11)}
   \frac{\partial^2\pi}{\partial X^2}={\widetilde{v}}^{\prime\prime}\left(X-g\left(Y\right)\right)
\end{equation}

In the same way, we get
\begin{equation} \label{(12)}
   \frac{\partial}{\partial X}\left(\frac{\partial^2\pi/\partial X^2}{\partial^2\pi/\partial X\partial Y}\right)=0
\end{equation}

We then take these derivatives as in Equation (9)-(11) above for the probability $p\left(X,Y\right)$ of the forward model, and we let $M\left(X,Y\right)=\frac{\partial^2\pi}{\partial X\partial Y}$ , $N\left(X,Y\right)=\frac{\partial^2\pi}{\partial X^2}$.
\begin{equation} \label{(13)}
\frac{\partial\pi}{\partial X}=\xi^\prime\left(X\right)+v^\prime\left(Y-f\left(X\right)|X\right)
\end{equation}
\begin{equation} \label{(14)}
M\left(X,Y\right)=-\sum_{k=1}^{K}{{w_kv_k}^{\prime\prime}\left(y_k-f_k\left(x_k\right)\right){f_k}^\prime\left(x_k\right)}
\end{equation}
\begin{equation} 
\begin{aligned}
   N\left(x,y\right)&=\sum_{k=1}^{K}{{w_kv_k}^{\prime\prime}\left(y_k-f_k\left(x_k\right)\right)}\left({f_k}^\prime\left(x_k\right)\right)^2 \\
   &-\sum_{k=1}^{K}{w_k{v_k}^\prime\left(y_k-f_k\left(x_k\right)\right){f_k}^{\prime\prime}\left(x_k\right)}+\xi^{\prime\prime}\left(X\right)
\end{aligned}
\end{equation}

For convenience to express, we write $v^{\prime\prime}\left(Y-f\left(X\right)\right)$ as $v^{\prime\prime}$, $\sum_{k=1}^{K}{{w_kf_k}^\prime\left(x_k\right)}$ as $\sum_{k=1}^{K}{w_kf_k}^\prime$ and $\xi^{\prime\prime}(X)$ as $\xi^{\prime\prime}$. Let $U\left(X,Y\right)=\sum_{k=1}^{K}{{w_kv_k}^{\prime\prime}({{f_k}^\prime)}^2}-\sum_{k=1}^{K}{{w_kv_k}^\prime{f_k}^{\prime\prime}}$, thus it can be obtained that

\begin{equation}
\left\{
\begin{aligned}
    & N\left(X,Y\right)=U\left(X,Y\right)+\xi^{\prime\prime} \\
    & M\left(X,Y\right)=-\sum_{k=1}^{K}{{w_kv_k}^{\prime\prime}{f_k}^\prime}
\end{aligned}
\right.
\label{(16)}
\end{equation}

Like Equation (12), the same derivative can be obtained as
\begin{equation} \label{(17)}
  \frac{\partial}{\partial X}\left(\frac{N}{M}\right)=0
\end{equation}

Therefore,
\begin{equation} \label{(18)}
  \frac{\partial N}{\partial X}M-N\frac{\partial M}{\partial X}=0
\end{equation}

Simplifying Equation (18) yields
\begin{equation} \label{(19)}
   \left(H\left(X,Y\right)+\xi^{\prime\prime\prime}\right)M\left(X,Y\right)=\left(U\left(X,Y\right)+\xi^{\prime\prime}\right)G\left(X,Y\right)
\end{equation}

The formula (20) is the required.
\begin{equation} \label{(20)}
   \xi^{\prime\prime\prime}=\frac{G\left(X,Y\right)}{M\left(X,Y\right)}\xi^{\prime\prime}+\frac{(U\left(X,Y\right)G(X,Y)}{M\left(X,Y\right)}-H\left(X,Y\right)
\end{equation}
where $M(X,Y)\neq0$, and $H\left(X,Y\right)=\frac{\partial U}{\partial X}=\sum_{k=1}^{K}{w_k(-{v_k}^{\prime\prime\prime}({f_k}^\prime)^3+3{v_k}^{\prime\prime}{f_k}^\prime{f_k}^{\prime\prime}-{v_k}^\prime({f_k}^{\prime\prime\prime})}, G(X,Y) $$=\frac{\partial M}{\partial X}=\sum_{k=1}^{K}w_k{(v_k}^{\prime\prime\prime}{{{(f}_k}^\prime)}^2-{v_k}^{\prime\prime}{f_k}^{\prime\prime})$.

First, through the joint distribution of $p\left(X,Y\right)$, expand it into the form of formula (5). When $X\rightarrow Y$, if there is an inverse mixture ANM, $\frac{\partial}{\partial X}(\frac{\partial^2 \pi/\partial X^2}{\partial^2 \pi/\partial X\partial Y})=0$ in backward holds. Here $\pi=\sum_{k=1}^{K}{w_klogp_{{\widetilde{\epsilon}}_k}\left(x_k-g_k\left(y_k\right)\right)}+logp\left(Y\right)$. Because $p(X,Y)$ in the forward and reverse direction is equal, and still exist $\frac{\partial}{\partial X}(\frac{\partial^2 \pi/\partial X^2}{\partial^2 \pi/\partial X\partial Y})=0
$. Then we use the positive direction$\ \pi=\sum_{k=1}^{K}{w_klogp_\epsilon\left(y_k-f_k\left(x_k\right)\right)+log}p(X)]$ to obtain equation (7). That is to say, in general, there will not be a HANM satisfying the condition from $Y\rightarrow X$.

The idea of this proof follows from ANM-MM and ANM, but the specific items in it are inconsistent with it. This lemma 1 shows that if a forward HANM exists, it is almost impossible to have a reverse HANM. Therefore, we strengthen the assumption that there is a forward HANM $X\rightarrow Y$, and thus the following theorem exists.

\noindent \textbf{\emph{Theorem 1.}} \emph{Let X→Y, and there is a forward hybrid ANM, if a reverse hybrid ANM}
 $$X = \sum_{k=1}^{K}{w_k(g_k\left(y_k\right)+{\widetilde{\epsilon}}_k)}, \quad \text{\emph{subject to}} \quad {\widetilde{\epsilon}}_k \perp \!\!\! \perp y_k.$$
\emph{exists, the following equation must be satisfied}.
\begin{equation} \label{(21)}
   \xi^{\prime\prime\prime}=\frac{G_k\left(X,Y\right)}{M_k\left(X,Y\right)}\xi^{\prime\prime}+\frac{U_k\left(X,Y\right)G_k\left(X,Y\right)}{M_k\left(X,Y\right)}-H_k\left(X,Y\right)
\end{equation}
\emph{where $\xi$, $G_k$, $M_k$, $U_k$, and $H_k$ are the same as Lemma 1}.

\noindent\textbf{\emph{Proof.}} Assuming that there is a HANM in the reverse direction, as in the proof of Lemma 1, according to the fact that the reverse and forward $p\left(X,Y\right)$ are equal, then through$\ \ \frac{\partial}{\partial X}(\frac{\partial^2 \pi/\partial X^2}{\partial^2 \pi/\partial X\partial Y})=0$, here $\pi=logp\left(X,Y\right)$, it can be obtained that equation (21) holds. In other words, the set of solutions $logp_X$ is contained in a three-dimensional affine space. It is almost impossible to exist a hybrid ANM satisfying the condition from $Y\rightarrow X$.

\subsection{Proposed likelihood criterion based on the HANM}
In hybrid ANM method, there will be no reverse smooth HANM in general. Similar to the CANM method, we use the variational method to solve the likelihood and use the log-likelihood as the causality criterion. But here, we neither focus on the intermediate hidden variable problem nor assume that the distribution of noise variables is normal. Compared with the independence of $\epsilon$ and $X$ obtained by regressing $Y$ to get a casual direction in ANM, we switch to solving the likelihood to judge the causality of the proposed hybrid model. As show in formula (22), when $X\perp \!\!\! \perp {\epsilon} $, we have:
\begin{equation} 
\begin{aligned}
   logp\left(X,\epsilon\right)&=logp\left(X\right)+\sum_{k=1}^{K}{w_klogp_{\epsilon_k}\left(y_k-f_k\left(x_k\right)|x_k\right)} \\
   &=\log{\left(X\right)}+\log{\left(Y\middle| X\right)}
\end{aligned}
\end{equation}

We mainly concentrate on binary variables under fully observed data. Here, we establish a mixture conditional variational auto-encoder to solve $logp(Y|X)$, which can fit each component function to get Y to deal with $logp\left(X,\epsilon\right)$.

Now turn to solving the problem on how to get $logp\left(X,\epsilon\right)$. Fig. 1. shows the proposed mixture conditional variational encoder, our regression scheme, to regress $Y$. In the conditional encoding part, we use the Enconder2 module for conditional feature expression to get $Z_{prior}$. For the expression of hidden variables, we use $K$ MLPs structures for Gaussian mixture expression to obtain the mean ${\{{\mu}_k\}}_{k=1}^K$ and variance ${\{{\sigma}_k\}}_{k=1}^K$.Then, use the reparameterization technique to get ${\{{Z}_k\}}_{k=1}^K$, and use the condition $Z_{con}$ and ${\{{Z}_k\}}_{k=1}^K$ to decode to obtain ${\{{\widetilde{Y}}_k\}}_{k=1}^K$. $q(c_k|X,Y)$ indicates the possibility
\begin{figure}[!t]
\centering
\includegraphics[width=3.8in]{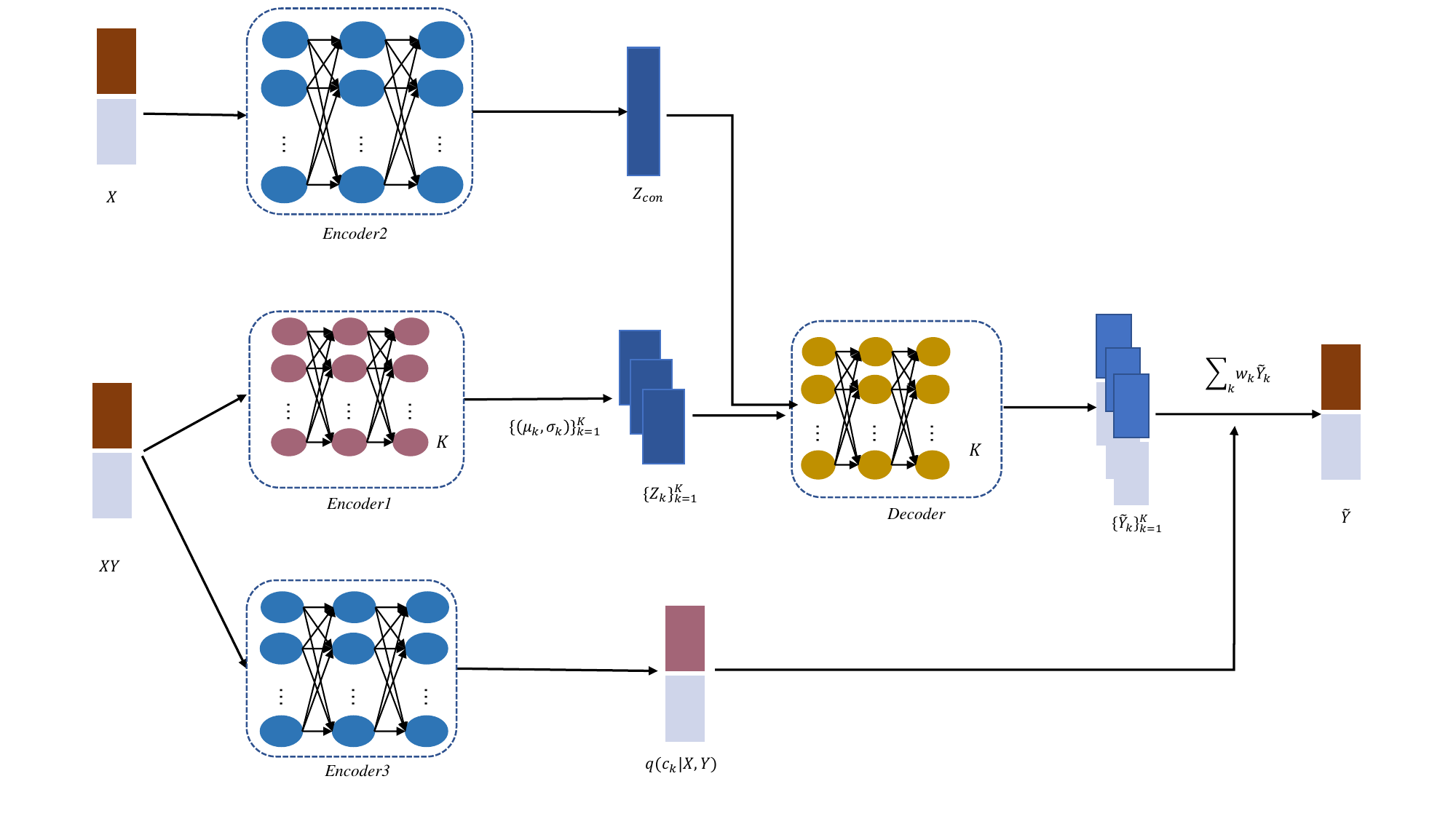}
\caption{Regression model of the mixture conditional variational auto-encoder.}
\label{fig_1}
\end{figure}
that the current data sample belongs to the $kth$ component.Where $p\ \left(c_k=1\right)=w_k$ and $\sum_{k=1}^{K}w_k$ is calculated by the Encoder3 part with the Softmax layer. Finally, we use the obtained mixed weight to perform the sum of $w_k{\{{\widetilde{Y}}_k\}}_{k=1}^K$ to solve the regression variable $\widetilde{Y}$ we want.

Referring to CVAE \cite{ref15}, the Loss function of the regression model we constructed can be derived with$\ logp(Y|X)$. For a given observation $X$, $Z_{con}$ is obtained from the condition distribution $p_\theta\left(Z_{con}\middle| X\right)$. And the output Y is generated by the distribution $P_\theta\left(Y\middle| X,Z\right)$, $Z_{con}$ and$\ Z_k$ form $Z$ together, where $k=1,\cdots,K$. Therefore, we have the Equation (23).

\begin{equation} 
\begin{aligned}
  &logp_\theta\left(Y\middle| X\right)=E_{q\left(Z,c_k\middle| X,Y\right)}\left[log\frac{p_\theta\left(Y,Z,c_k\middle| X\right)}{p_\theta\left(Z,c_k\middle| X,Y\right)}\right] \\
   &=E_{q\left(Z,c_k\middle| X,Y\right)}\left[log\frac{p_\theta\left(Y,Z,c_k\middle| X\right)}{q_\varphi\left(Z,c_k\middle| X,Y\right)}\frac{q_\varphi\left(Z,c_k\middle| X,Y\right)}{p_\theta\left(Z,c_k\middle| X,Y\right)}\right]\\
   &=E_{q\left(Z,c_k\middle| X,Y\right)}\left[logp_\theta\left(Y,Z,c_k\middle| X\right)-logq_\varphi\left(Z,c_k\middle| X,Y\right)\right]\\
   &+KL(q_\varphi\left(Z,c_k\middle| X,Y\right)||p_\theta\left(Z,c_k\middle| X,Y\right)
\end{aligned}
\end{equation}

Because $KL(q_\theta\left(Z,c_k\middle| X,Y\right)||p_\theta\left(Z,c_k\middle| X,Y\right)\geq0$, then
\begin{equation} 
\begin{split}
  &\log p_\theta\left(Y\middle| X\right) \geq  E_{q\left(Z,c_k\middle| X,Y\right)}\left[\log\frac{p_\theta\left(Y,Z,c_k\middle| X\right)}{p_\theta\left(Z,c_k\middle| X,Y\right)}\right] \\
  &= E_{q\left(Z,c_k\middle| X,Y\right)}\left[\log p_\theta\left(Y,Z,c_k\middle| X\right)-\log q_\varphi\left(Z,c_k\middle| X,Y\right)\right] \\
  &= E_{q\left(Z,c_k\middle| X,Y\right)}\left[\log p_\theta\left(Z_{con},c_k\middle| X\right) + \log p_\theta\left(Y\middle| Z,X,c_k\right) \right.\\
  &\quad - \left. \log q_\varphi\left(Z,c_k\middle| X,Y\right)\right] \\
  &= E_{q\left(Z,c_k\middle| X,Y\right)}\left[\log p_\theta\left(Y\middle| Z,X,c_k\right) \right.\\
  &\quad - \left.KL\left(q_\varphi(Z,c_k|X,Y)\parallel p_\theta(Z_{con},c_k|X)\right)\right]\\
  &:=ELBO
\end{split}
\end{equation}

In summary, a mixture conditional variational generative model is built to regress y by maximizing $ELBO$, which is also minimizing the Loss of the constructed network, where $Loss=-ELBO$ and ELBO is the lower bound of the conditional probability. Then use the adaptive Adam optimization algorithm to optimize the entire network.
Because $KL(q_\theta\left(Z,c_k\middle| X,Y\right)||p_\theta\left(Z,c_k\middle| X,Y\right)\geq0$, then
\begin{equation}
\begin{split}
  &L_{X\rightarrow Y} = \log p\left(X,\epsilon\right) \\
  &= \log p\left(X\right) + E_{q\left(Z,c_k\middle| X,Y\right)}\left[\log p_\theta\left(Y\middle| Z,X,c_k\right) \right.\\
  &\quad - \left. KL\left(q_\varphi\left(Z,c_k\middle| X,Y\right)\parallel p_\theta(Z_{con},c_k|X)\right)\right]
\end{split}
\end{equation}

\begin{algorithm}[H]
\caption{The MCVCI algorithm.}\label{alg:alg1}
\begin{algorithmic}
\STATE 
\STATE \textbf{Input:}$X={\{x_i\}}_{i=1}^m$, $Y={\{y_i}\}_{i=1}^m$,dataset $D={X,Y}$, and learning rate $\lambda$.
\STATE \textbf{Output:}The casual direction.
\STATE 1: Standardize the dataset $D$ and divide the data into training and testing sets;
\STATE 2: Compute the correlation $ corr(X,Y)$, if $corr>0$, then perform next step 3, else perform step 7.
\STATE 3:  hyperparameters $K$  $  \gets\ $  training the model in Fig. 1. by formula (24);
\STATE 4: $L_{X\rightarrow Y}$ $\gets\ $ Use formula (25) through the test set;
\STATE 5: $L_{Y\rightarrow X}$ $\gets\ $  repeat the steps 2-3, retrain the reverse model, and use the formula (26);
\STATE 6: Compare the value of $L_{X\rightarrow Y}$ and $L_{Y\rightarrow X}$;
\STATE 7: \textbf{Return} the casual relationship of $X$ and $Y$.
\end{algorithmic}
\label{alg1}
\end{algorithm}

\begin{equation}
\begin{split}
  & L_{Y\rightarrow X} = \log p\left(Y,\widetilde{\epsilon}\right) \\
  &= \log p\left(Y\right) + E_{q\left(Z,c_k\middle| X,Y\right)}\left[\log p_\theta\left(X\middle| Z,Y,c_k\right) \right.\\
  &\quad - \left. KL\left(q_\varphi\left(Z,c_k\middle| X,Y\right)\parallel p_\theta(Z_{con},c_k|Y)\right)\right]
\end{split}
\end{equation}

Eventually, we propose a causal inference algorithm MCVCI based on the mixture likelihood, where forward likelihood is the formula (25). The detailed steps of algorithm MCVCI are given in Algorithm 1. 

In Step 2, this means that when the correlation value $corr>0$, continue to judge the causal relationship between $X$ and $Y$, otherwise the output $X$ and $Y$ has no causal relationship. In Step 7, if $L_{X\rightarrow Y}>L_{Y\rightarrow X}$, then return $X\rightarrow Y$; if $L_{X\rightarrow Y}<L_{Y\rightarrow X}$, we can get the result $Y\rightarrow X$, and in other cases we cannot decide the casual relationship.

\subsection{ Causality Mechanism Clustering}
In the causal mechanism clustering part, we assume that the data correspond to our proposed hybrid additive noise model. The scenario of this mechanism clustering is more suitable for label offset due to certain factors. And our proposed algorithm is to cluster the shifted categories under the factors with relatively significant influence. We use the $w\epsilon_c$ term we seek in the true causal direction as the extracted causal feature space and then cluster on it. Here $w$ control the shifted values. We regard $w\epsilon_c$ term as $\vartheta$ and $u$ is the cluster center.

Therefore, our clustering objective function is shown in Equation (27), and the $C$ is the cluster number.
Assuming a causal relationship exists between the two variables, our algorithm 2 MCVCC has been shown in the paper.
\begin{equation} \label{(27)}
   \mathrm{\Psi}=argmin \sum_{i=1}^{C} \left \|\vartheta-u_i  \right \|^2 
\end{equation}

\section{EXPERIMENTAL RESULTS}
In this chapter, we first validate the proposed causal inference method MCVCI method on the three publicly simulated datasets and real data CEP, composed of 41 datasets. And in the application scenario of our proposed causal mechanism clustering method MCVCC, we constructed our simulateddataset for verification. At last, we verified the effectiveness of the causal mechanism clustering method on real BAFU air data.
\begin{minipage}{\linewidth}
\begin{algorithm}[H]
\caption{The MCVCC algorithm.}\label{alg:alg2}
\begin{algorithmic}
\STATE 
\STATE \textbf{Input:}$X={\{x_i\}}_{i=1}^m$, $Y={\{y_i}\}_{i=1}^m$,dataset $D={X,Y}$, and learning rate $\lambda$,Cluster number $C$.
\STATE \textbf{Output:}the clustering labels.
\STATE 1: Standardize the dataset $D$ and divide the data into training and testing sets;
\STATE 2:  hyperparameters $K$  $  \gets\ $  training the model in Fig. 1. by formula (24);
\STATE 3: $L_{X\rightarrow Y}$, $\widetilde{Y}$ $\gets\ $ Use formula (25) through the test set;
\STATE 4: $L_{Y\rightarrow X}$, $\widetilde{X}$ $\gets\ $  repeat the step 2, retrain the reverse model, and use the formula (26);
\STATE 5: $\vartheta$ $\gets$ Compare the value of $L_{X\rightarrow Y}$ and $L_{Y\rightarrow X}$; if $L_{X\rightarrow Y}<L_{Y\rightarrow X}$, perform $\vartheta=X-\widetilde{X}$, else $\vartheta=Y-\widetilde{Y}$;
\STATE 6: use function (27) to cluster on $\vartheta$;
\STATE 7: \textbf{Return} the clustering labels.
\end{algorithmic}
\label{alg2}
\end{algorithm}
\end{minipage}

\subsection{Experimental results and analysis of MCVCI}
\textbf{\emph{1) Introduction to comparison Methods.}} The classic methods LINGAM \cite{ref32}, ANM \cite{ref6}, IGCI \cite{ref7}, and PNL \cite{ref5} are included in the comparison algorithms. biCAM \cite{ref26} is a high dimensional based additive noise method. CURE \cite{ref25} uses the principle that the probability distribution $p_x$ cannot help $x$ to regress $y$, but $p_y$ may help $y$ to regress $x$ to determine cause and effect. RESIT [22] is a continuous additive noise model. QCCD \cite{ref10}, NNCL \cite{ref19}, and HEC \cite{ref11} are all causal methods for regression improvements in recent years. Sloppy \cite{ref18} and RECI \cite{ref8} are the model based on the regression noise error to determine the cause and effect. The neural network is used as one of the four basic regression methods in the RECI method. Here we regard CANM \cite{ref23} as a causal identification method of likelihood based on VAE regression. The ANM-MM \cite{ref12} method is the only causal model that uses the mixture mechanism.

In the comparison methods, the most experimental results without the marker $*$ are those obtained from the open-source code we ran. In contrast, the methods with the marker * are the results from the article QCCD or the original paper. Since RECI has no public source code in the author's paper, we found a version of RECI-PLOY in the toolkit \cite{ref20}. Then we implemented the RECI-nn with a three fully connected neural network for regression and used mean squared error as the loss function.

\textbf{\emph{2) On public simulated datasets.}} We use three publicly available artificial datasets in the paper \cite{ref24}, including SIM, SIM-G, and SIM-ln artificial datasets, each consisting of 100 causal pairs. While the SIM data set has no confounding factors, the SIM-G distribution is close to a Gaussian distribution, and the SIM-ln data is low-noise. The general form of the three datasets is

\begin{equation}
\left\{
\begin{aligned}
    & x^\prime\sim P_x,\epsilon\sim P_\epsilon \\
    & \epsilon_x \sim \left(0,\sigma_x\right),\epsilon_y\sim\left(0,\sigma_y\right) \\
    & x=x^\prime+\epsilon_x,y=f_y\left(x^\prime,\epsilon\right)+\epsilon_y
\end{aligned}
\right.
\end{equation}
where $\epsilon$ is addictive noise. For the specific parameter settings of the simulated data set, see the appendix of the paper \cite{ref24}. 

TABLE I first illustrates the causal inference accuracy of MCVCI and the comparison algorithms on the SIM, SIM-G, and SIM-ln datasets.There are no ablation experiments here because we merged these results into this table. HEC is a comparison of heterogeneous noise models. For the RECI-PLOY and RECI-nn method, using polynomial and DNN neural networks as a regression model, we mainly compare the regression error-type causal methods.

Overall, in TABLE I our algorithm MCVCI obtains the best experimental performance on these public simulated datasets. As all simulated datasets of the comparison experiments in this section are constructed using the GP algorithm. ANM uses the GP algorithm for regression, thus achieving a good causal identification accuracy. Although CANM is a causal algorithm for intermediate confounding variables, VAE-based regression superiority also performed well on SIM-G and SIM-ln. The comparison between the CANM and our methods proves that our proposed mixture CVAE regression part outperforms the VAE-based model.

\begin{table*}[!t]
\caption{THE INFERENCE ACCURACY OF MCVCI AND THE COMPARISON ALGORITHMS ON DIFFERENT TYPE DATA.\label{tab:table1}}
\centering
\begin{tabular}{c|ccccccccc}
\hline
Data   & LINGAM  & ANM    & IGCI* & PNL*      & biCAM*  & CURE* & RESIT* & QCCD*         \\ \hline
SIM    & 0.4     & 0.75   & 0.42  & 0.7       & 0.57    & 0.57  & 0.78   & 0.49          \\
SIM-G  & 0.28    & 0.71   & 0.54  & 0.64      & 0.78    & 0.5   & 0.77   & 0.76          \\
SIM-ln & 0.29    & 0.77   & 0.52  & 0.61      & 0.87    & 0.62  & 0.87   & 0.77          \\
CEP    & 0.6     & 0.6    & 0.67  & 0.64      & 0.57    & 0.6   & 0.53   & 0.66          \\ \hline
Data   & Sloppy* & NNCL   & HEC   & RECI-PLOY & RECI-nn & CANM  & ANM-MM & MCVCI         \\ \hline
SIM    & 0.64    & 0.6795 & 0.49  & 0.44      & 0.61    & 0.51  & 0.52   & \textbf{0.88} \\
SIM-G  & 0.81    & 0.709  & 0.56  & 0.39      & 0.77    & 0.77  & 0.4    & \textbf{0.87} \\
SIM-ln & 0.77    & 0.58   & 0.65  & 0.68      & 0.69    & 0.85  & 0.4    & \textbf{0.93} \\
CEP    & 0.74    & 0.6    & 0.59  & 0.63      & 0.62    & 0.54  & 0.57   & \textbf{0.81} \\ \hline
\end{tabular}
\end{table*}

\textbf{\emph{3) On real data CEP.}} We use the CEP dataset [26] of the causal research team at the University of Tübingen, which is relatively common in causal pairs data. There are 108 in the latest updated data, and causal pairs consist of 41 datasets. Among them, six causal pairs are not included because of high dimensionality, including pair 52, 53, 54, 55, 71, and 105. While causal pairs 107 and 108 are excluded as there is no comparison in the paper of the QCCD. In the parameter selection part, for each of the 37 datasets, we selected an appropriate hyperparameter $K$ through training the model. TABLE I shows the causal inference accuracy of MCVCI and the comparison algorithms on the CEP dataset. Overall, the algorithm MCVCI we proposed has the highest inference accuracy, and Sloppy is based on the improvement of RECI and also offers a good performance.

\textbf{\emph{4)MCVCI Confidence Analysis}}. In the RECI and sloppy methods, the minimum and maximum values of the error items, which can divide the casual directions, can be expressed as confidence estimation. Using the same strategy, we correspondingly propose a confidence measure in our decision and define it as:
$$\tau = 1 - \frac{\min(L_{X\rightarrow Y}, L_{Y\rightarrow X})}{\max(L_{X\rightarrow Y}, L_{Y\rightarrow X})}$$
The higher the value of $\tau$, the more correct our decision will be. Furthermore, we can set a threshold t to require $\tau\geq t$. When $\tau<t$,The higher the value of $\tau$, the more correct our decision will be. Furthermore, we can set a threshold t to require $\tau\geq t$. When $\tau<t$, we can think its credibility is not high. In other words, the causal direction cannot be determined.
\begin{figure}[!t]
\centering
\includegraphics[width=3.6in]{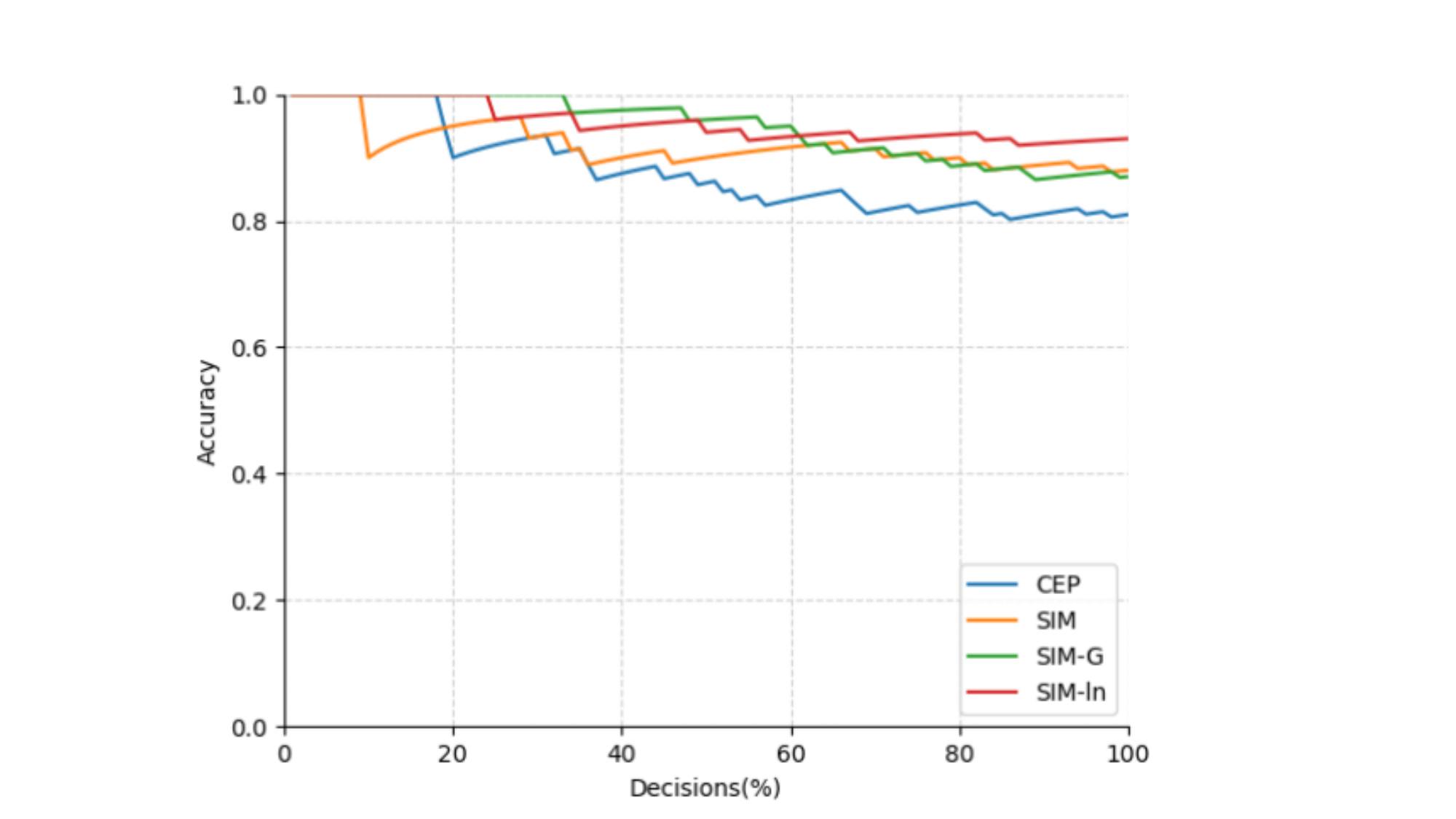}
\caption{Decision rate curves of MCVCI under the top $k\%$ on different datasets.}
\label{fig_2}
\end{figure}

Highly correlated with the confidence value is the decision rate. In particular, if we rank a set of decisions in order of the top $k\%$, we get a decision rate confidence value. Fig. 2. shows the decision rate of the top $k\%$ of MCVCI under several different datasets. For most datasets, our top 10\% of decisions are correct. On the whole, our decision-making accuracy is higher than 80\%.

\begin{table*}[!t]
\caption{CLUSTERING RESULTS OF COMPARISON METHODS AND MCVCC ON DIFFERENT FUNCTIONS CONDITION.\label{tab:table2}}
\centering
\begin{tabular}{c|cccccccccc}
\hline
             & \multicolumn{2}{c}{$f_1$}      & \multicolumn{2}{c}{$f_2$}         & \multicolumn{2}{c}{$f_3$}          & \multicolumn{2}{c}{$f_4$}          & \multicolumn{2}{c}{$f_5$}          \\ \hline
Measure (\%) & ARI          & NMI          & ARI           & NMI            & ARI            & NMI            & ARI            & NMI            & ARI            & NMI            \\
k-means      & 0.72         & 0.89         & 3.12          & 2.63           & -0.5           & 0              & 4.84           & 4.15           & -0.14$\pm$0.03           & 0.25$\pm$0.03\\
SpeClu       & 0.72         & 0.89         & 2.07          & 1.85           & -0.47          & 0              & 0.0485         & 4.44              & 0.12         & 3.55           \\
GMM          & 0            & 0.58         & 7.3           & 2.41           & 12.7           & 0.59           & 18.72$\pm$11.45          & 0.58$\pm$16.12           & -0.04          & 0.43           \\
CVAE-km      & 0.31         & 0.59         & 0.31          & 0.59           & 0.31           & 0.59           & 0.31           & 0.59           & 0.05$\pm$0.05           & 0.42$\pm$0.04           \\
ANM-MM       & 98           & 95.95        & 62.22         & 51.89          & 22.66          & 18.27          & 67.07          & 58.16          & 3.19$\pm$0.81          & 4.02$\pm$0.62          \\
MCVCC        & \textbf{100} & \textbf{100} & \textbf{88.3} & \textbf{83.47} & \textbf{47.36} & \textbf{43.95} & \textbf{84.56} & \textbf{77.51} & \textbf{43.28} & \textbf{34.67} \\ \hline
\end{tabular}
\end{table*}

\begin{table*}[!t]
\caption{CLUSTERING RESULTS OF COMPARISON METHODS AND MCVCC ON DIFFERENT CLUSTER NUMBERS CONDITION.\label{tab:table3}}
\centering
\begin{tabular}{c|cccccc}
\hline
             & \multicolumn{2}{c}{$C=2$}             & \multicolumn{4}{c}{$C$}                                    \\
             & \multicolumn{2}{c}{(a) $\sigma=0.2, 0.05$} & \multicolumn{2}{c}{(b) $C=3$} & \multicolumn{2}{c}{(c) $C=4$} \\ \hline
Measure (\%) & ARI              & NMI              & ARI          & NMI          & ARI          & NMI         \\
k-means      & -0.33            & 0.12             & 15.48        & 24.61        & 21.07        & 36.84       \\
SpeClu       & -0.39            & 0.07             & 18.78        & 26.78        & 23.46        & 39.14       \\
GMM          & -0.48            & 0.58             & 44.14$\pm$0.18        & 0.87$\pm$0.37         & 28.42$\pm$0.26        & 1.75$\pm$1.01        \\
CVAE-km      & 0.31             & 0.58             & 0.38         & 0.87         & 0.86$\pm$0.01         & 1.76$\pm$0.01        \\
ANM-MM       & 19.85            & 15.7             & 78.98        & 76.49        & 53.67$\pm$0.04        & 59.95$\pm$0.11       \\
MCVCC        & \textbf{57.55}            & \textbf{49.29$\bm{\pm}$1.48}            & \textbf{84.26}        & \textbf{83.83}        & \textbf{58.52$\bm{\pm}$1.69}        & \textbf{67.27$\bm{\pm}$1.1}       \\ \hline
\end{tabular}
\end{table*}
\subsection{Experiment Results and Analysis of MCVCC}
\textbf{\emph{1) Comparison methods}}. In the previous part, we thoroughly analyzed our causal method MCVCI. Causal clustering is an application to MCVCI via data characteristics, so we only compared the methods related to us. K-means \cite{ref27} is a classic method based on Euclidean distance as a similarity measure. Spectral clustering \cite{ref28} is suitable for nonlinear data, referred to SpeClu here. Because our regression method is an improvement on CVAE, the method which uses k-means clustering after CVAE extracts features is also our competitor,
which we abbreviate as CVAE-km. As for GMM \cite{ref29}, the idea of GMM is related to our MCVCI method. ANM-MM is the only method we found for binary causal mechanism clustering. We use the traditional clustering measures ARI \cite{ref31} and NMI \cite{ref30} for evaluation. And in experiments related to k-means, we used k-means to initialize the cluster center to keep it stable. We ran the experiment 20 times and obtained the following ARI and NMI mean values and standard deviation.

\textbf{\emph{2) Causal mechanism clustering and analysis on constructed simulated datasets.}} We first tested our method MCVCC under several different mechanisms, where $f_1$ is $y=a_C\left(\frac{1}{1+{x_C}^2}+\epsilon_C\right)$, $f_2$ is $y = a_C \left(\exp(-2x_C) + \epsilon_C\right)$, $f_3$ is $y=a_C({x_C}^2+\epsilon_C)$, $f_4$ is $y=a_C\left(tanh{\left(x_C\right)}+\epsilon_C\right)$, and $f_5$ is $y=a_C\left(log{\left({5x}_C\right)}+\epsilon_C\right)$. We mainly control the different offsets of the data through $a_C$. Among them, $ a_1 \sim \mathcal{U}(1, 1.1)$, $a_2 \sim \mathcal{U}(0.5, 0.6)$, we set the data number for each 
class as 100, $\epsilon \sim N\left(0,\sigma\right)$, the default value of $\sigma$ is 0.05. Clustering results of comparison methods and MCVCC on different conditions are shown in TABLE II and TABLE III.

First of all, we show the mean value and standard deviation of ARI and NMI of MCVCC under different functions condition on TABLE II when the cluster number $C$ is 2 with $a=a_1$ and $a_2$. On TABLE III, (a) shows $f=f_2$ with $a=a_1$, and $\sigma=0.2$ and 0.05, (b) set $C=3$ that is we mixed $f_1$ with $a_1$ and $a_2$, $f_2$ with $a_1$. And (c) $C=4$ is the experiment with mixed data under conditions $f_1$ and $f_2$ with $a_1$ and $a_2$.

Overall, the MCVCC method ARI and NMI have the highest mean values relative to the other compared methods in TABLE  II and TABLE III. Although its performance decreases when the mixed number increases, the MCVCC method is still much better than existing methods. Additionally, the visualization of the best results (we plot the figure when getting the max ARI value) of all methods is shown below in Fig. 3.-10.

\textbf{\emph{3) Causal mechanism clustering on BAFU air data.}} Following the ANM-MM paper, we evaluated the average ARI and NMI values for MCVCC on real data using BAFU air data. This data included daily ozone and temperature values at two locations in Switzerland in 2009. In our experiments, the data was generated from two locations, with location as a factor influencing temperature and ozone. Finally, we hoped to cluster the data by the locations factor. After we preprocessed the data by deleting the null data, our results are shown in TABLE  IV and Fig. 11., where the MCVCC obtained the best results.

\begin{table*}[!t]
\caption{CLUSTERING RESULTS OF COMPARISON METHODS AND MCVCC ON DIFFERENT CLUSTER NUMBERS CONDITION.\label{tab:table3}}
\centering
\footnotesize
\begin{tabular}{l|cccccc}
\hline
    & k-means & SpeClu & GMM   & CVAE-km & ANM-MM & MCVCC          \\ \hline
ARI & -0.12   & 0.99   & -0.12 & 1.29    & 11.72  & \textbf{47.35} \\
NMI & 0.03    & 1.05   & 0.03  & 2.59    & 9.37   & \textbf{38.21} \\ \hline
\end{tabular}
\end{table*}

\section{CONCLUSION AND FUTURE WORK}
We have developed a Mixture Conditional Variational Causal Inference model MICVCI for observational data. Based on our hybrid additive noise model, we constructed our hybrid likelihood principle for causal identification through a proposed mixture conditional variational auto-encoder. Then, by utilizing the causal mechanism of the MCVCI method, we proposed the second algorithm MCVCC, which can exhibit causal  mechanism clustering of data in specific scenarios. The MCVCI and MCVCC methods show superior performance on Gaussian and other several different types of data. In future work, MCVCI can be extended to higher dimensions. It is also achievable to improve the MCVCC method by improving the causal method part to enhance causality expression, thus reducing the sensitivity to the mixture numbers.

\begin{figure*}[!t]
\centering
\subfloat[]{\includegraphics[width=1.8in]{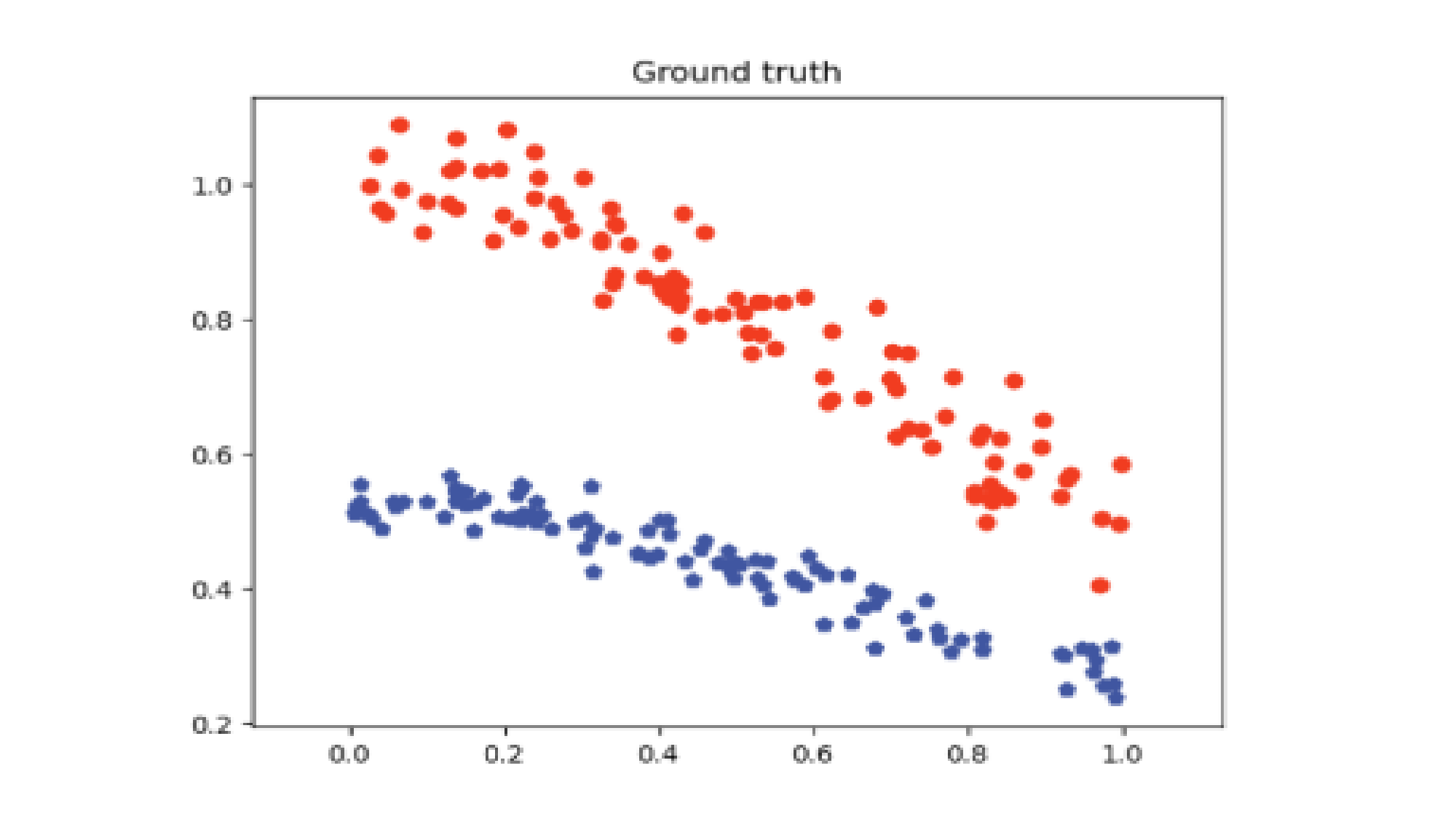}%
\label{fig_groundtruth}}
\subfloat[]{\includegraphics[width=1.8in]{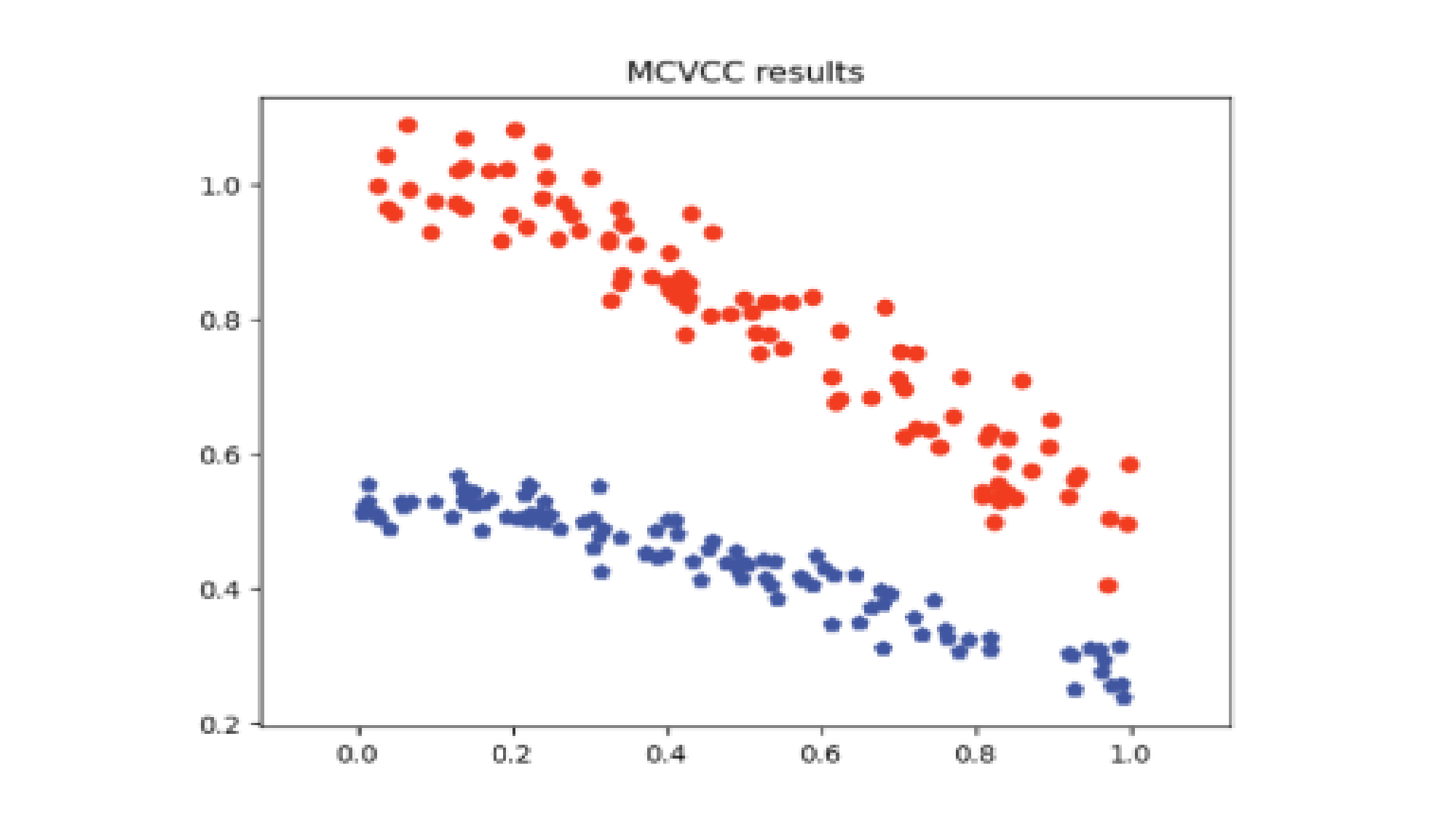}%
\label{fig_mcvcc}}
\subfloat[]{\includegraphics[width=1.8in]{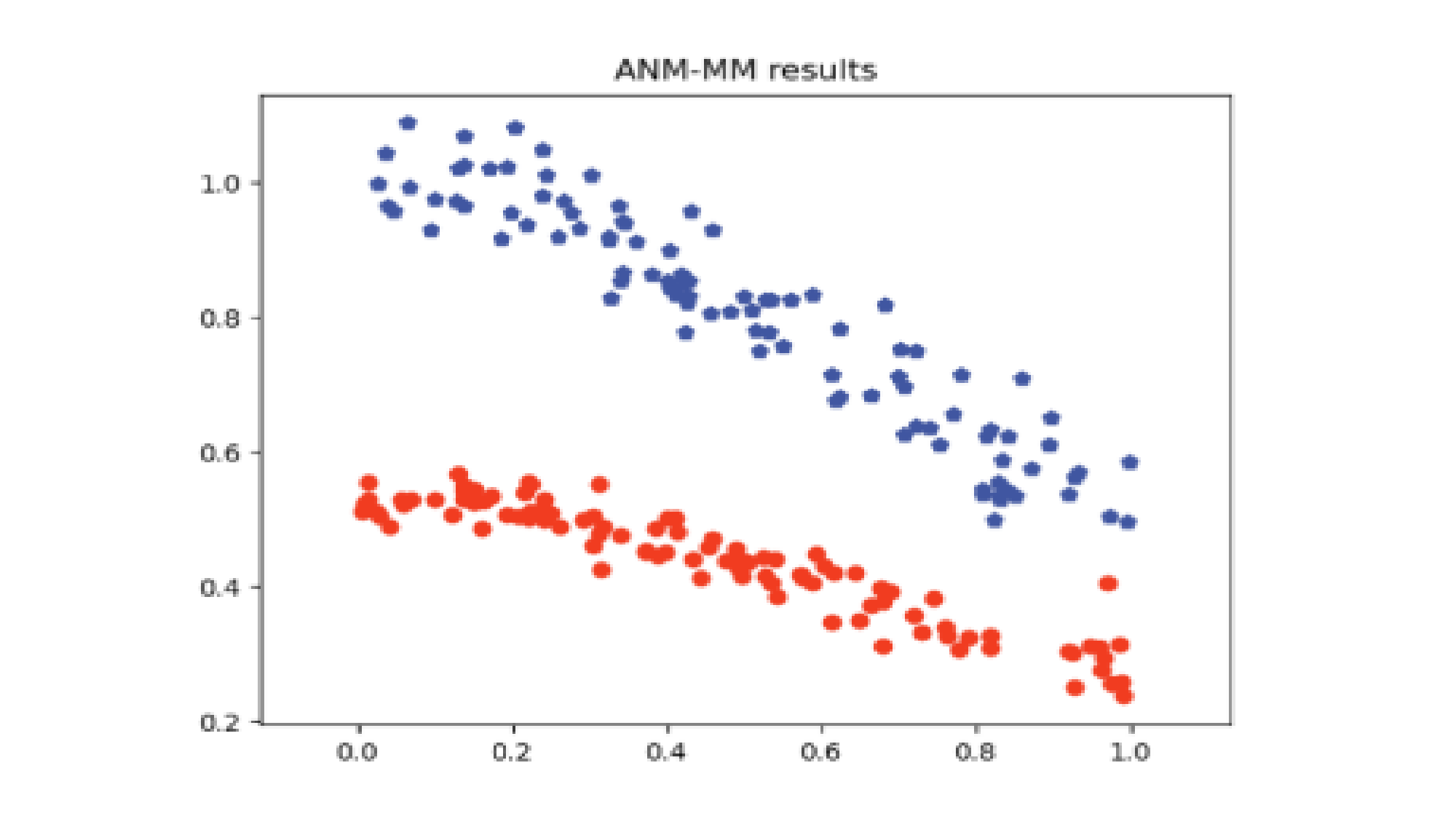}%
\label{fig_anm_mm}}
\subfloat[]{\includegraphics[width=1.8in]{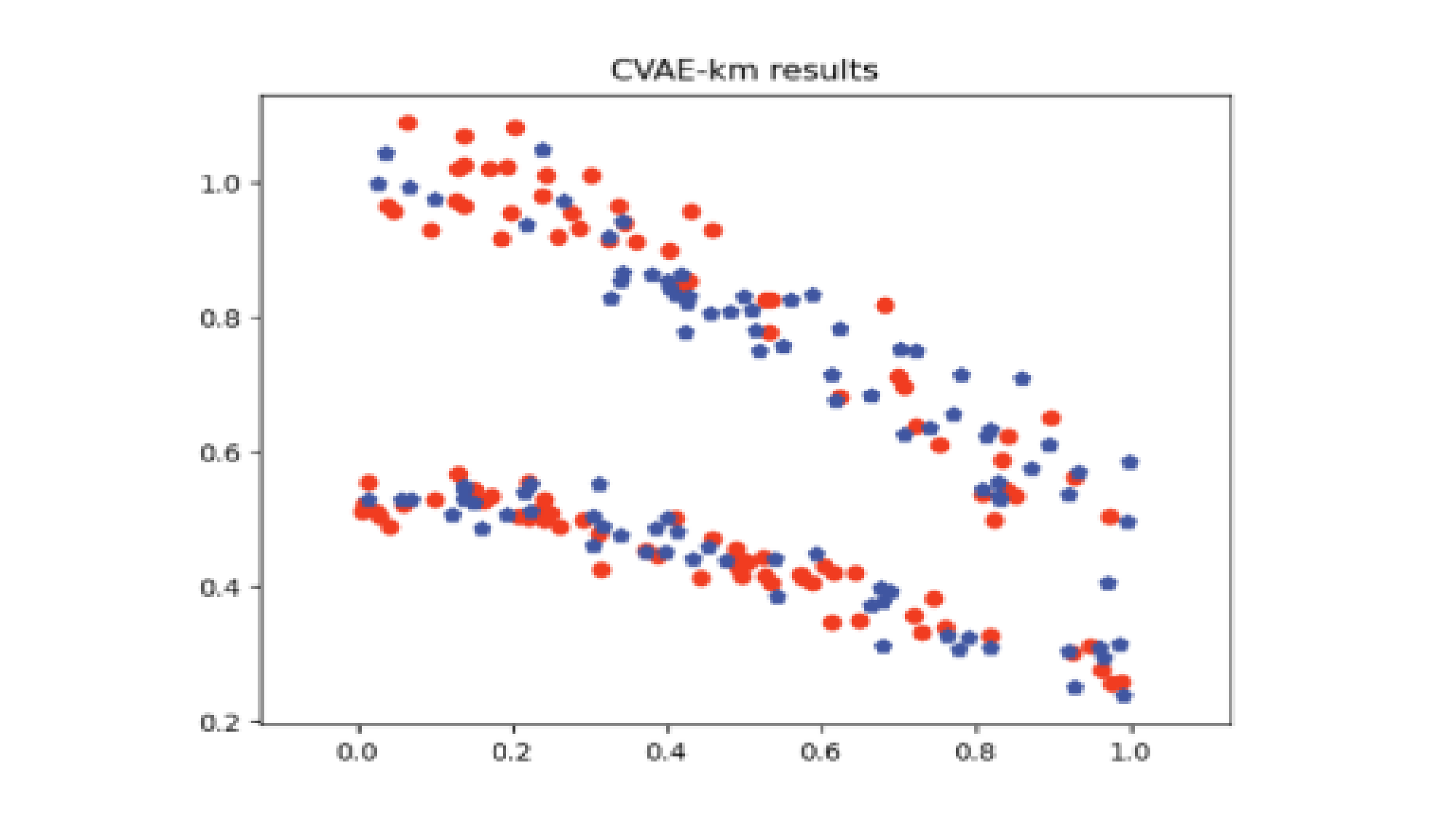}%
\label{fig_cvae_km}}
\hfil
\subfloat[]{\includegraphics[width=1.8in]{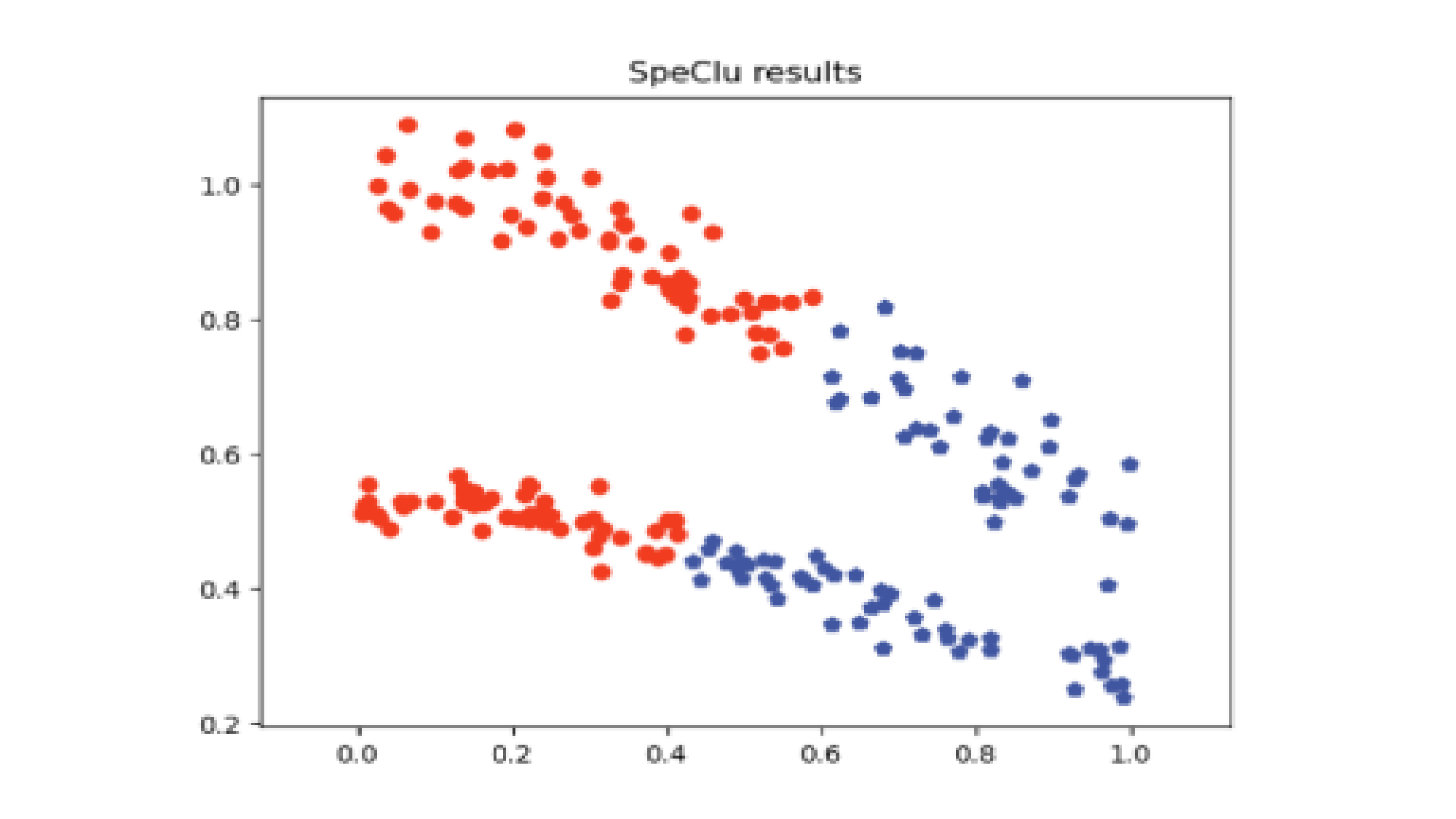}%
\label{fig_speclu}}
\subfloat[]{\includegraphics[width=1.8in]{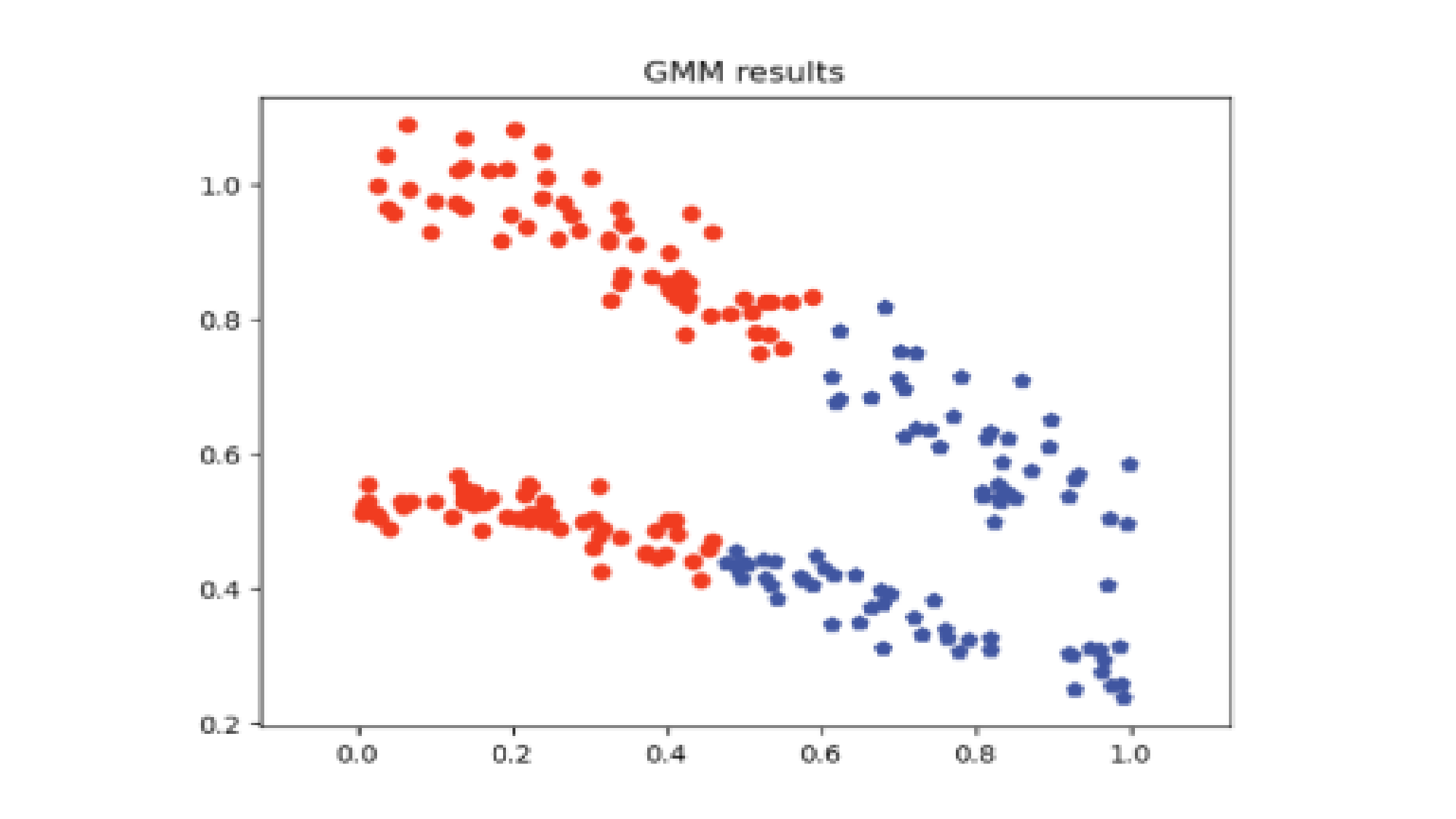}%
\label{fig_gmm}}
\subfloat[]{\includegraphics[width=1.8in]{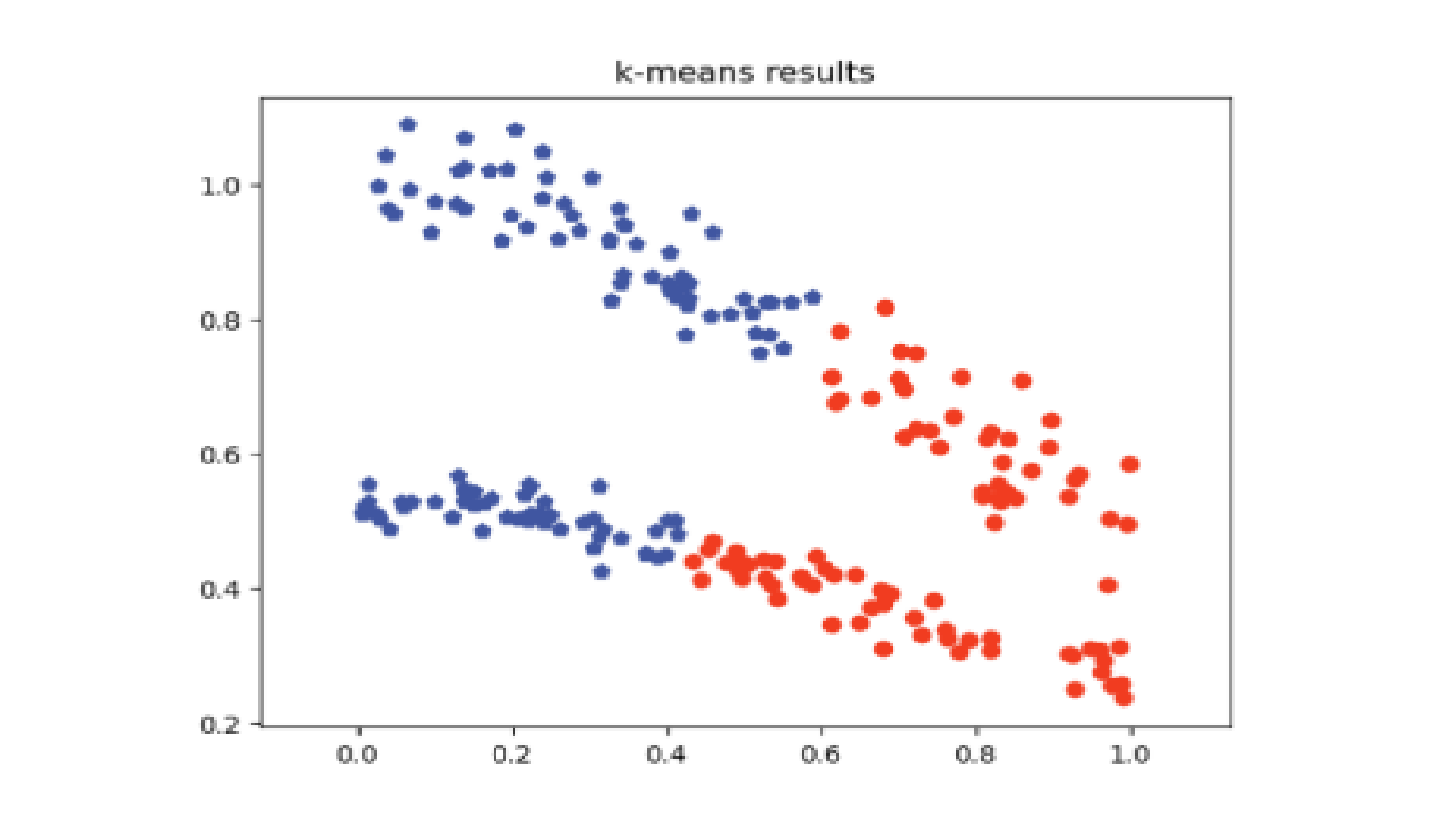}%
\label{fig_kmeans}}
\caption{The clustering results of MCVCC and comparison algorithms when $f=f_1$.}
\label{fig_comparison}
\end{figure*}

\begin{figure*}[!t]
\centering
\subfloat[]{\includegraphics[width=1.8in]{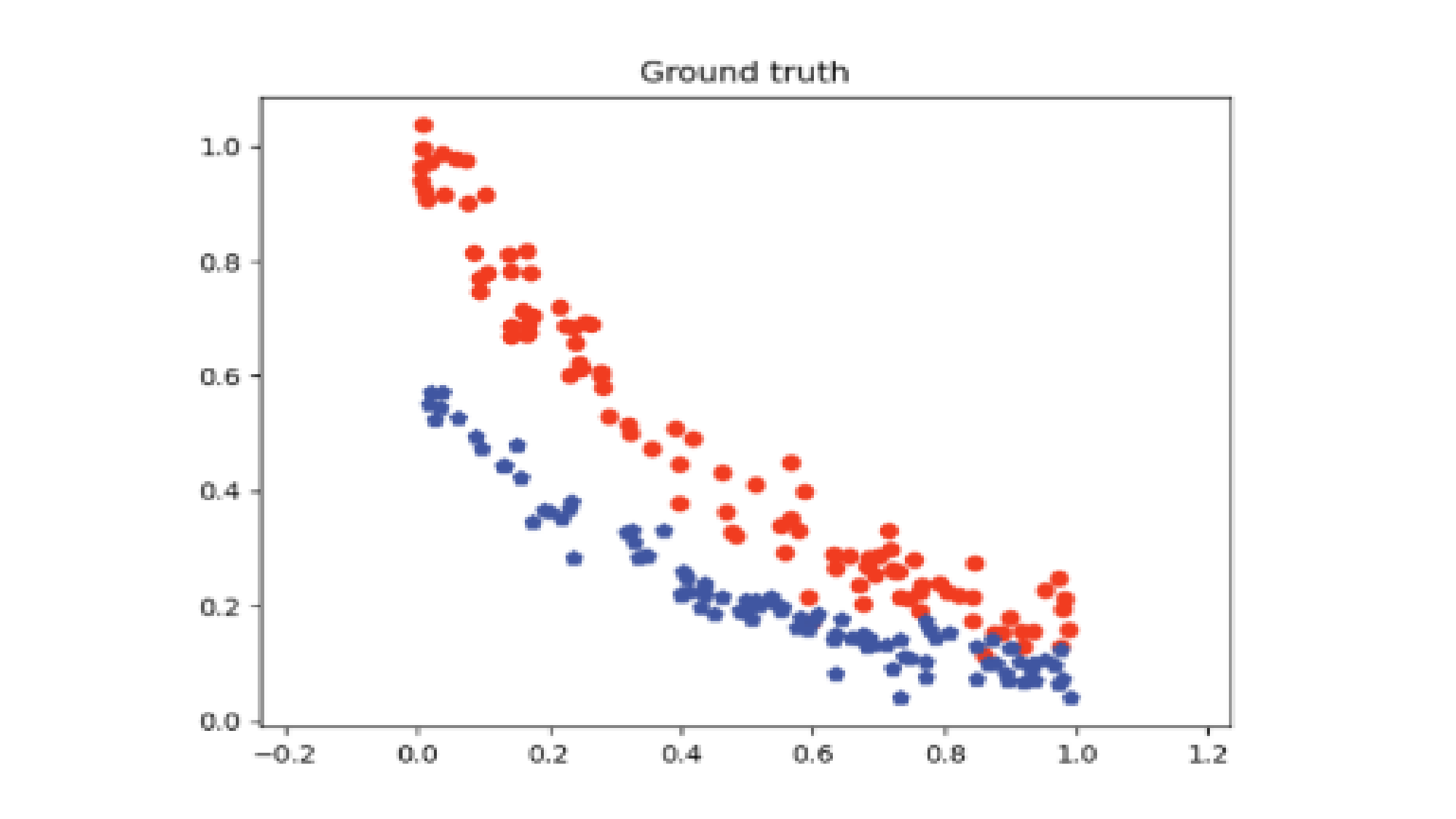}%
\label{fig_groundtruth}}
\subfloat[]{\includegraphics[width=1.8in]{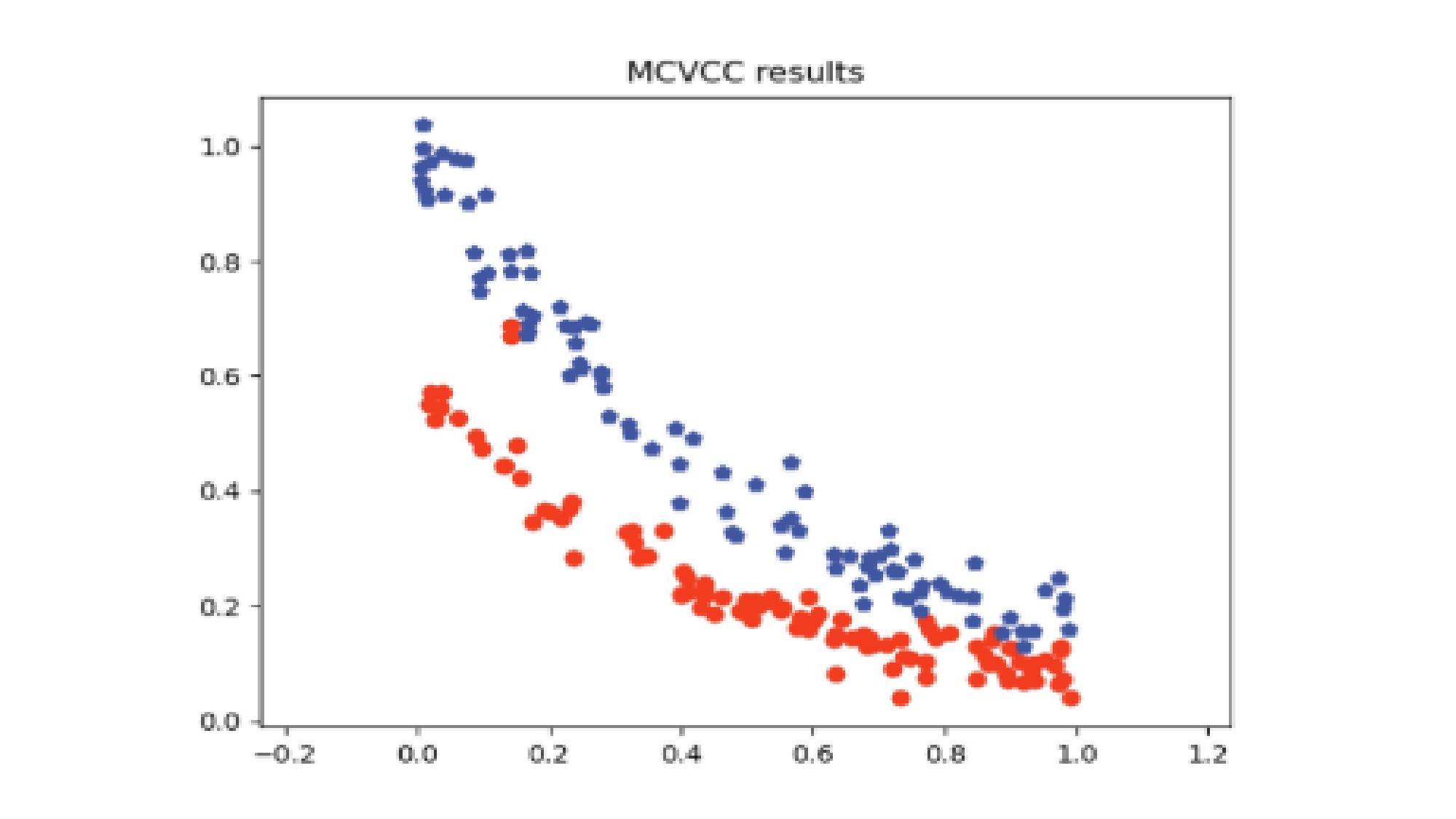}%
\label{fig_mcvcc}}
\subfloat[]{\includegraphics[width=1.8in]{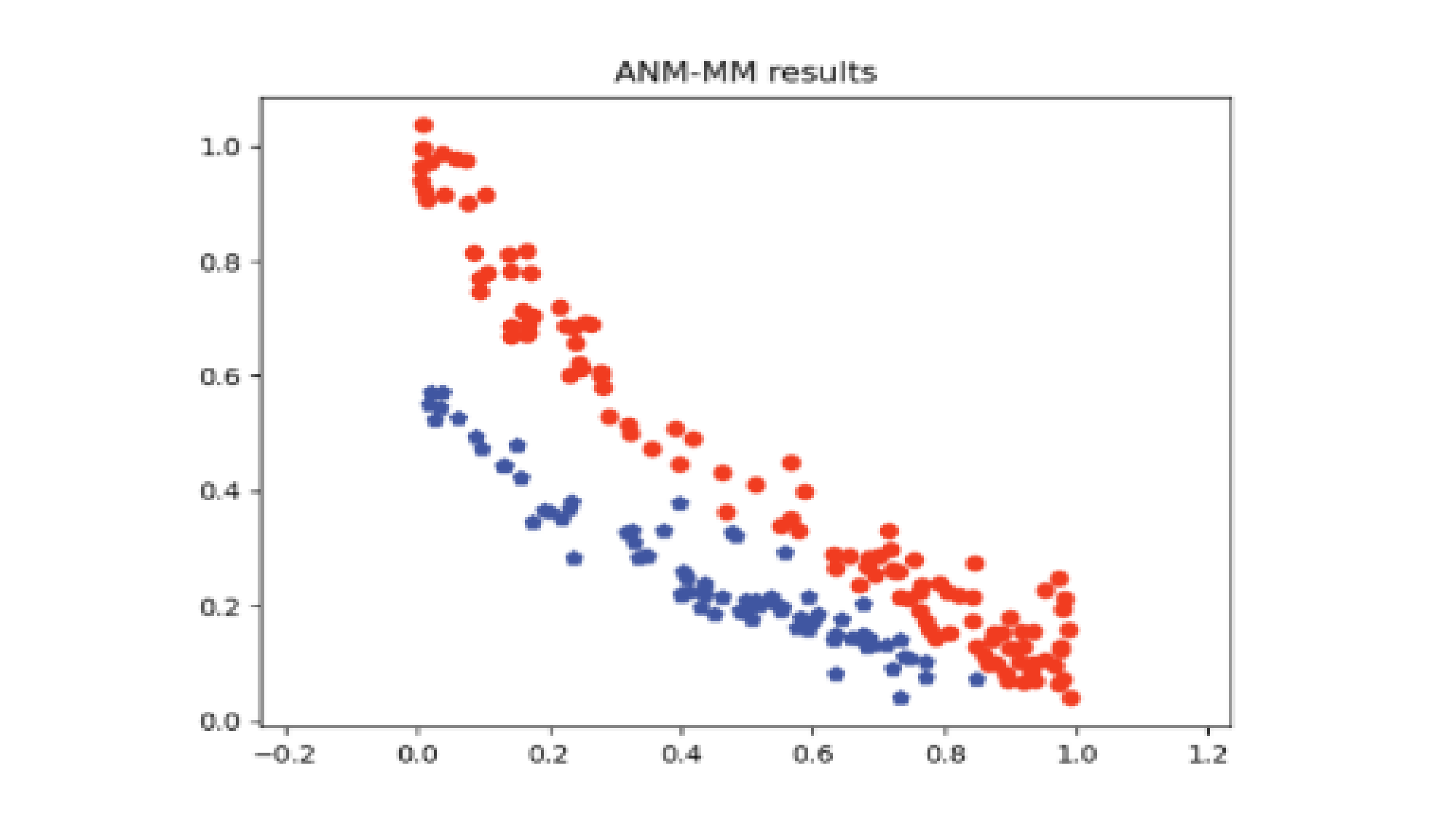}%
\label{fig_anm_mm}}
\subfloat[]{\includegraphics[width=1.8in]{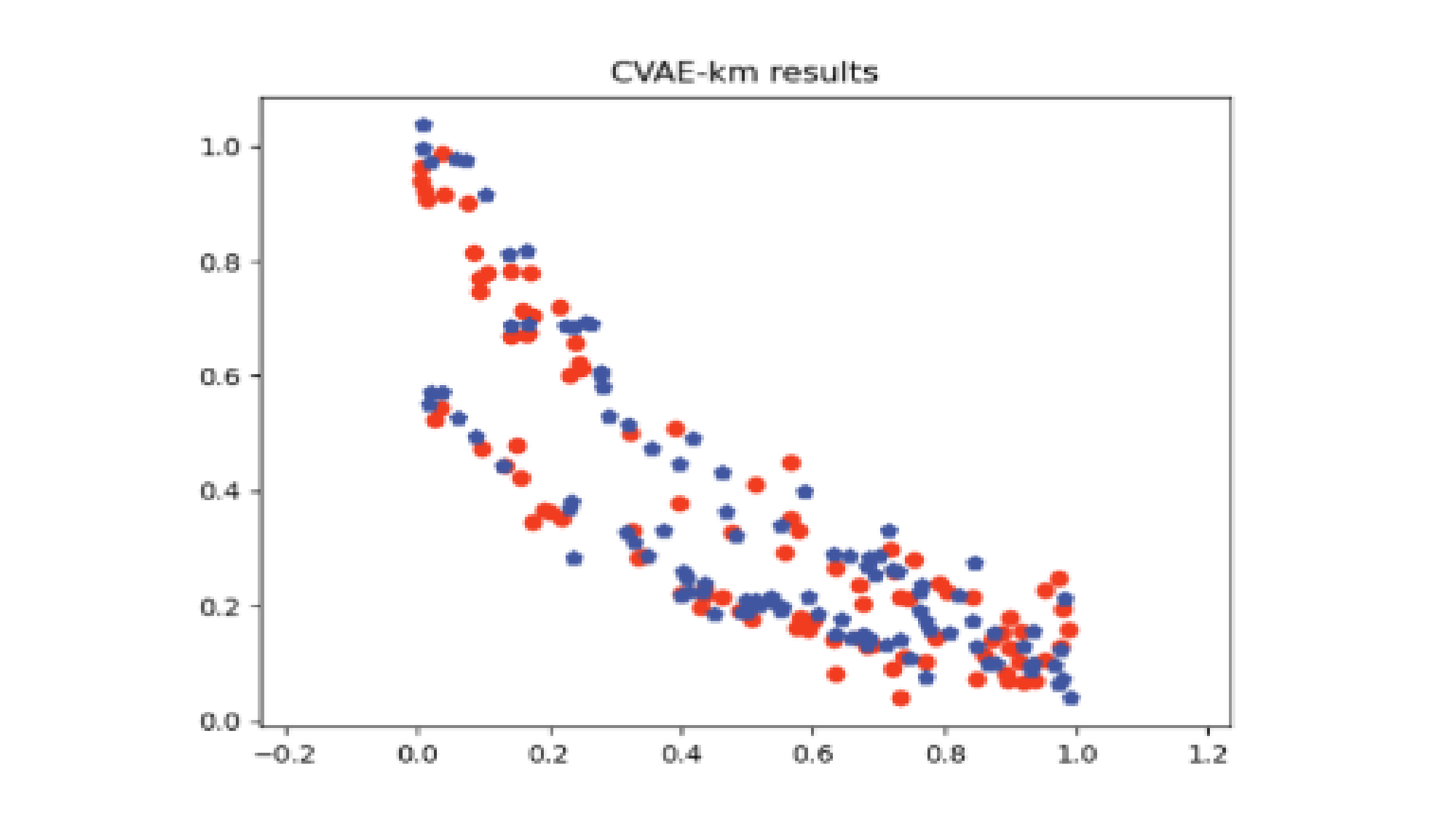}%
\label{fig_cvae_km}}
\hfil
\subfloat[]{\includegraphics[width=1.8in]{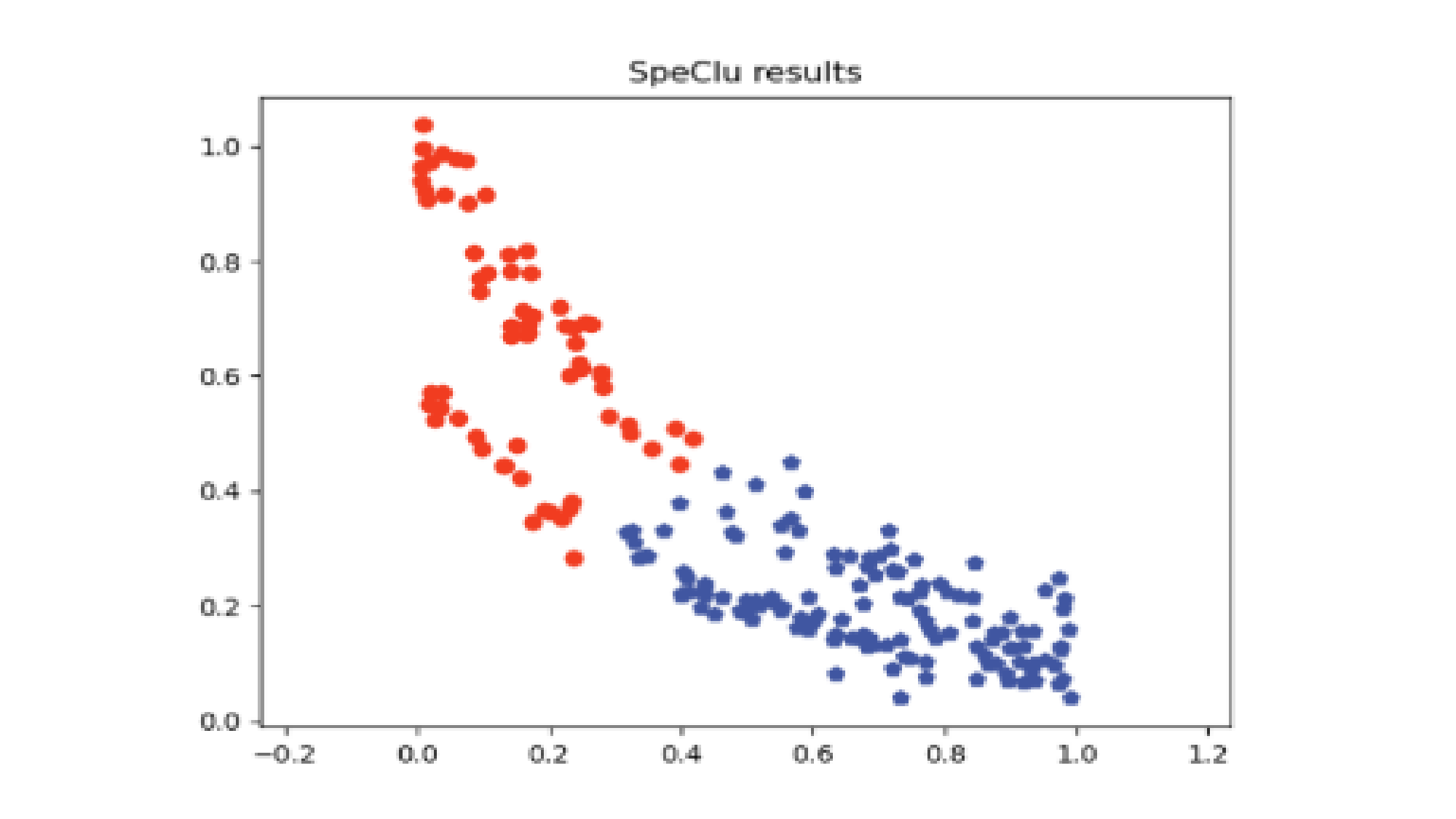}%
\label{fig_speclu}}
\subfloat[]{\includegraphics[width=1.8in]{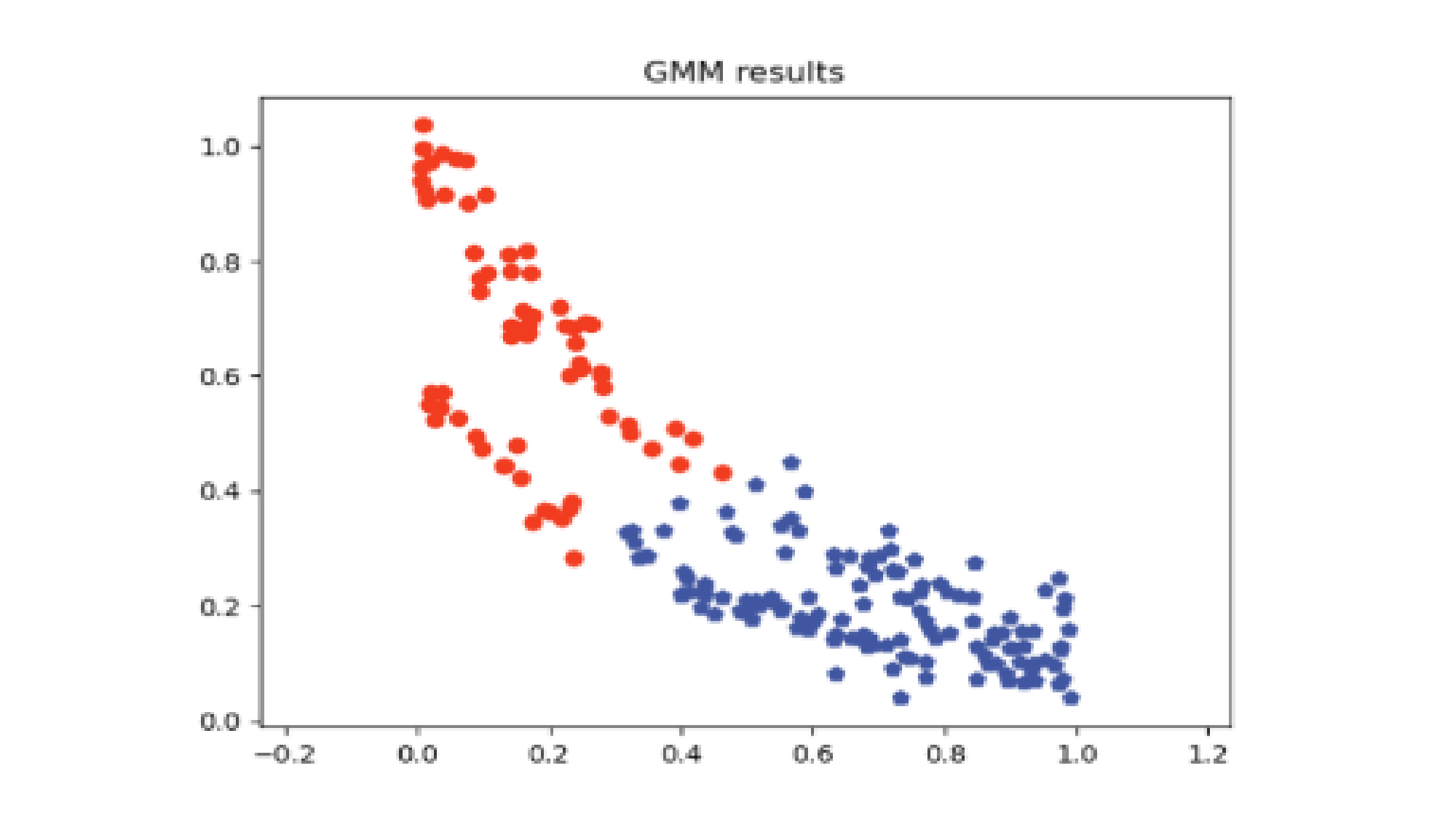}%
\label{fig_gmm}}
\subfloat[]{\includegraphics[width=1.8in]{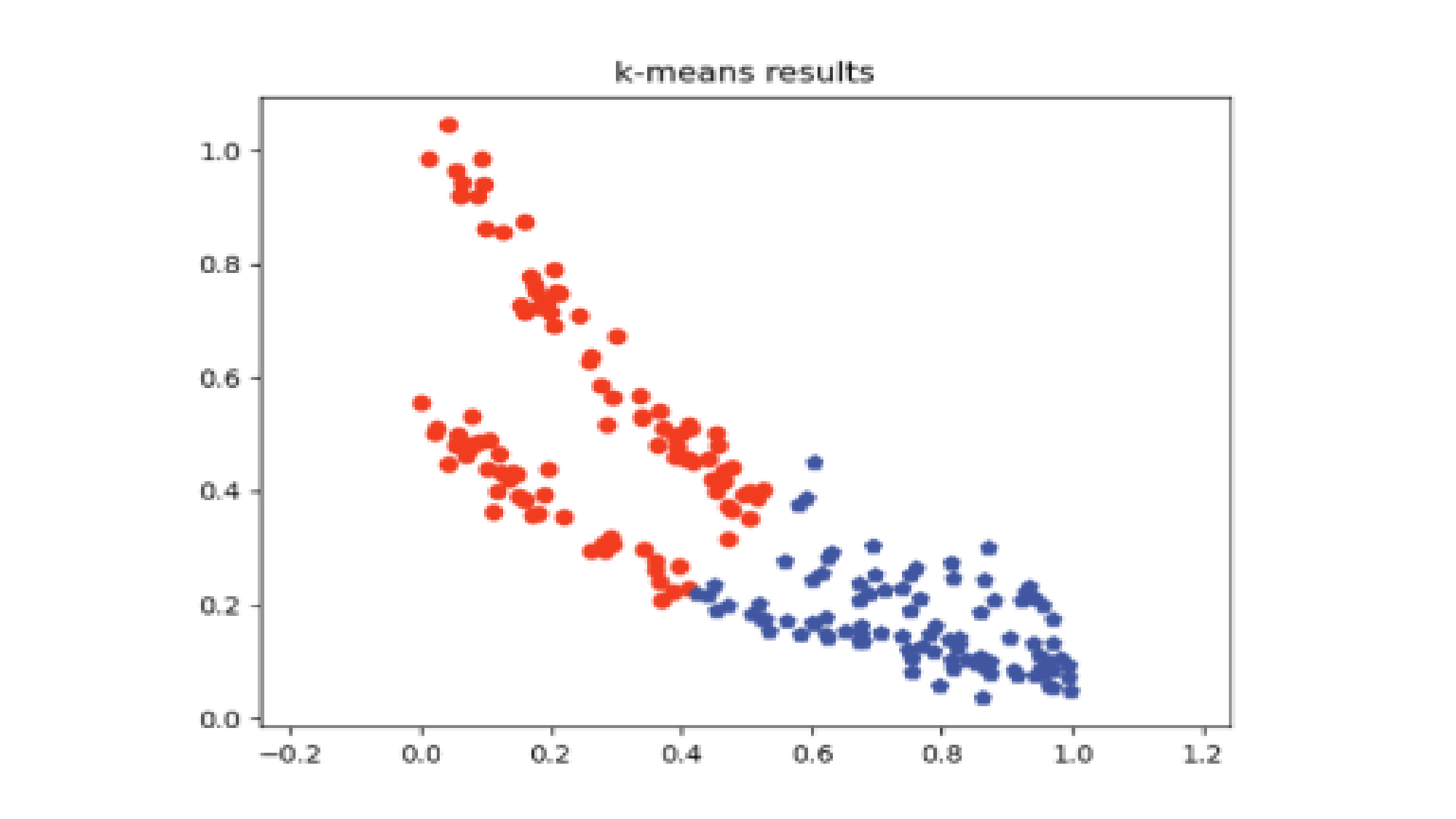}%
\label{fig_kmeans}}
\caption{The clustering results of MCVCC and comparison algorithms when $f=f_2$.}
\label{fig_comparison}
\end{figure*}

\begin{figure*}[!t]
\centering
\subfloat[]{\includegraphics[width=1.8in]{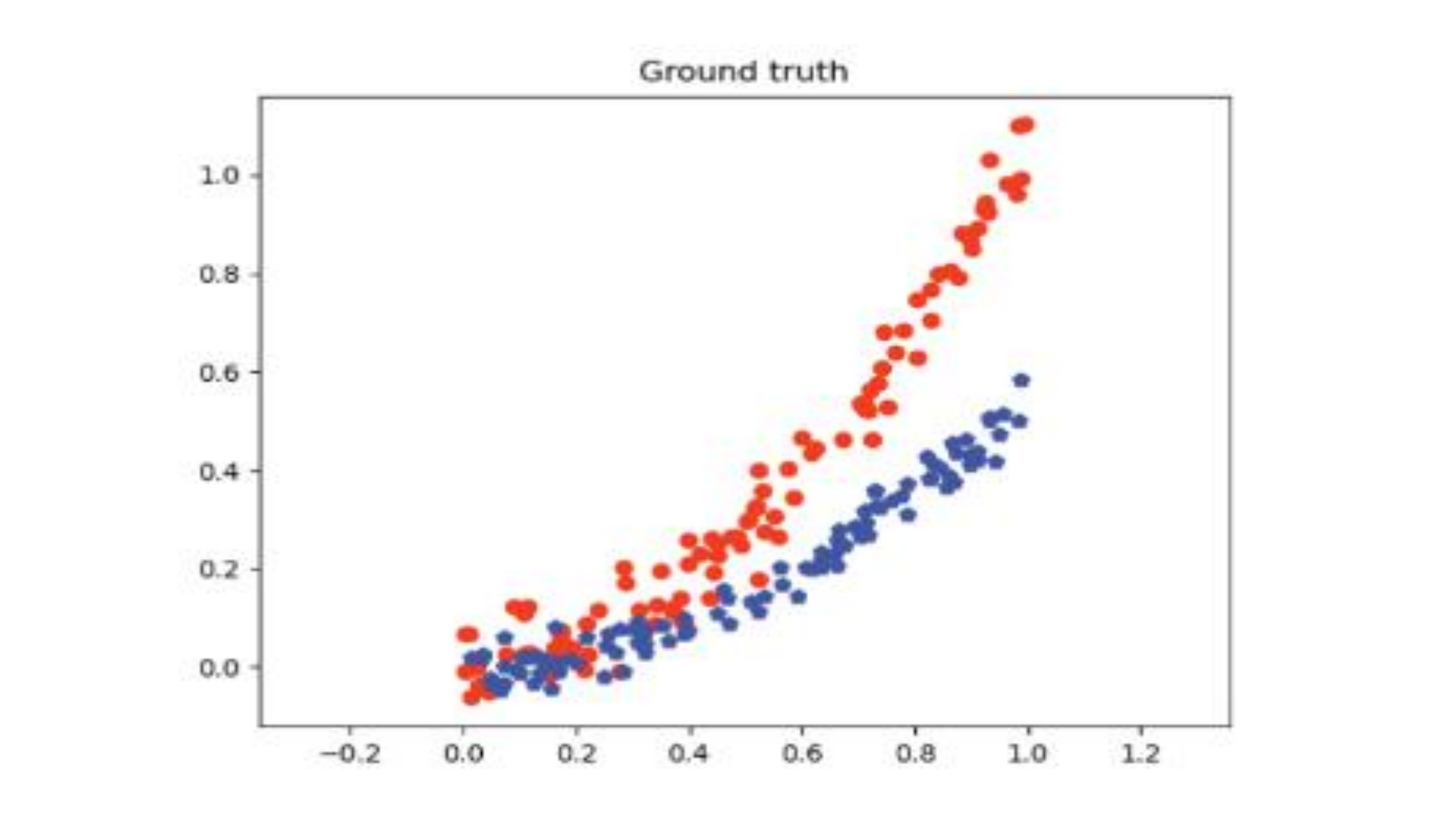}%
\label{fig_groundtruth}}
\subfloat[]{\includegraphics[width=1.8in]{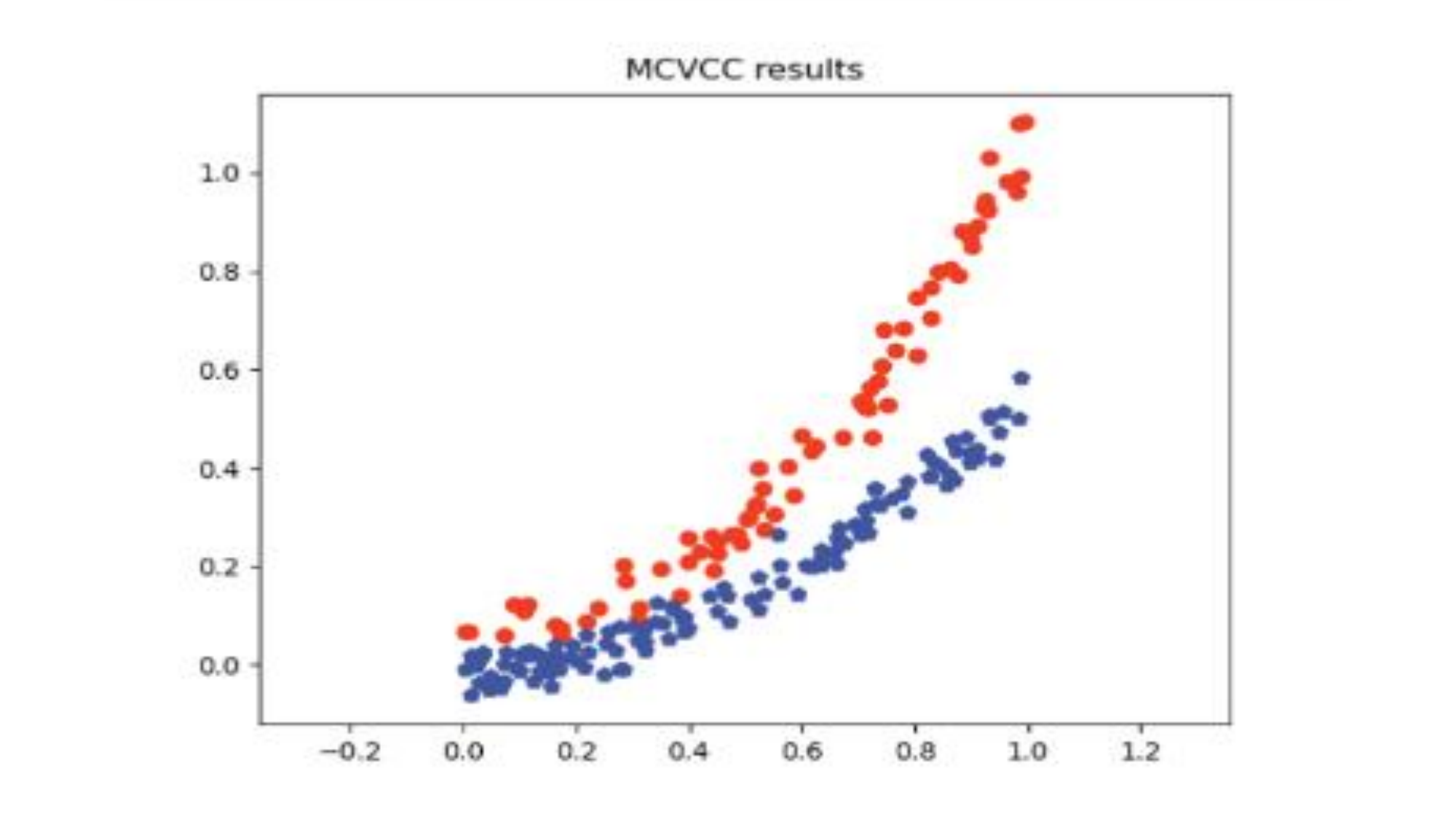}%
\label{fig_mcvcc}}
\subfloat[]{\includegraphics[width=1.8in]{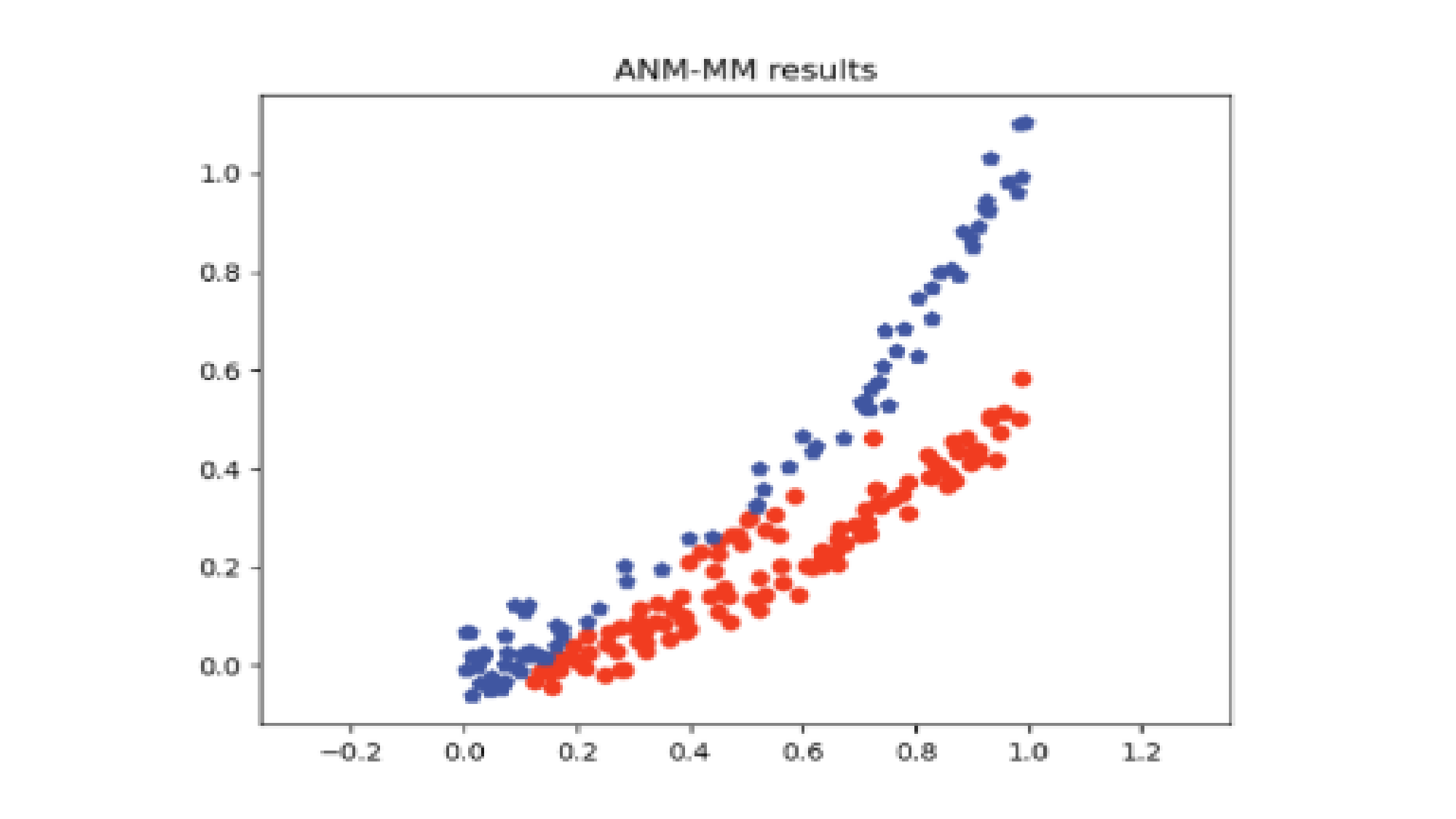}%
\label{fig_anm_mm}}
\subfloat[]{\includegraphics[width=1.8in]{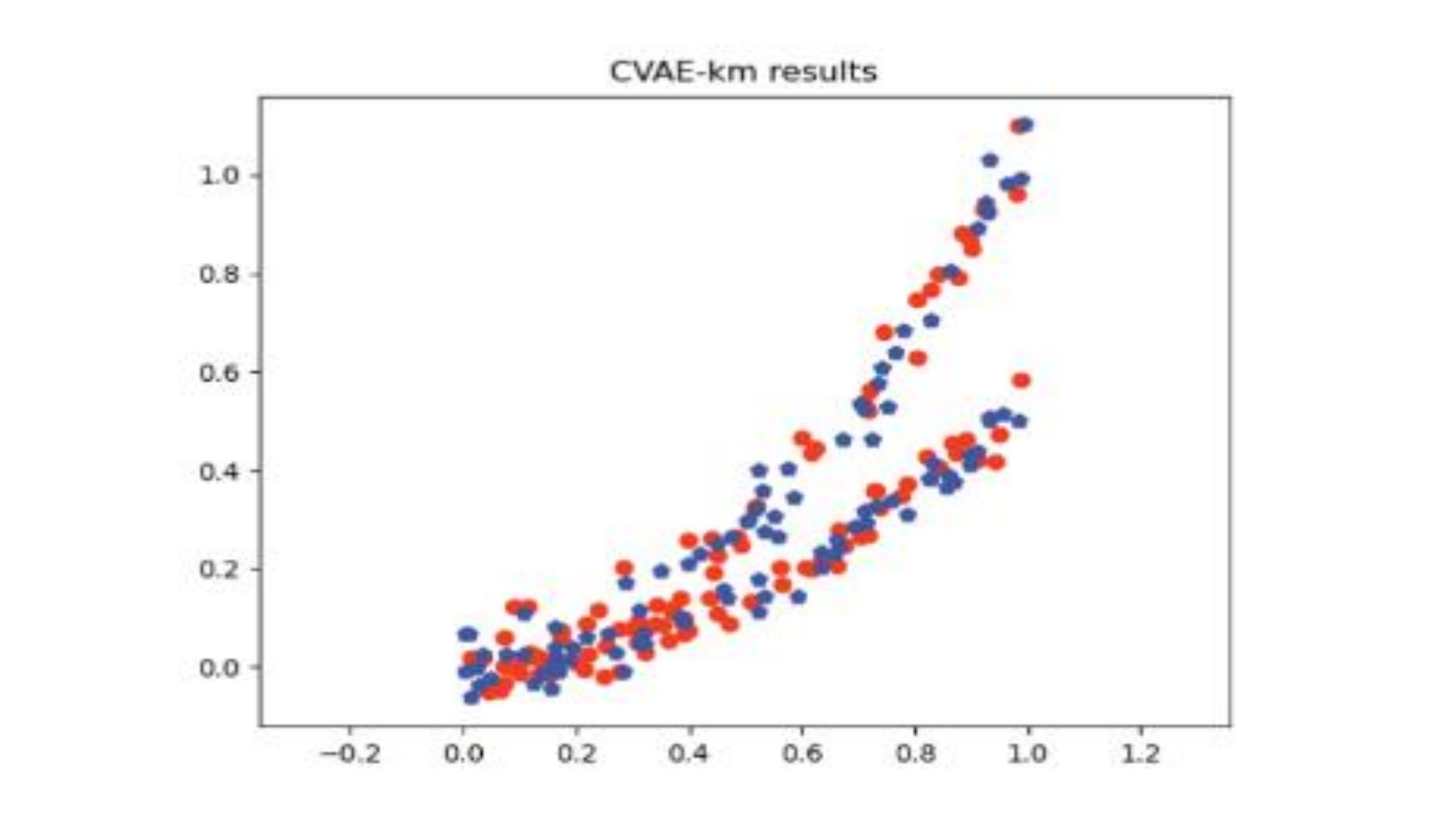}%
\label{fig_cvae_km}}
\hfil
\subfloat[]{\includegraphics[width=1.8in]{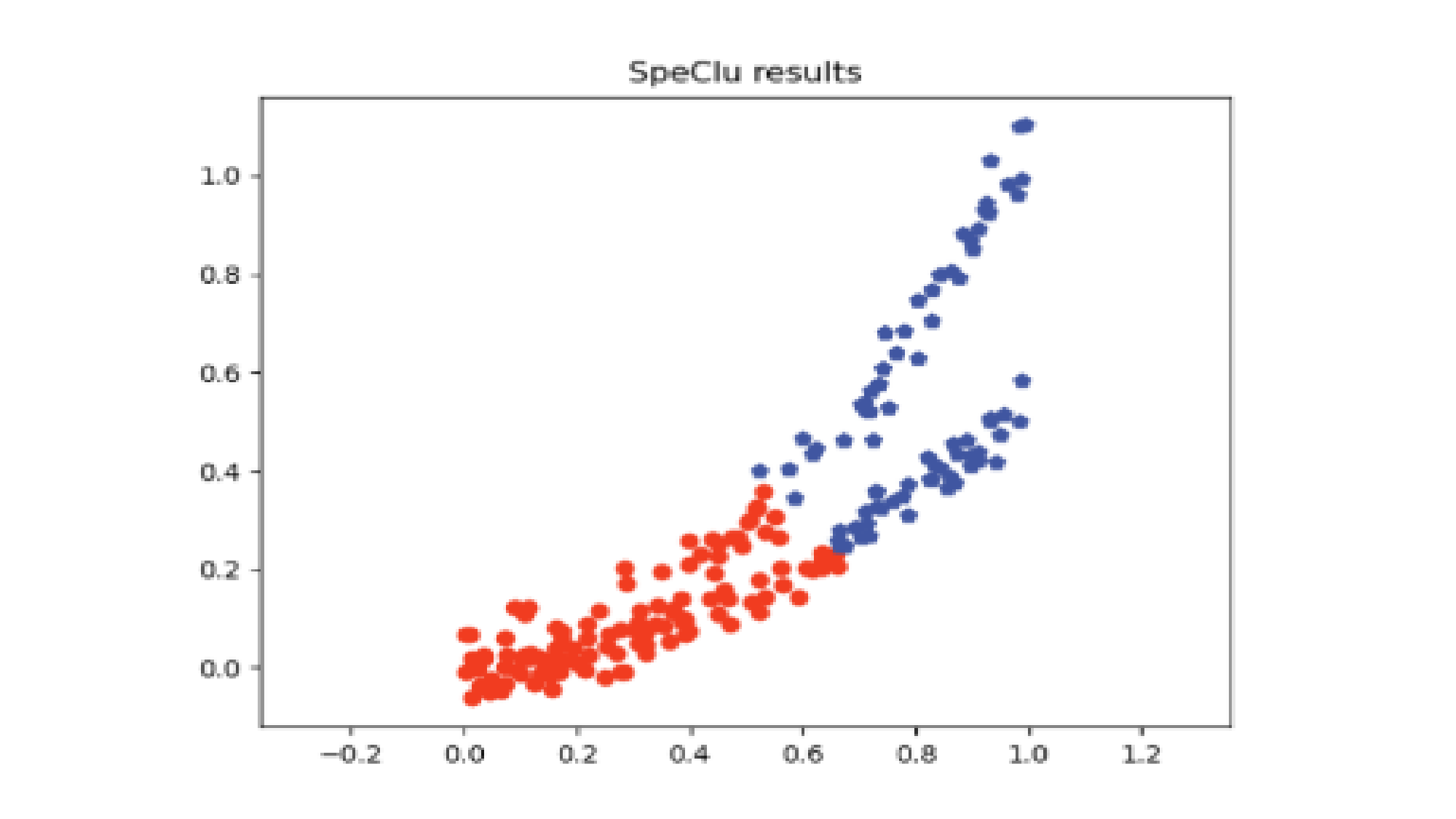}%
\label{fig_speclu}}
\subfloat[]{\includegraphics[width=1.8in]{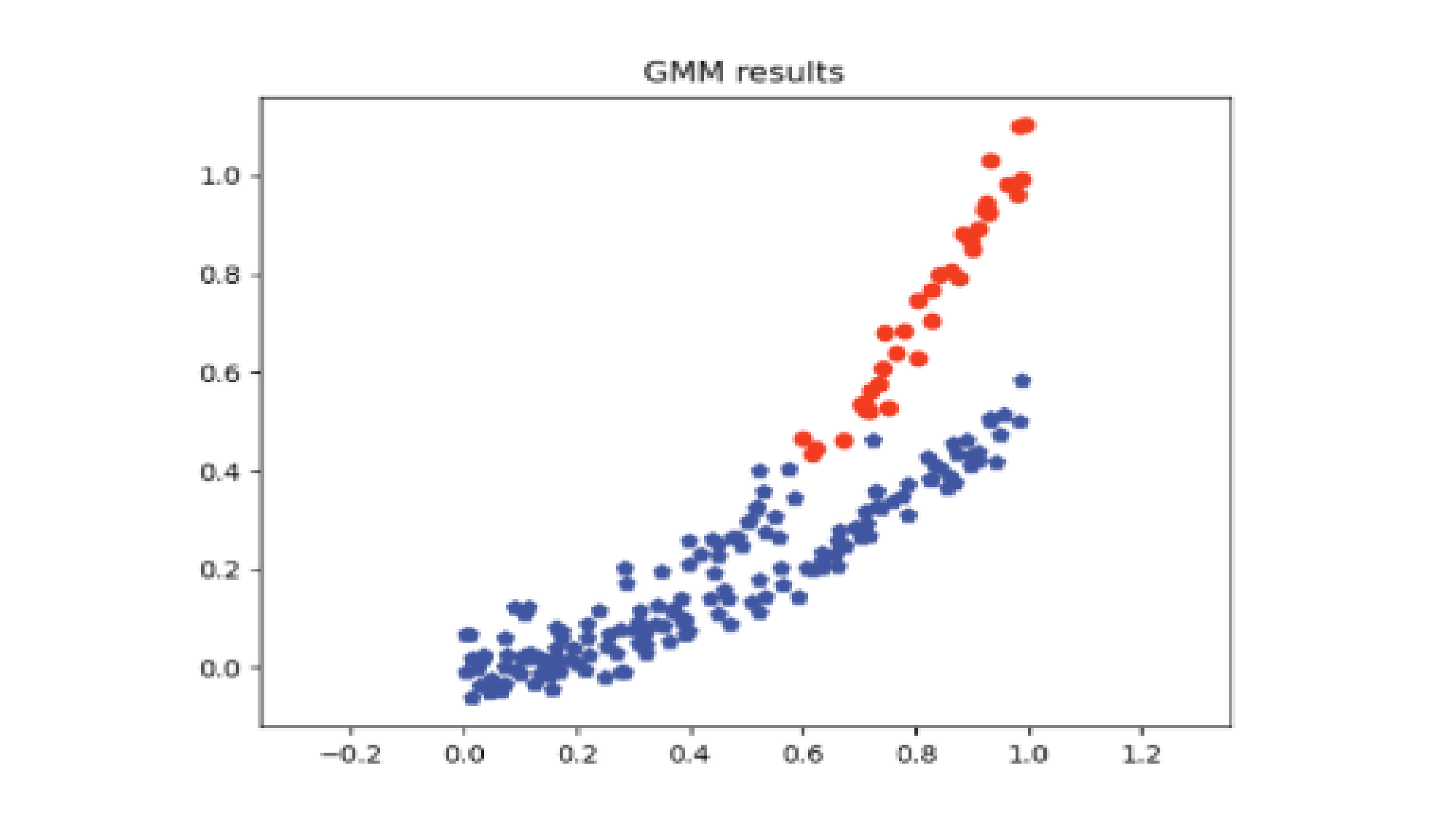}%
\label{fig_gmm}}
\subfloat[]{\includegraphics[width=1.8in]{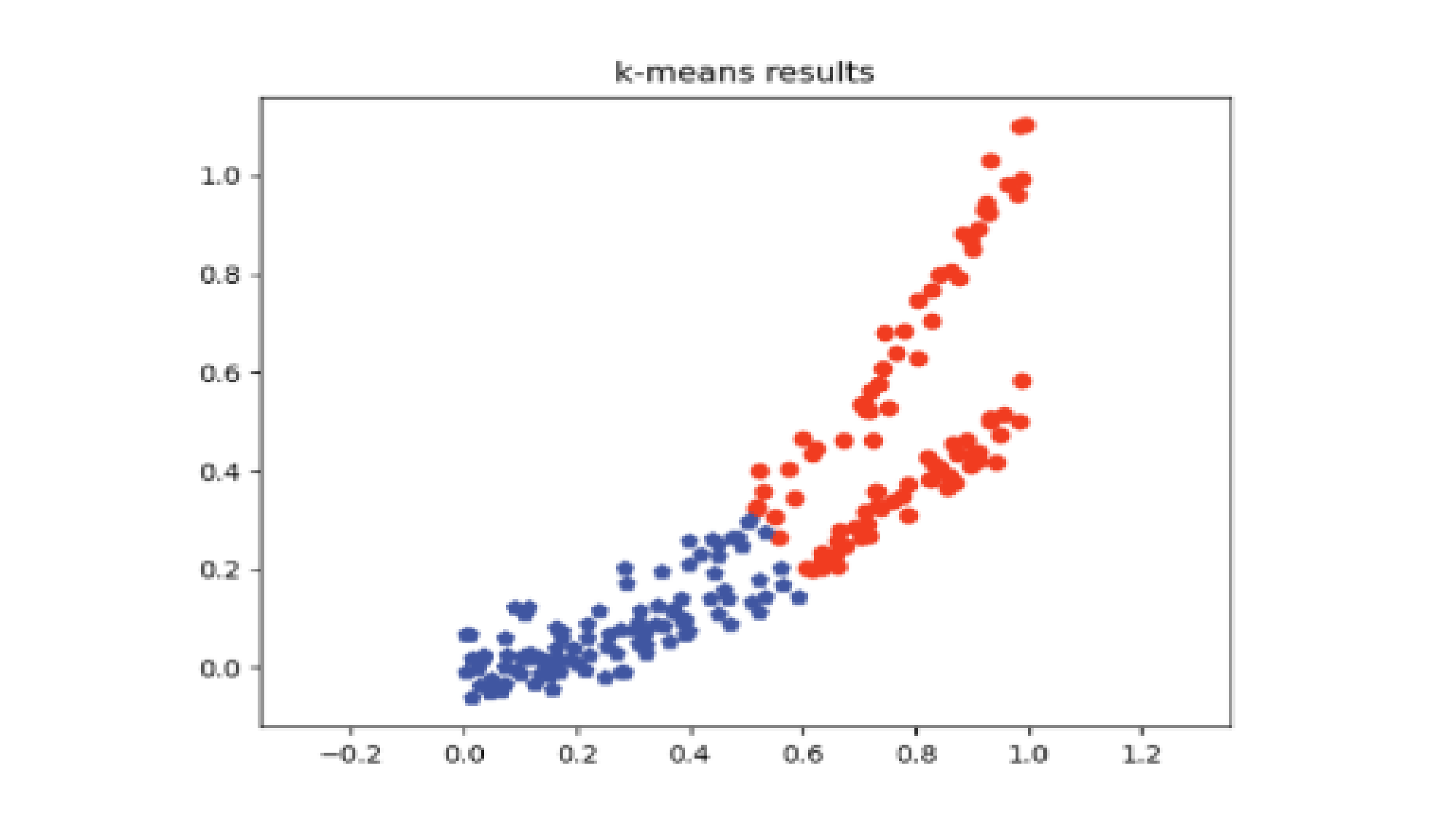}%
\label{fig_kmeans}}
\caption{The clustering results of MCVCC and comparison algorithms when $f=f_3$.}
\label{fig_comparison}
\end{figure*}

\begin{figure*}[!t]
\centering
\subfloat[]{\includegraphics[width=1.8in]{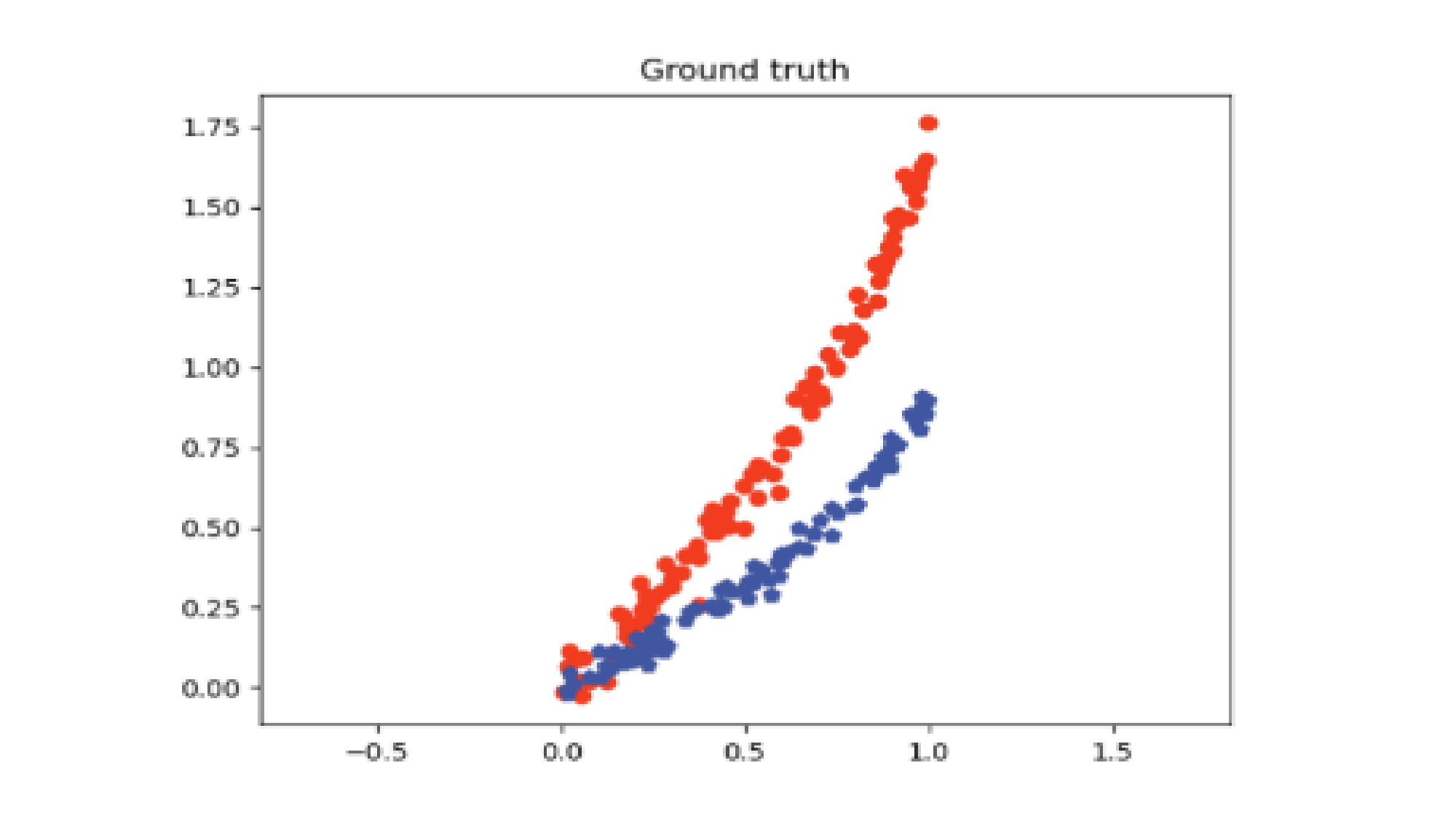}%
\label{fig_groundtruth}}
\subfloat[]{\includegraphics[width=1.8in]{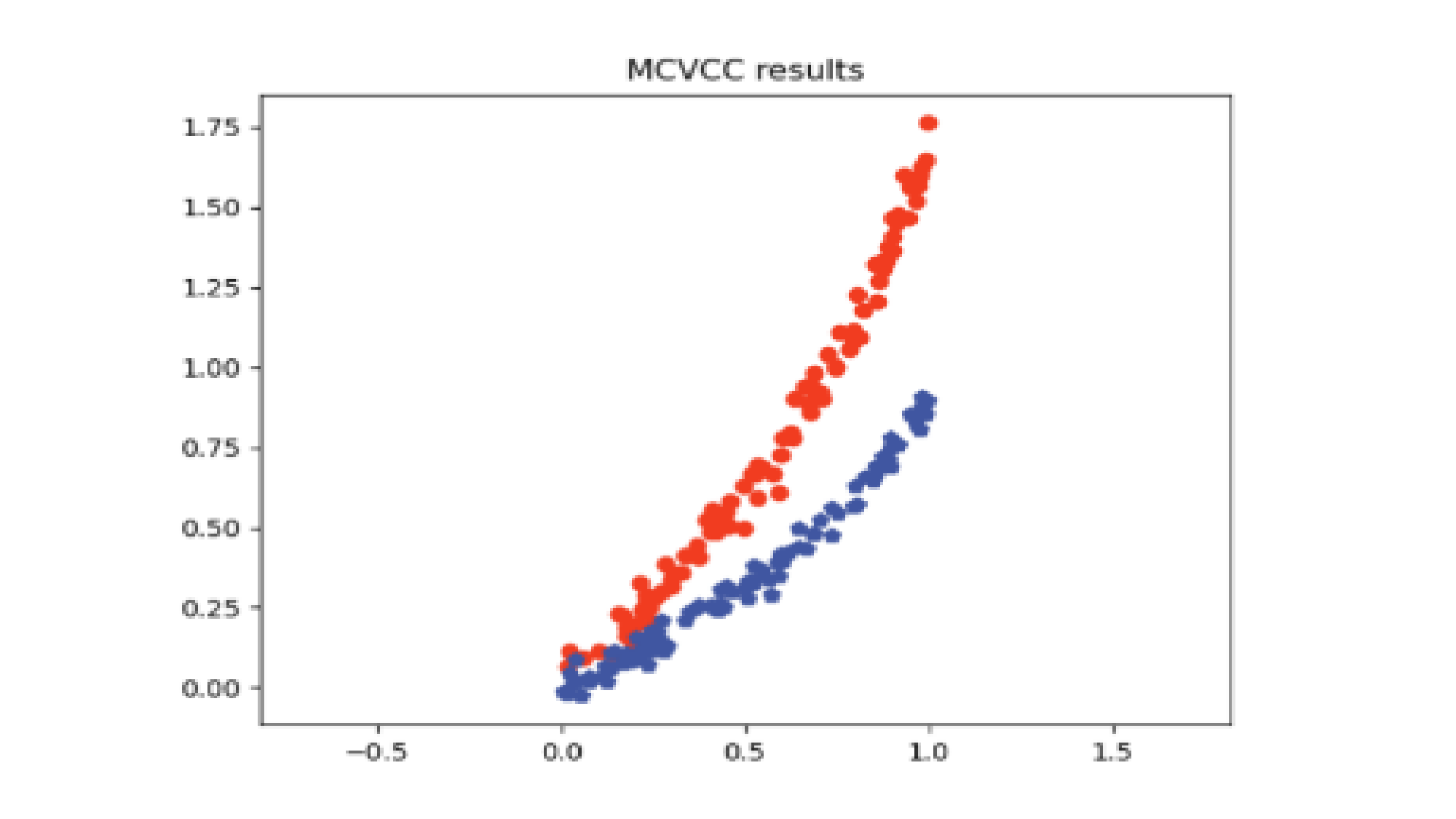}%
\label{fig_mcvcc}}
\subfloat[]{\includegraphics[width=1.8in]{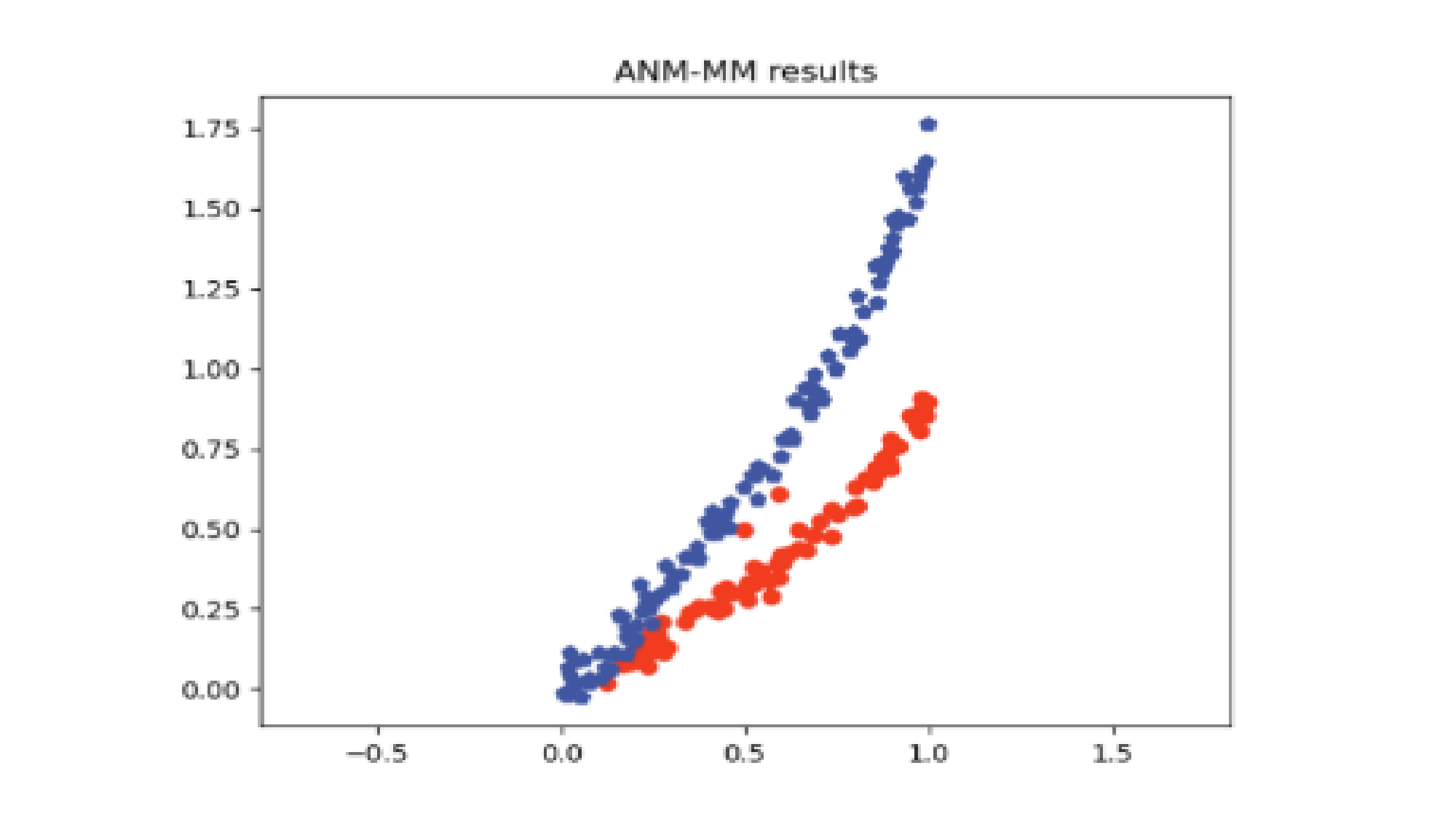}%
\label{fig_anm_mm}}
\subfloat[]{\includegraphics[width=1.8in]{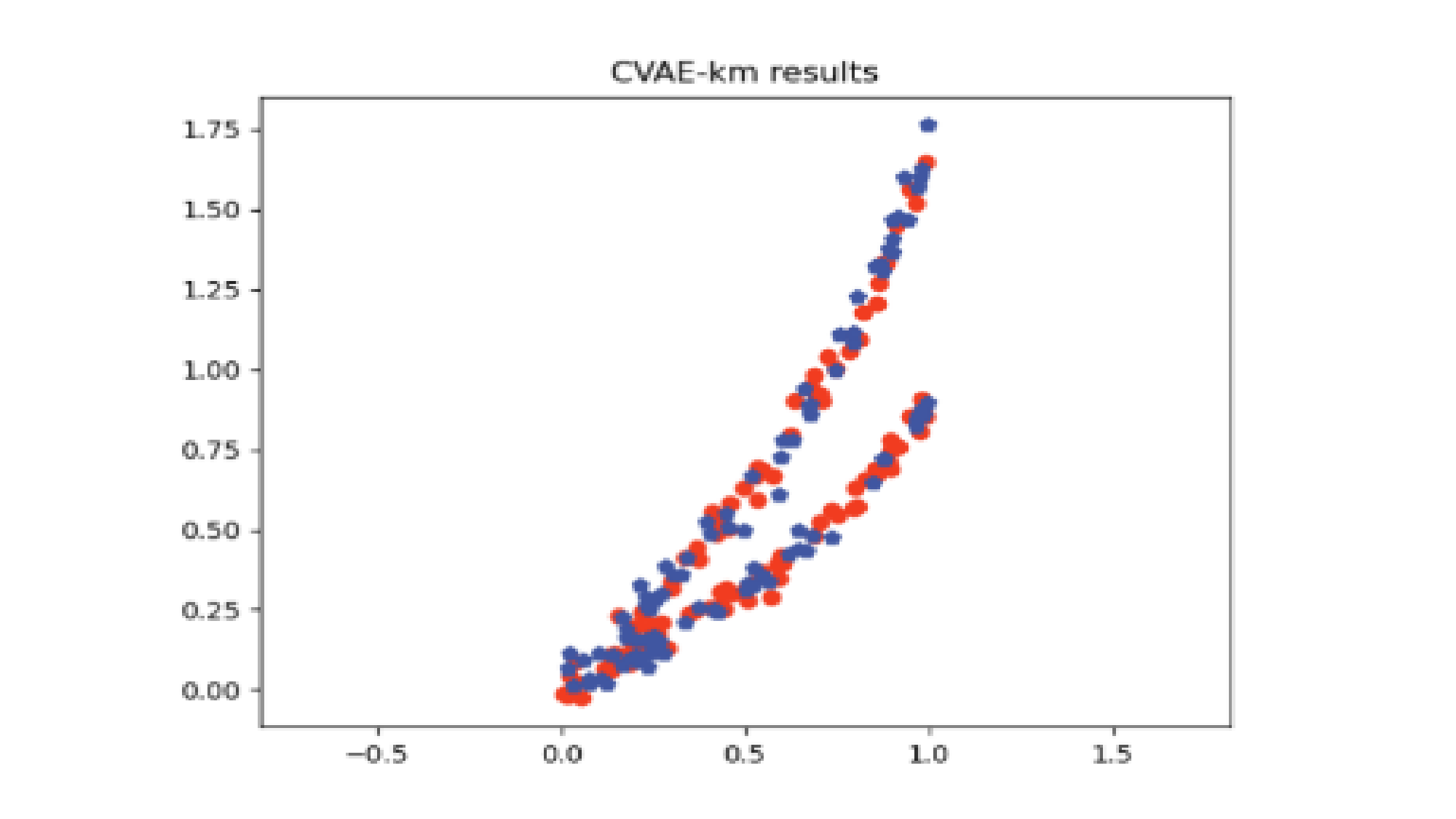}%
\label{fig_cvae_km}}
\hfil
\subfloat[]{\includegraphics[width=1.8in]{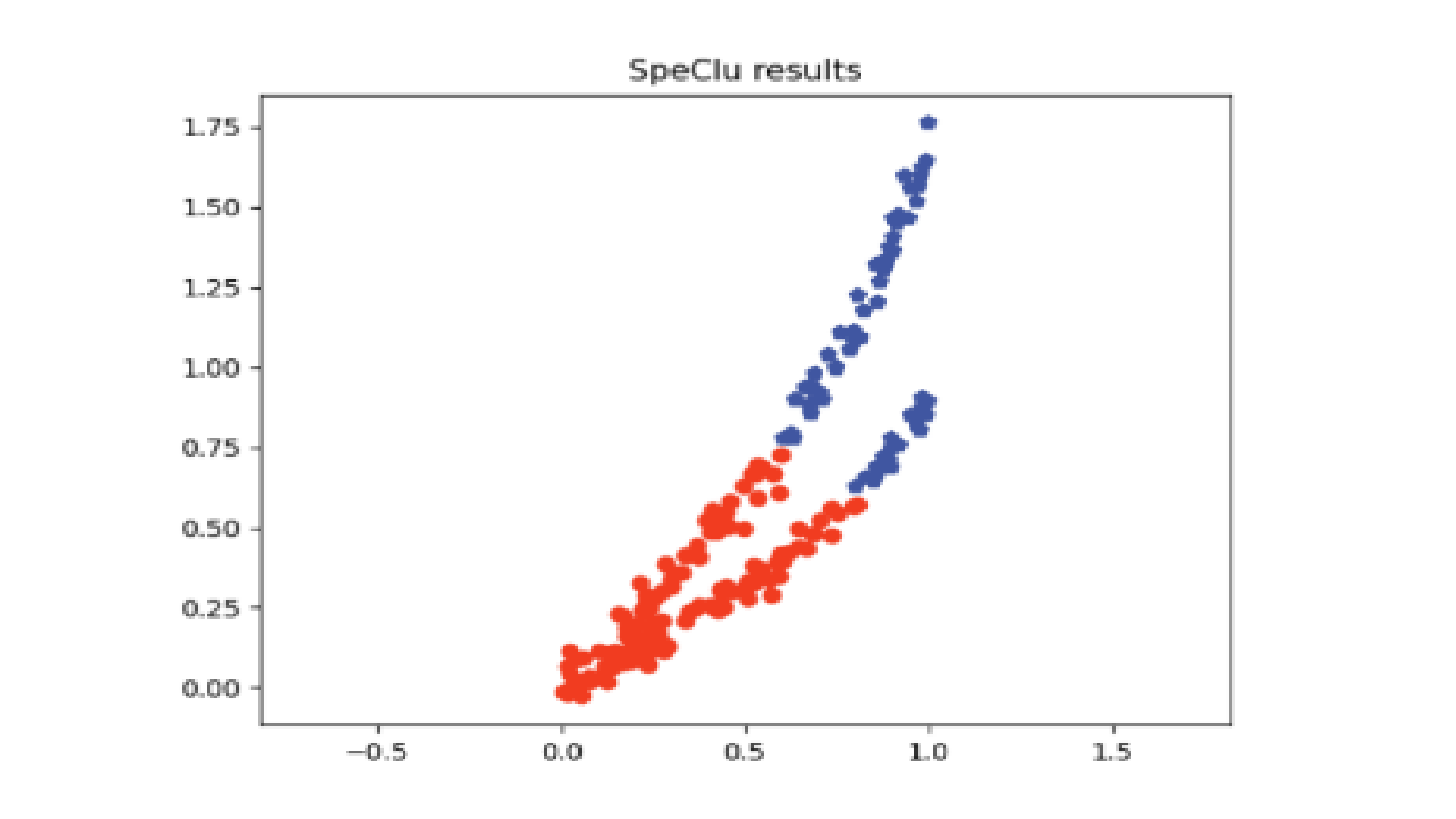}%
\label{fig_speclu}}
\subfloat[]{\includegraphics[width=1.8in]{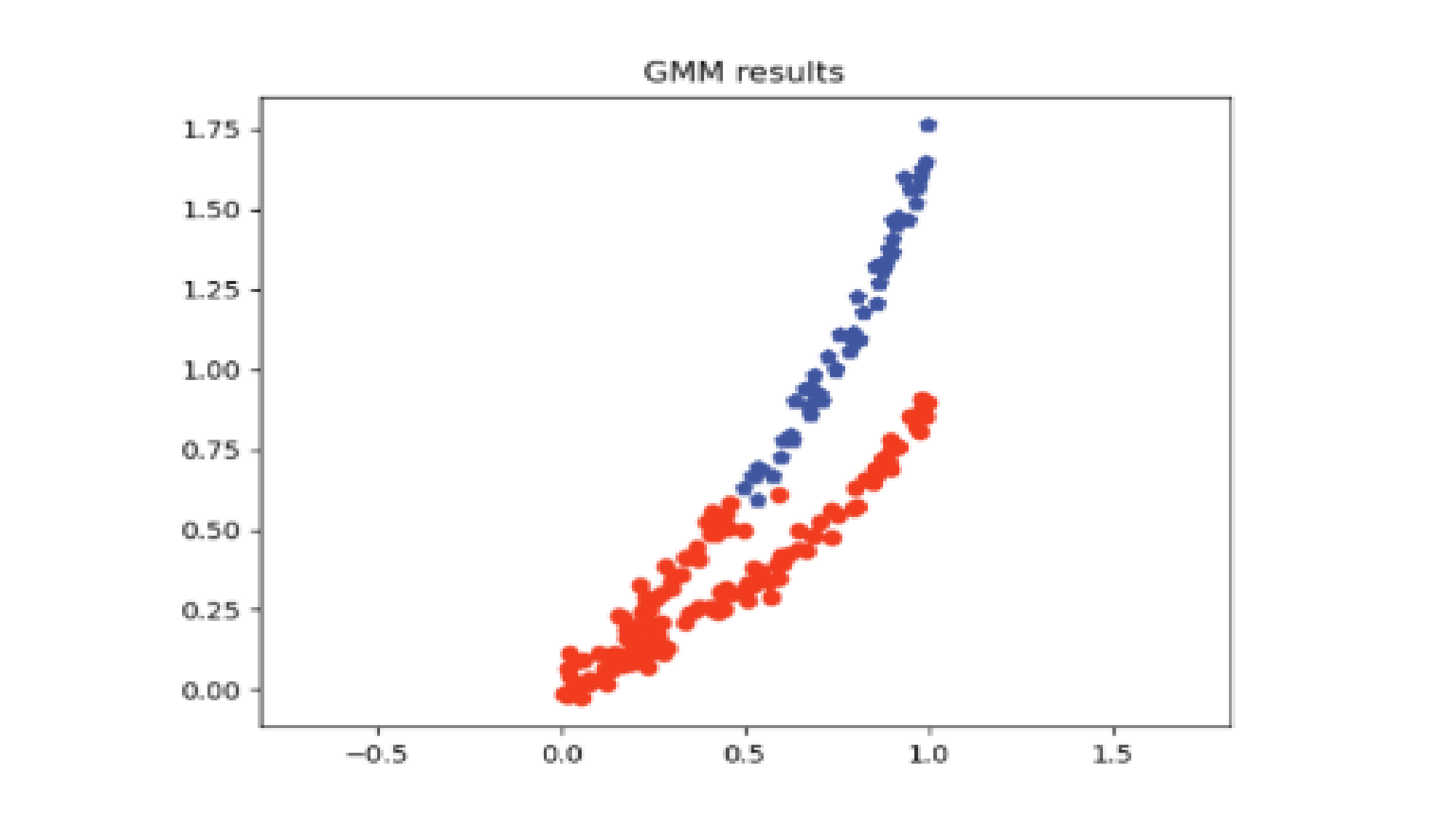}%
\label{fig_gmm}}
\subfloat[]{\includegraphics[width=1.8in]{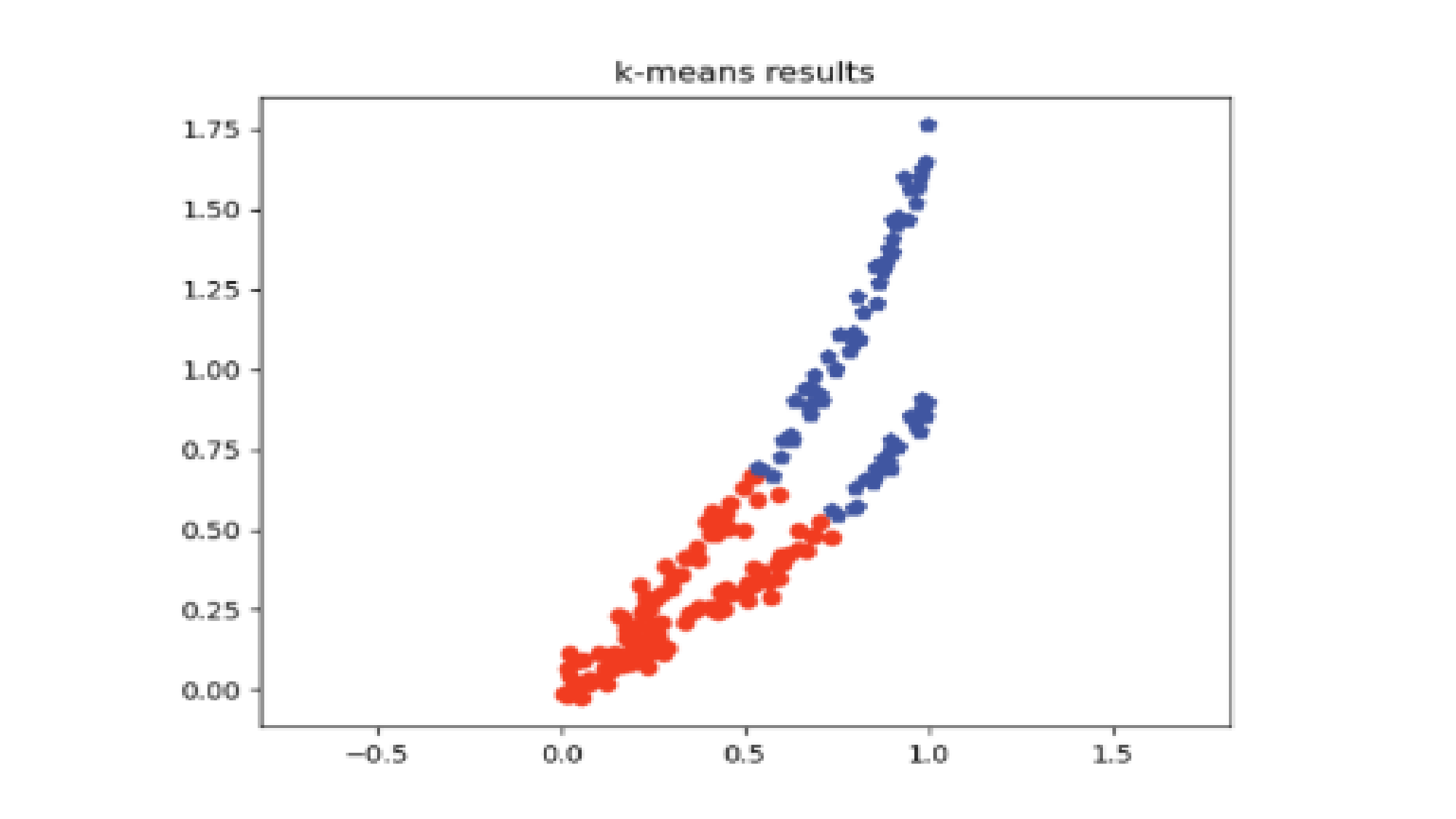}%
\label{fig_kmeans}}
\caption{The clustering results of MCVCC and comparison algorithms when $f=f_4$.}
\label{fig_comparison}
\end{figure*}

\begin{figure*}[!t]
\centering
\subfloat[]{\includegraphics[width=1.8in]{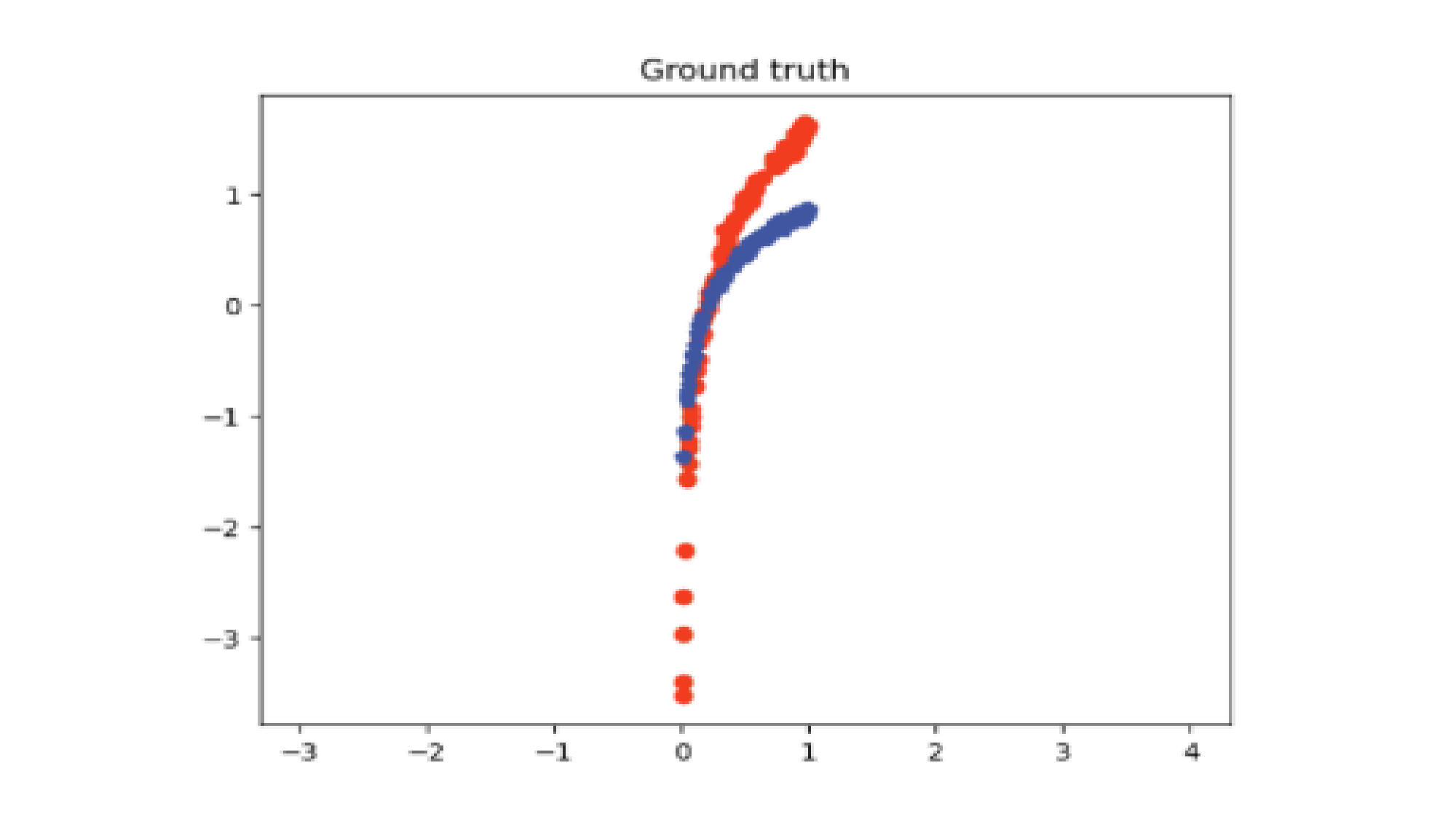}%
\label{fig_groundtruth}}
\subfloat[]{\includegraphics[width=1.8in]{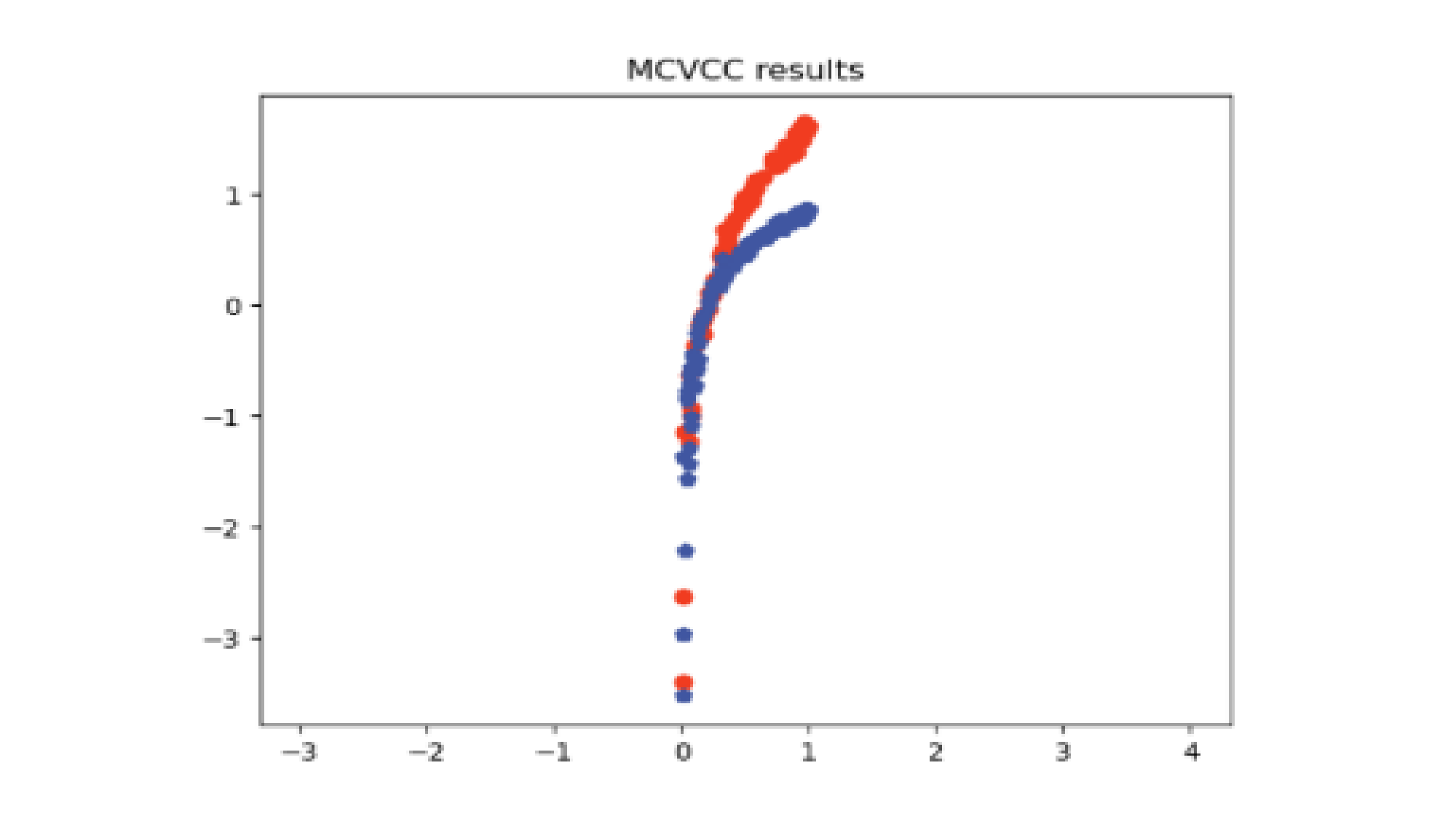}%
\label{fig_mcvcc}}
\subfloat[]{\includegraphics[width=1.8in]{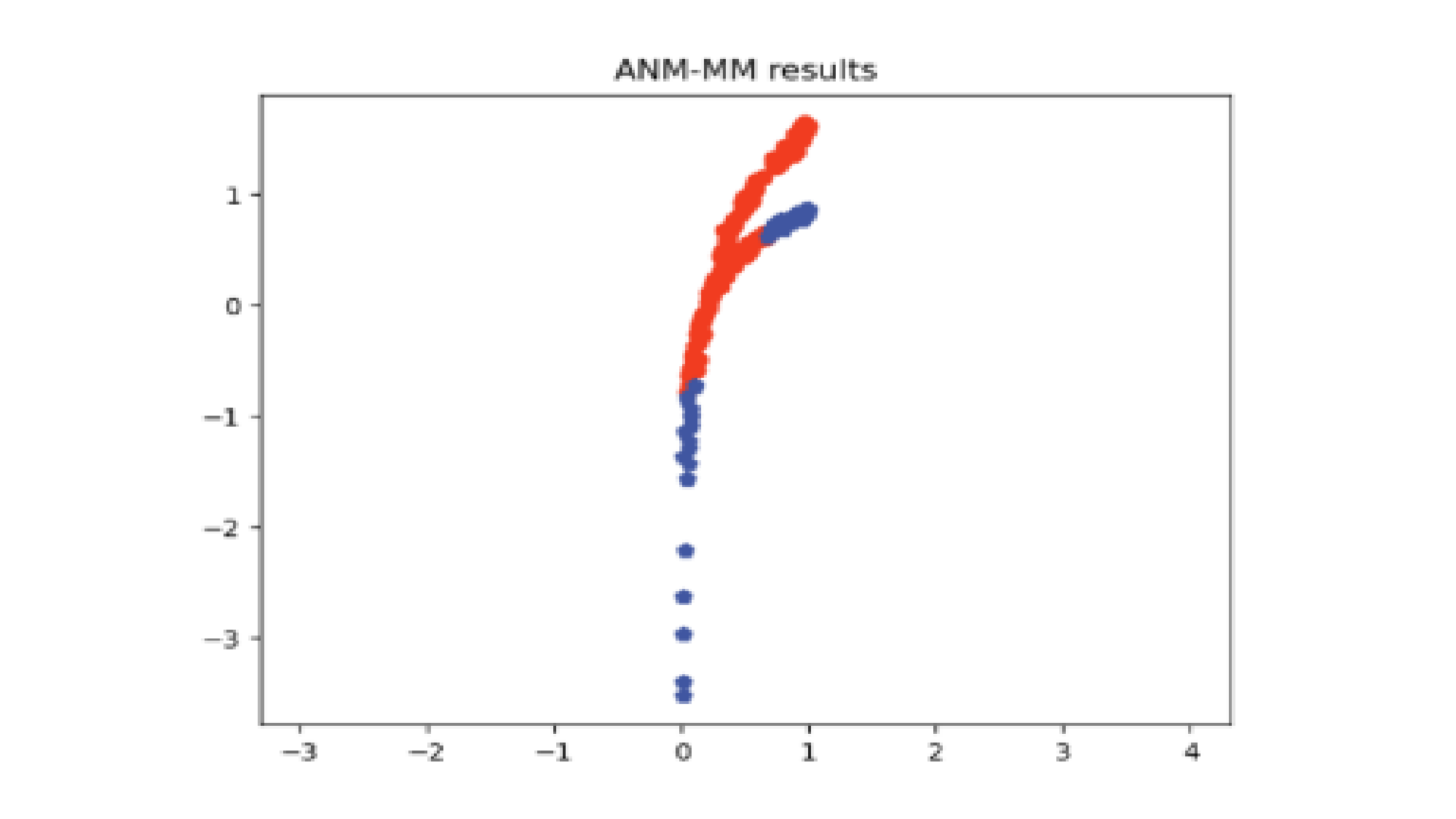}%
\label{fig_anm_mm}}
\subfloat[]{\includegraphics[width=1.8in]{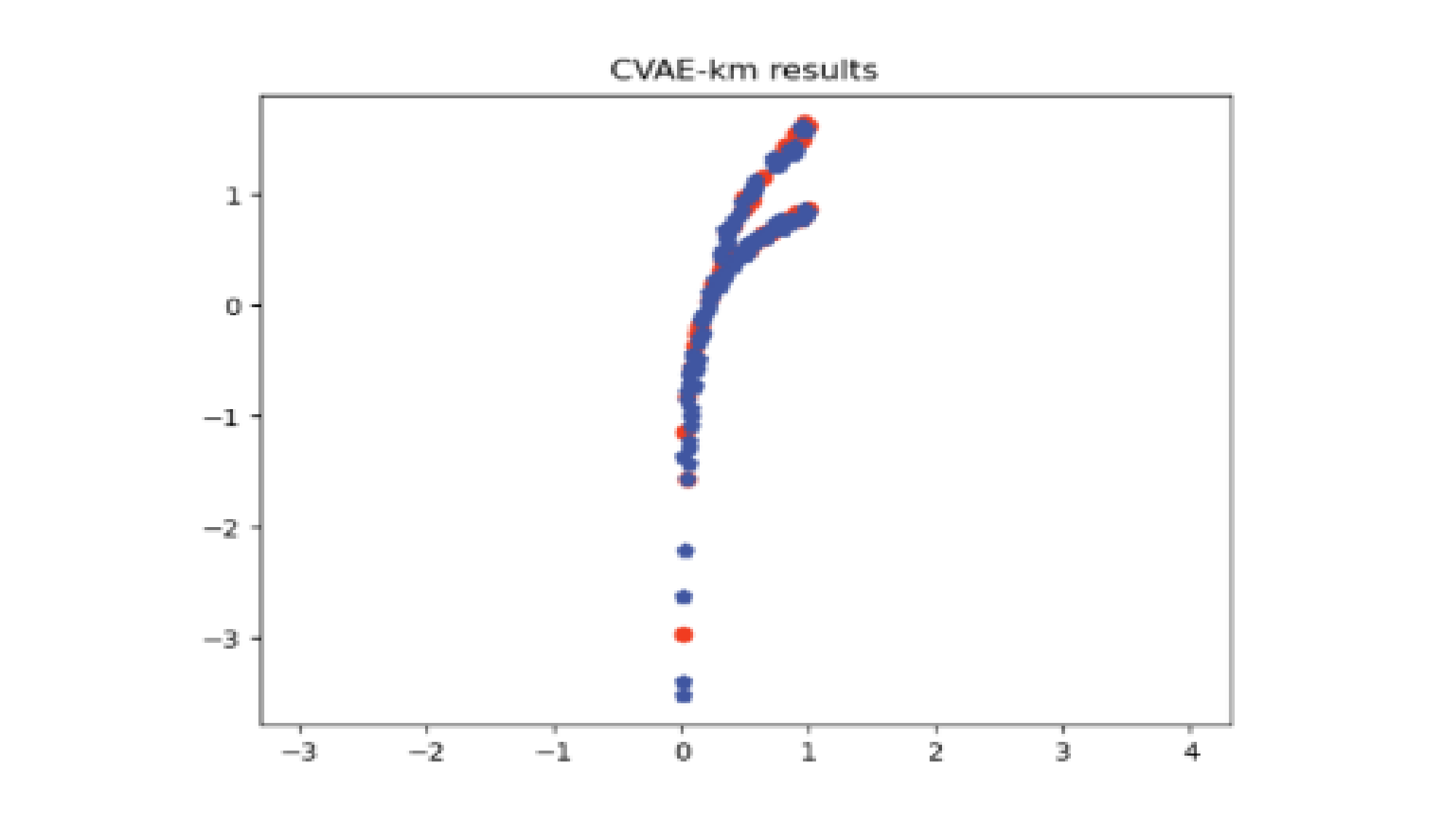}%
\label{fig_cvae_km}}
\hfil
\subfloat[]{\includegraphics[width=1.8in]{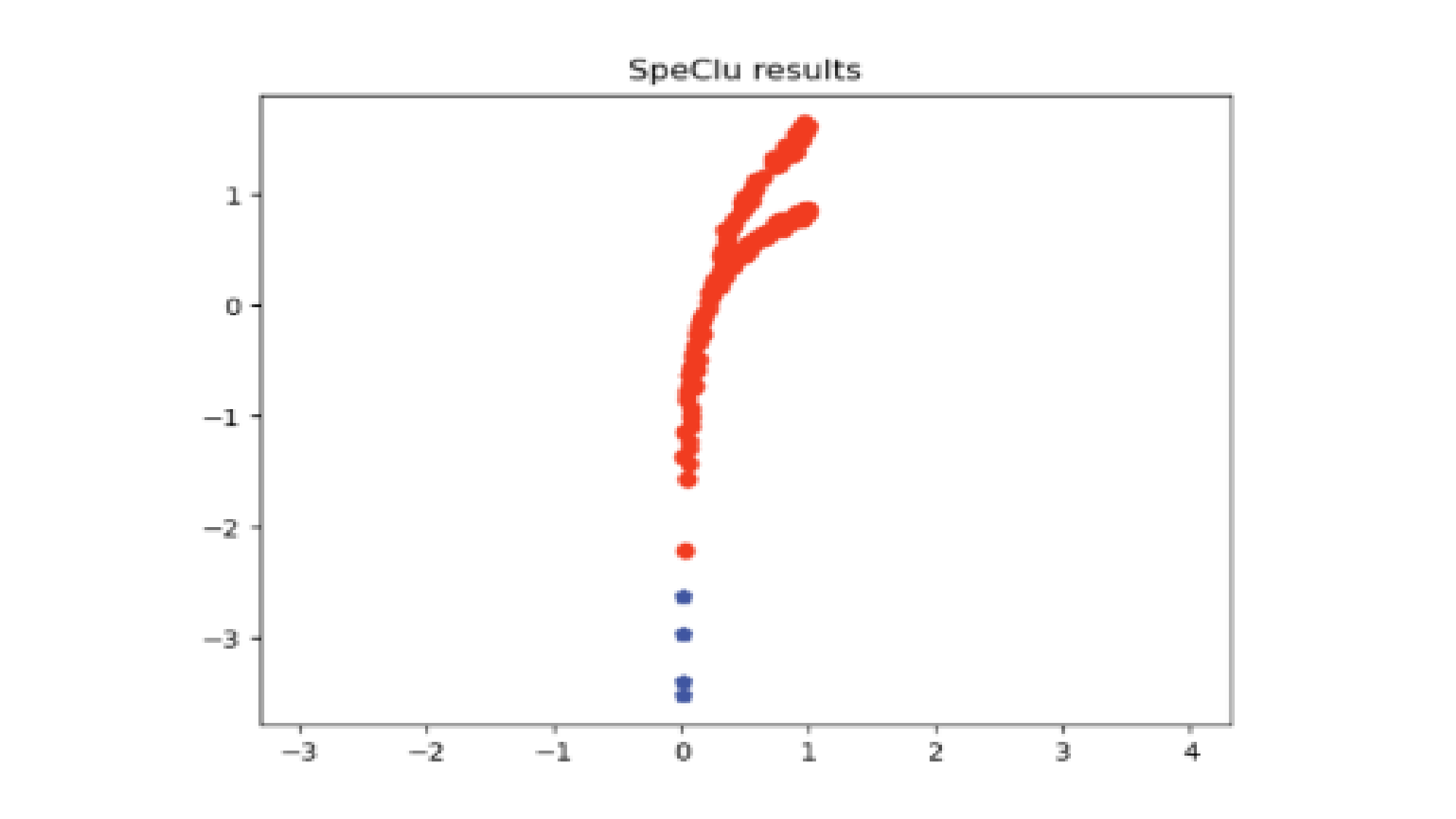}%
\label{fig_speclu}}
\subfloat[]{\includegraphics[width=1.8in]{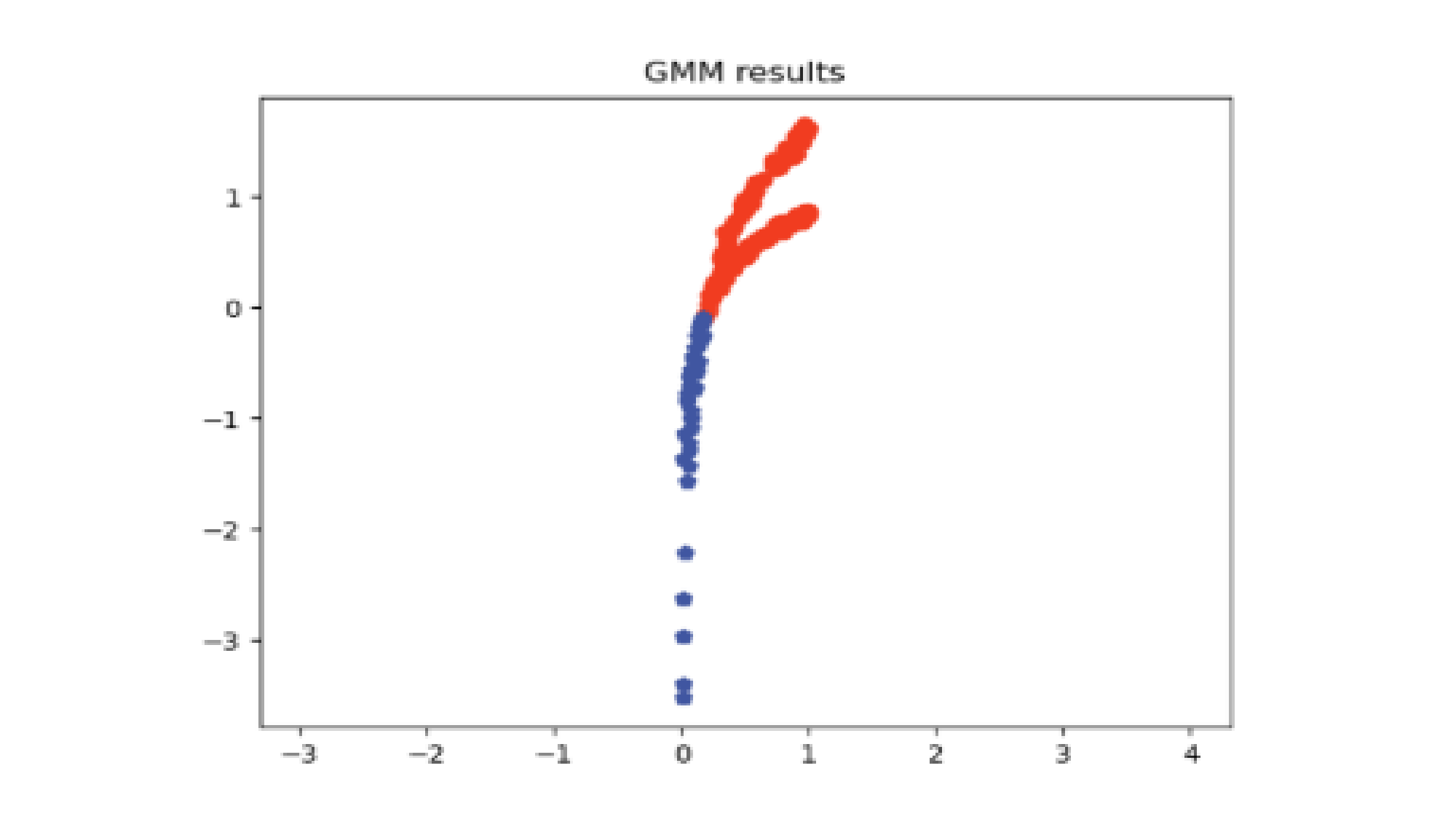}%
\label{fig_gmm}}
\subfloat[]{\includegraphics[width=1.8in]{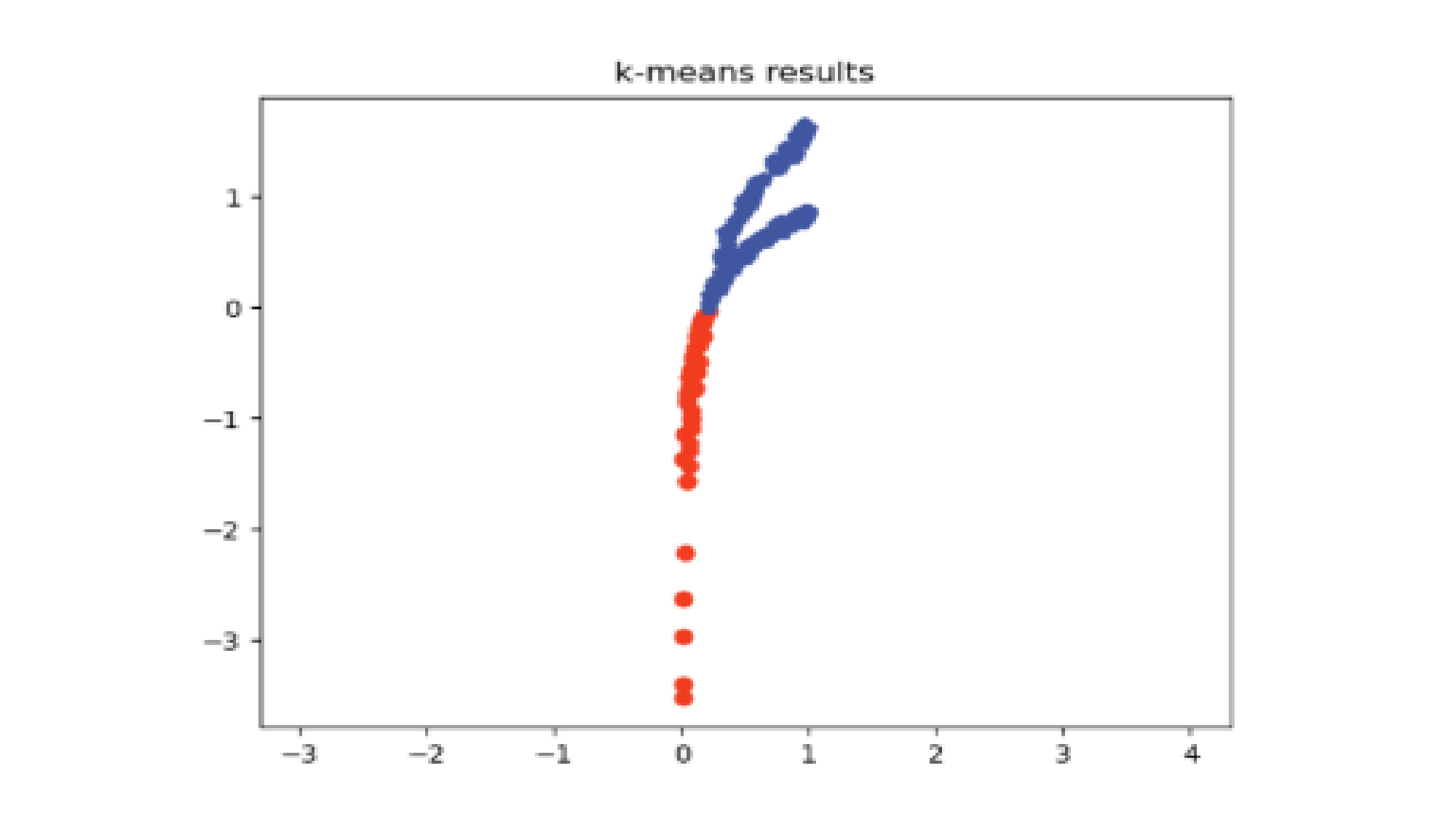}%
\label{fig_kmeans}}
\caption{The clustering results of MCVCC and comparison algorithms when $f=f_5$.}
\label{fig_comparison}
\end{figure*}

\begin{figure*}[!t]
\centering
\subfloat[]{\includegraphics[width=1.8in]{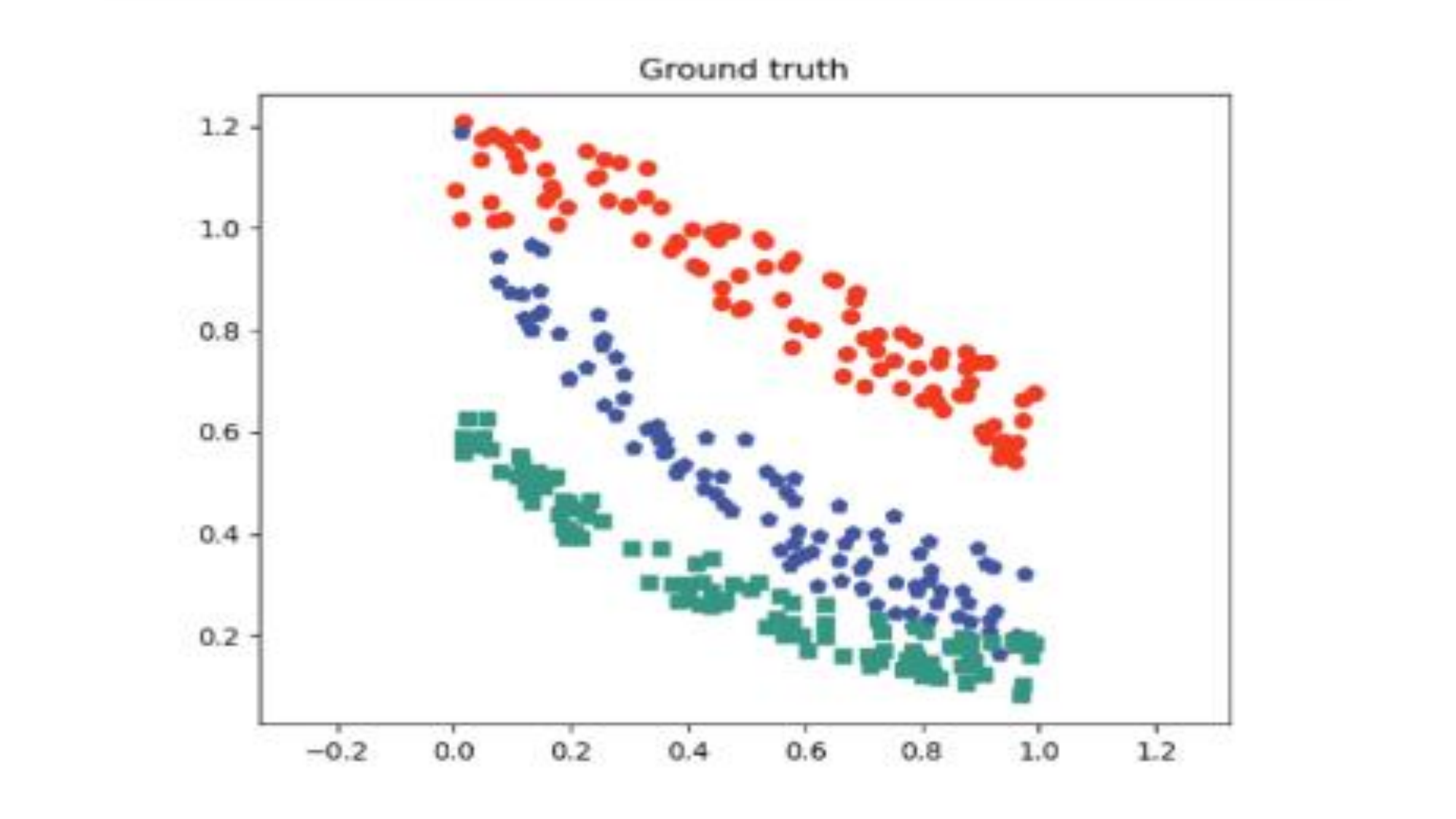}%
\label{fig_groundtruth}}
\subfloat[]{\includegraphics[width=1.8in]{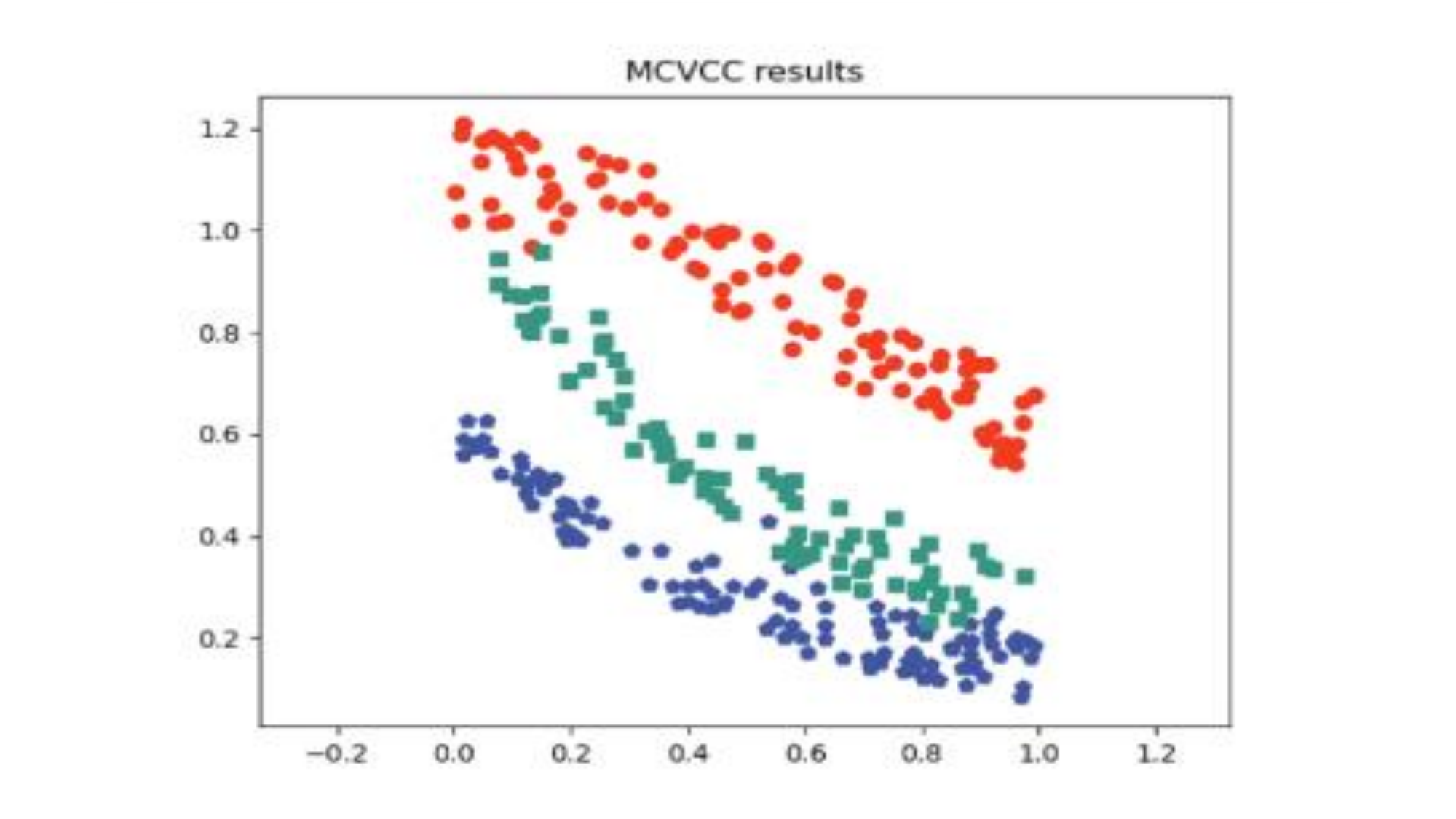}%
\label{fig_mcvcc}}
\subfloat[]{\includegraphics[width=1.8in]{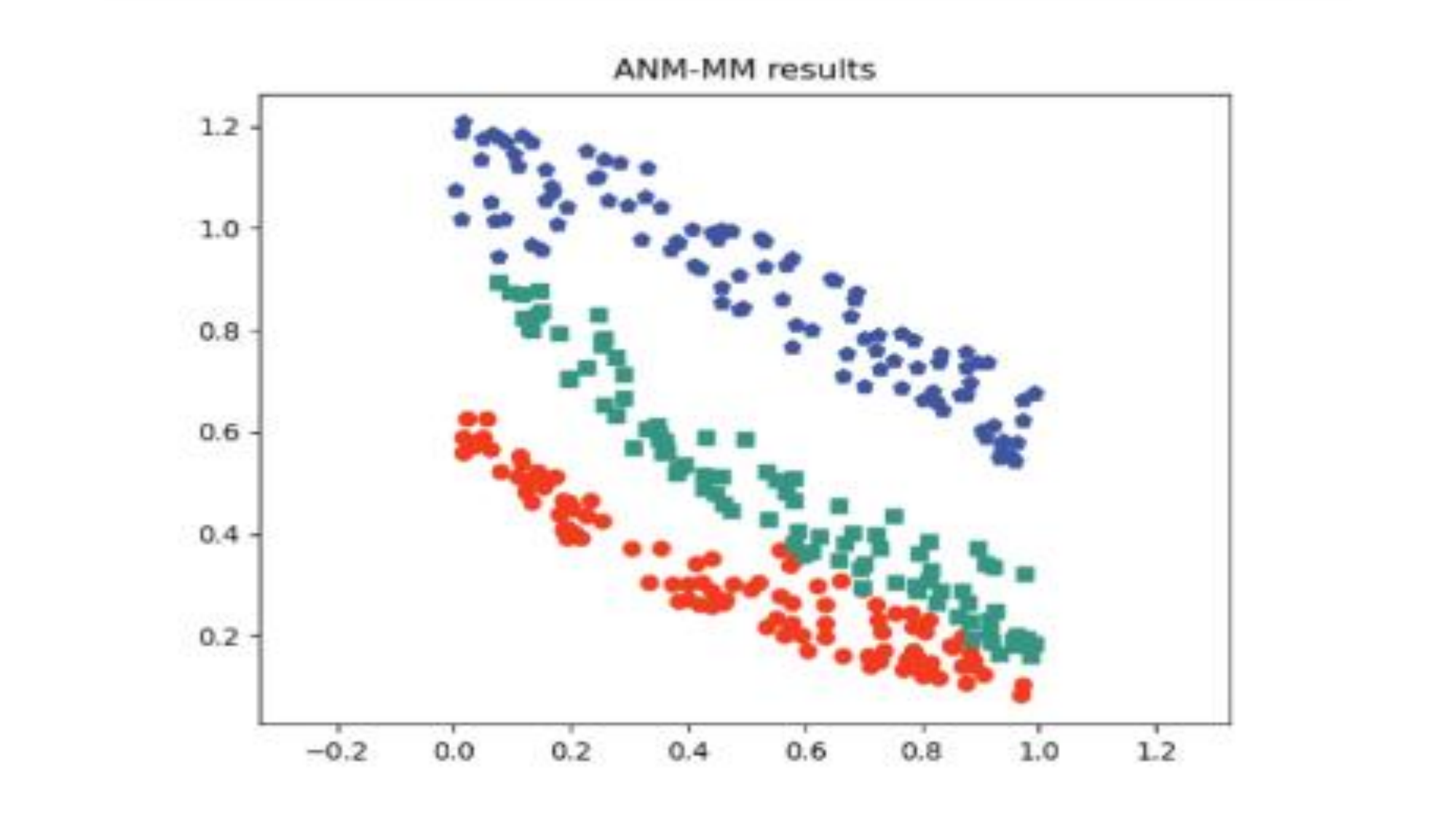}%
\label{fig_anm_mm}}
\subfloat[]{\includegraphics[width=1.8in]{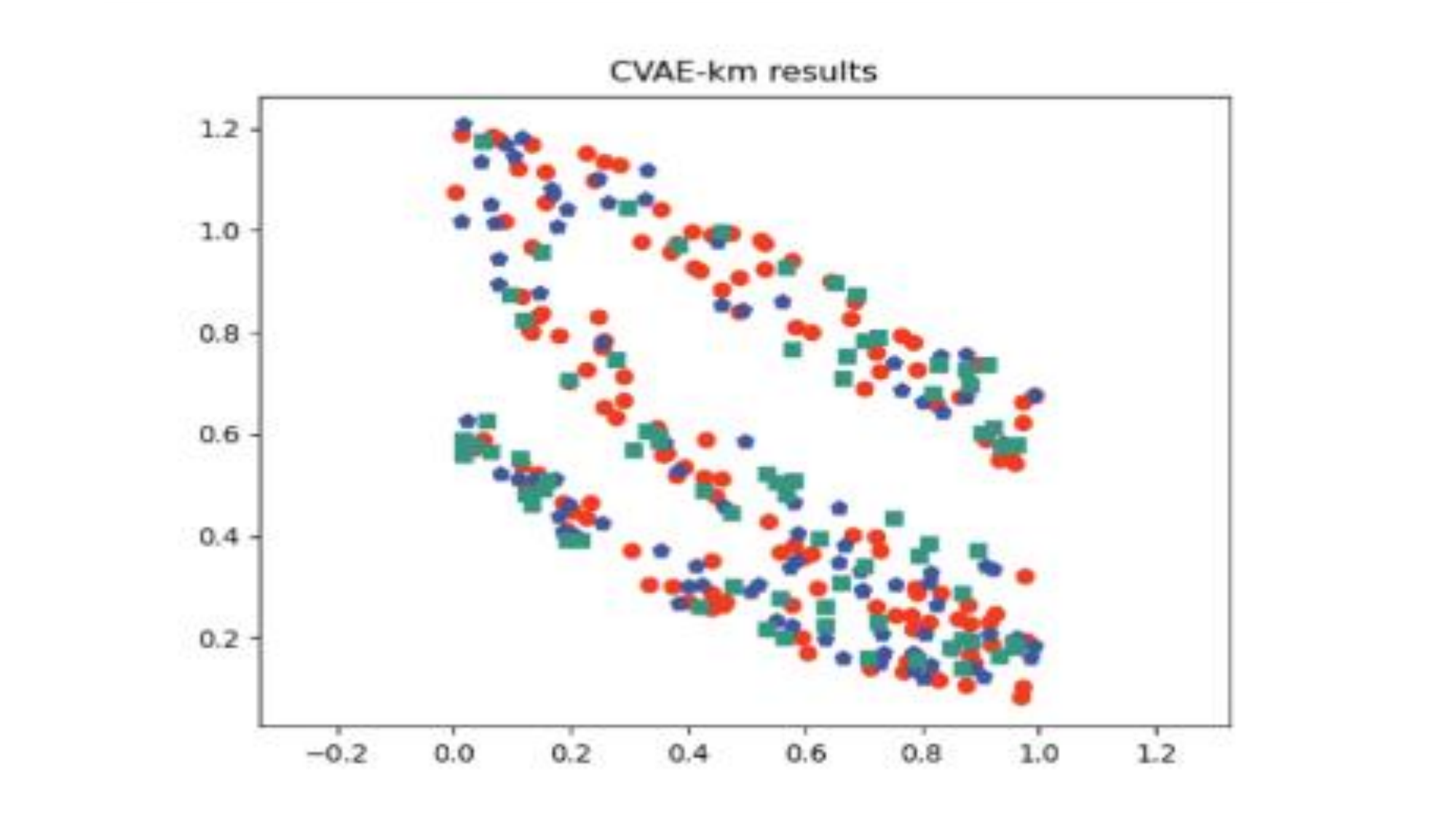}%
\label{fig_cvae_km}}
\hfil
\subfloat[]{\includegraphics[width=1.8in]{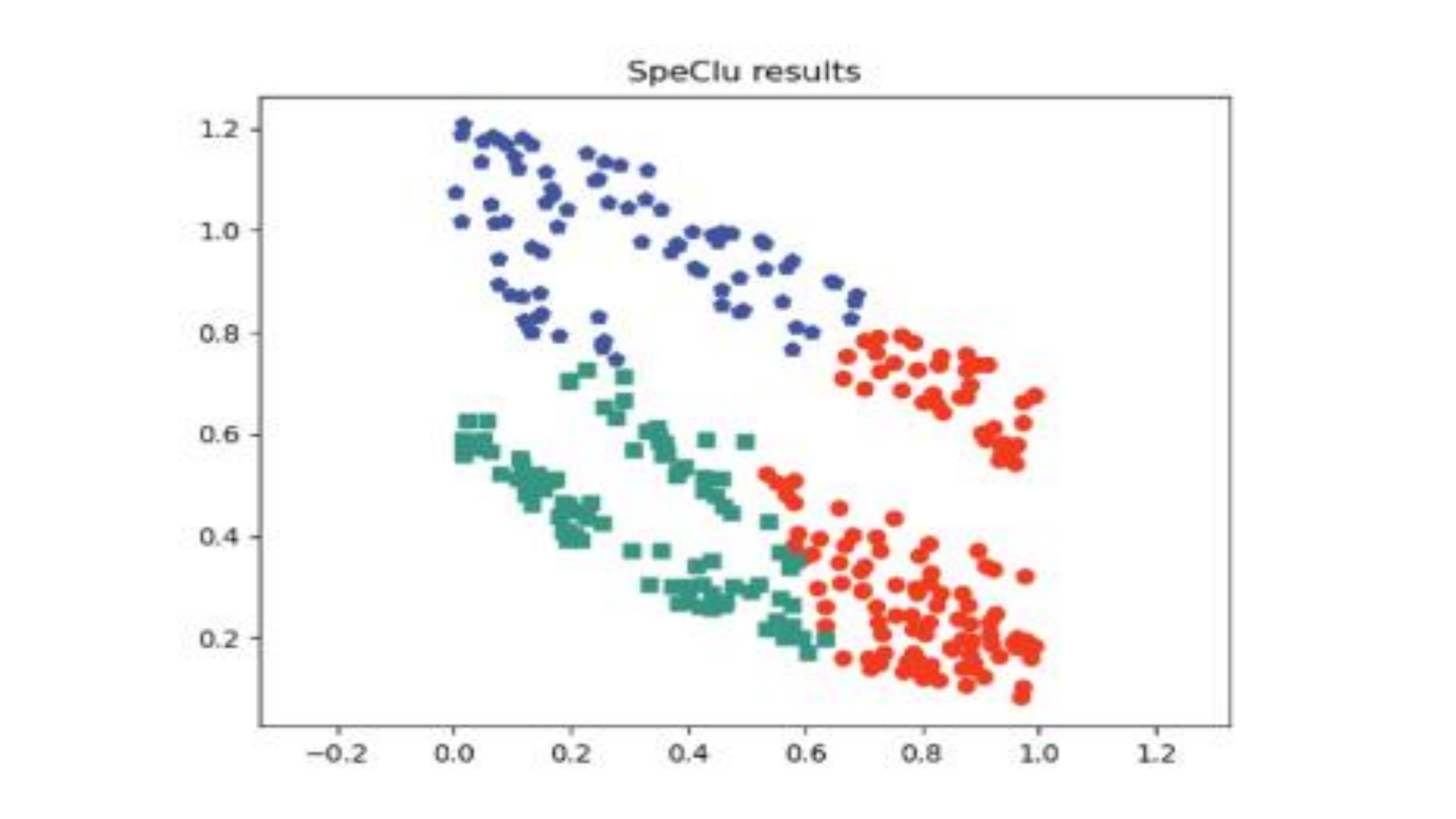}%
\label{fig_speclu}}
\subfloat[]{\includegraphics[width=1.8in]{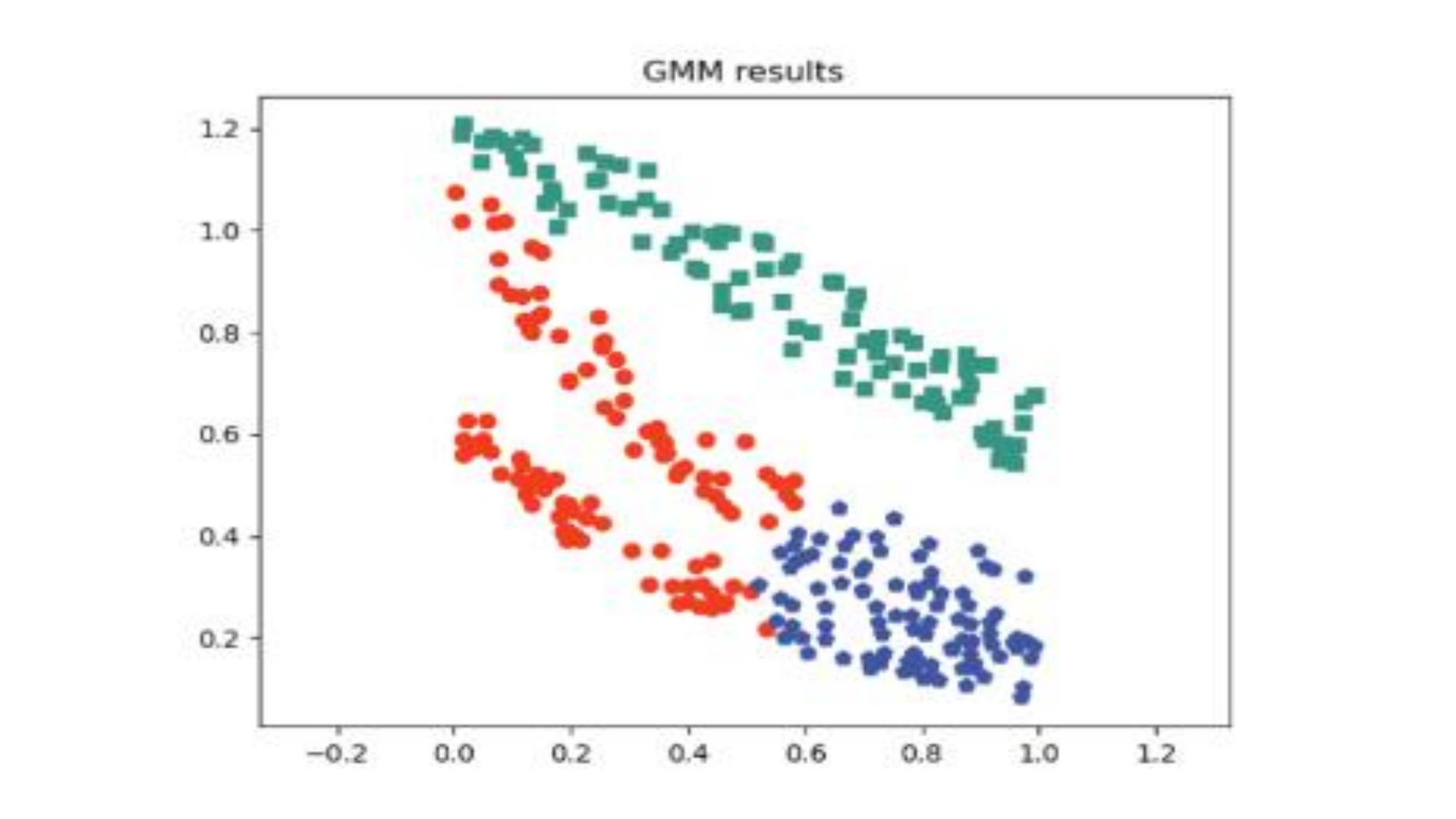}%
\label{fig_gmm}}
\subfloat[]{\includegraphics[width=1.8in]{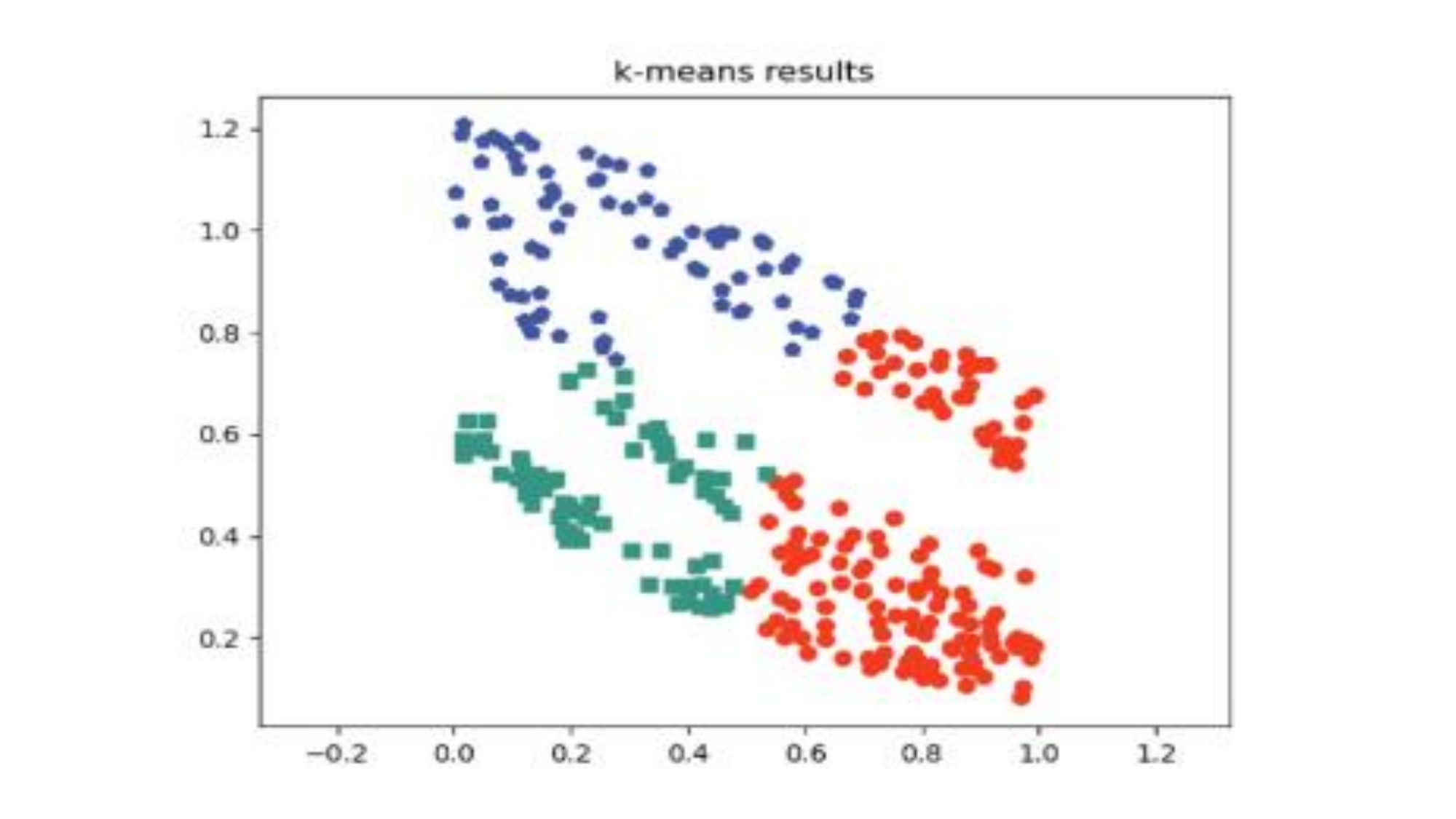}%
\label{fig_kmeans}}
\caption{The clustering results of MCVCC and comparison algorithms when $C=3$.}
\label{fig_comparison}
\end{figure*}

\begin{figure*}[!t]
\centering
\subfloat[]{\includegraphics[width=1.8in]{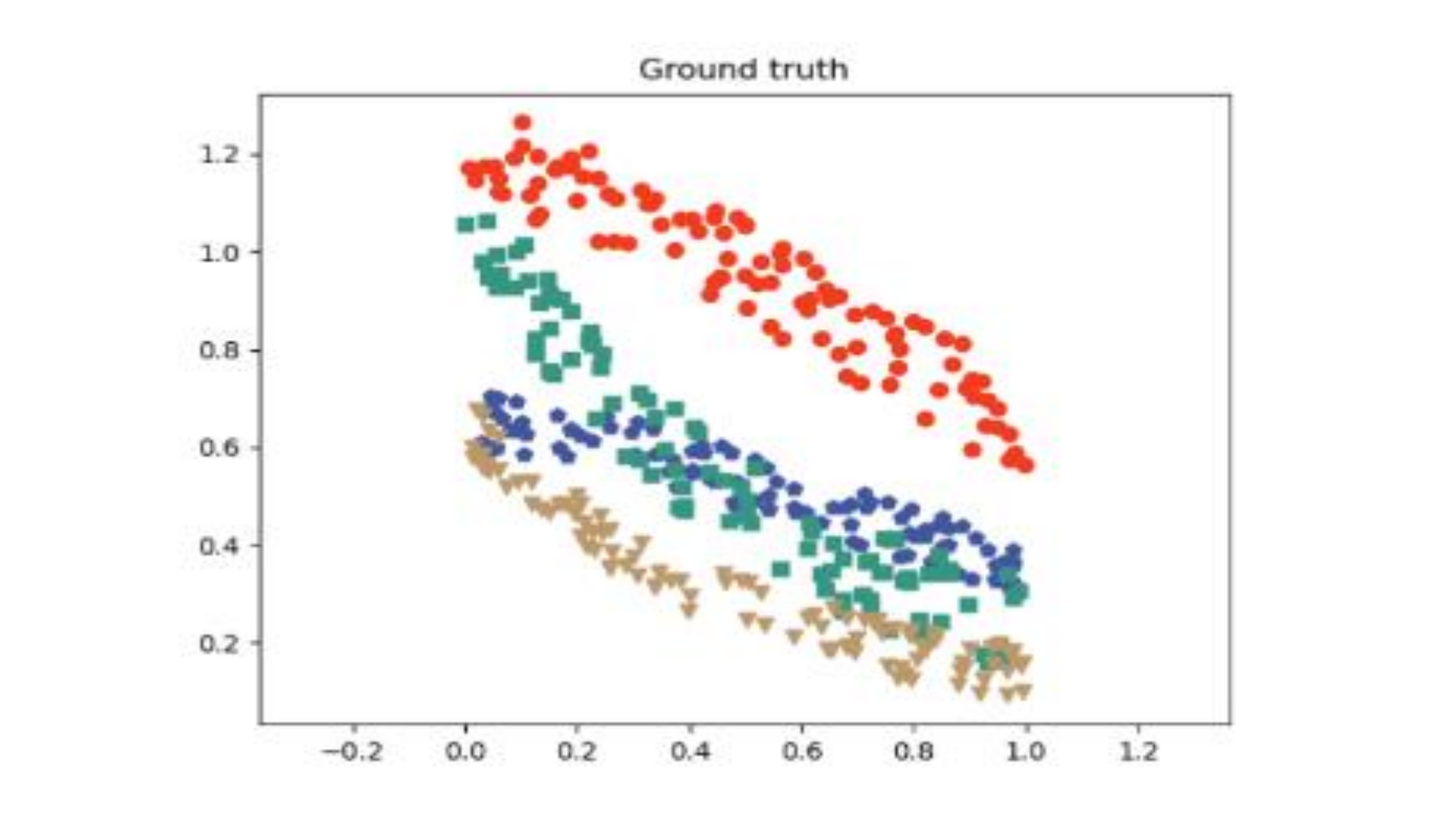}%
\label{fig_groundtruth}}
\subfloat[]{\includegraphics[width=1.8in]{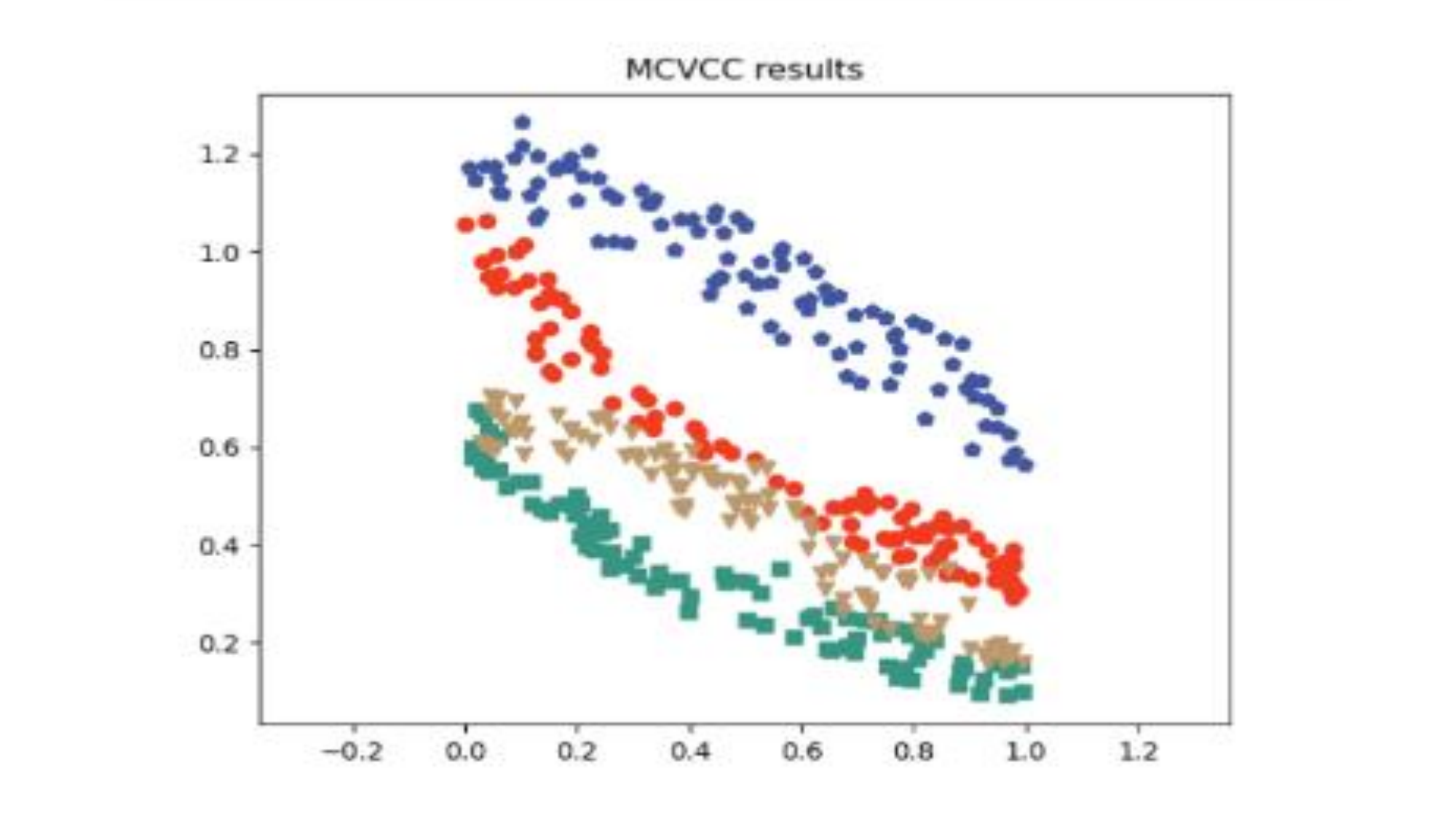}%
\label{fig_mcvcc}}
\subfloat[]{\includegraphics[width=1.8in]{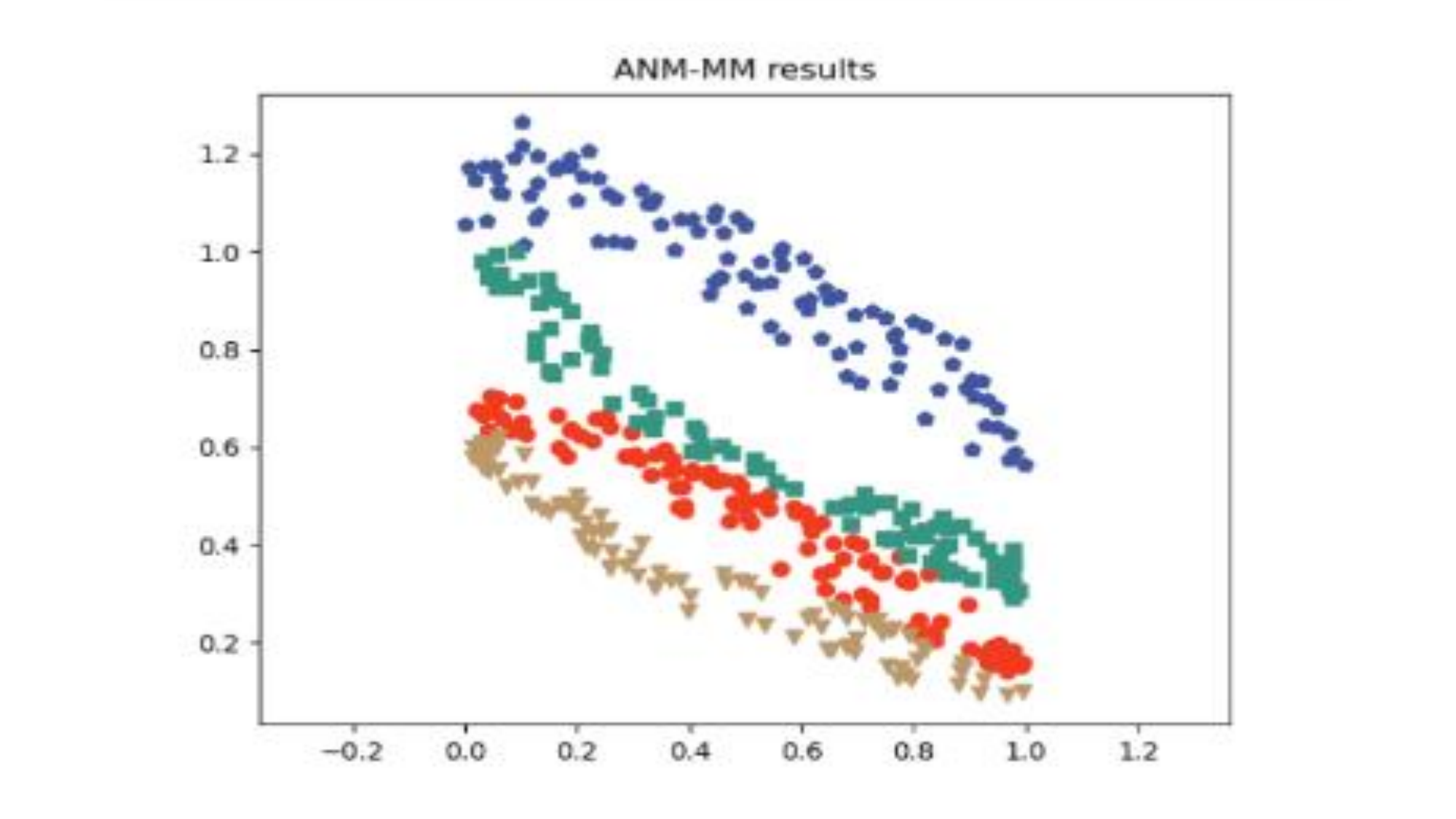}%
\label{fig_anm_mm}}
\subfloat[]{\includegraphics[width=1.8in]{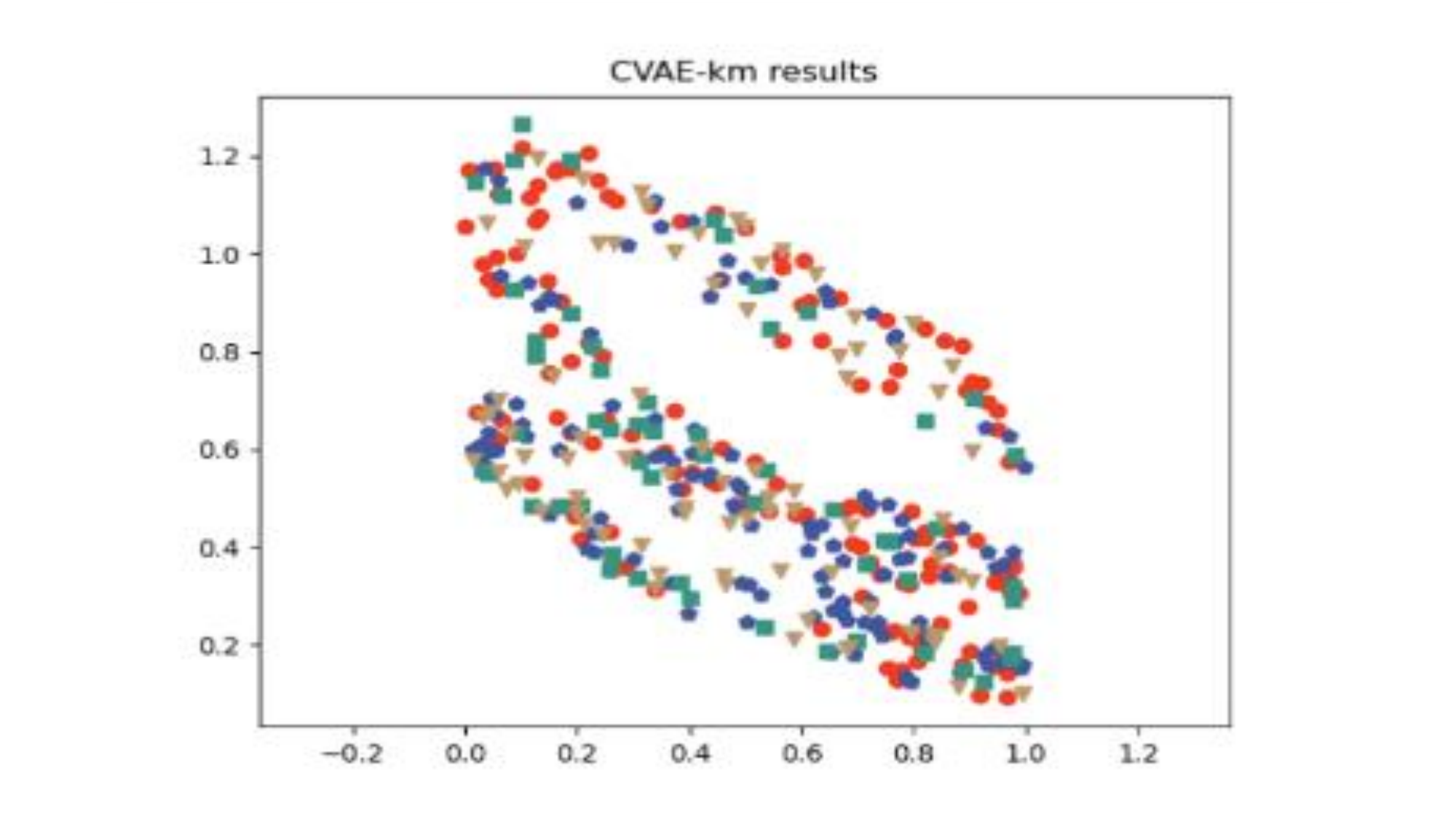}%
\label{fig_cvae_km}}
\hfil
\subfloat[]{\includegraphics[width=1.8in]{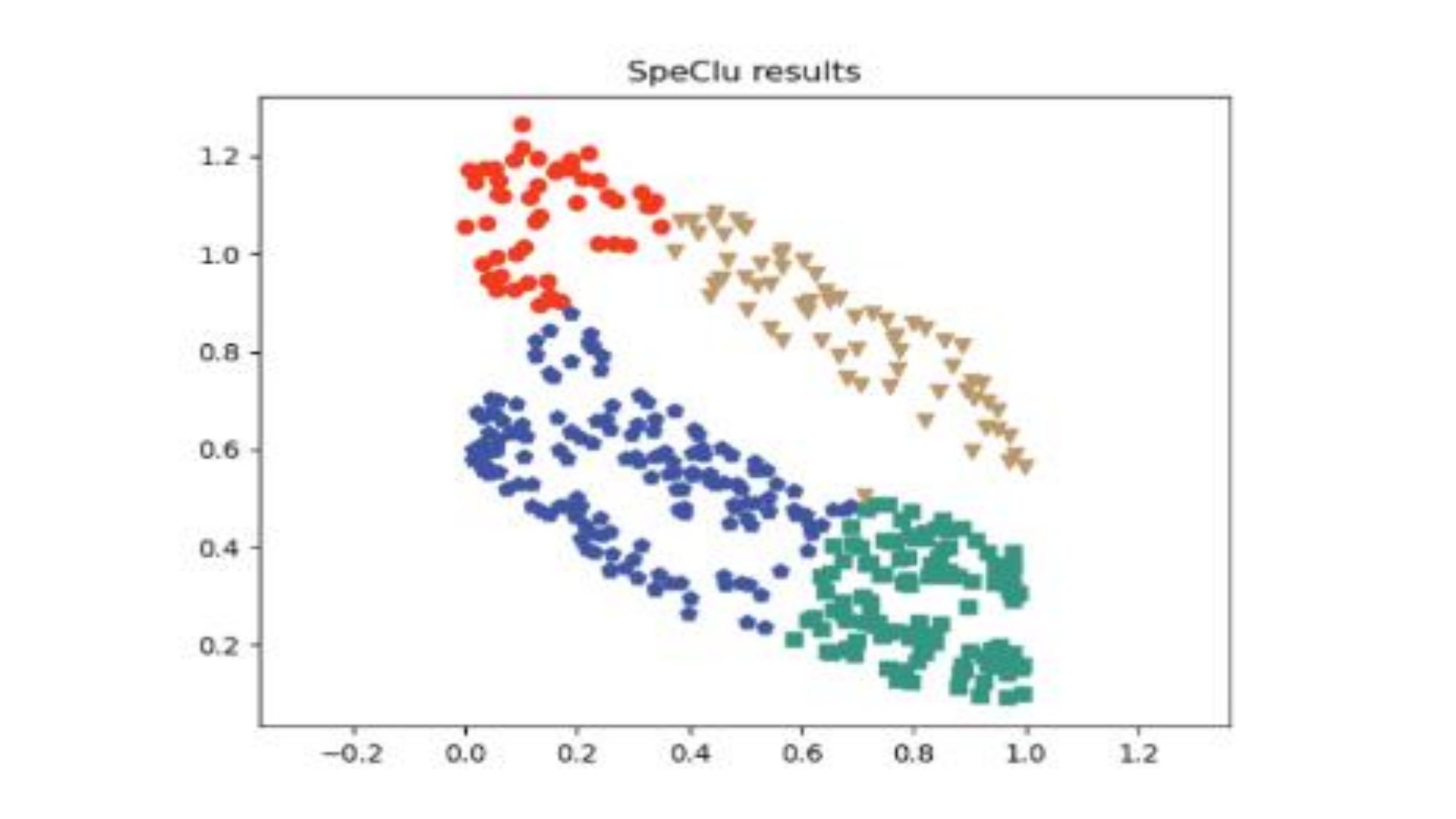}%
\label{fig_speclu}}
\subfloat[]{\includegraphics[width=1.8in]{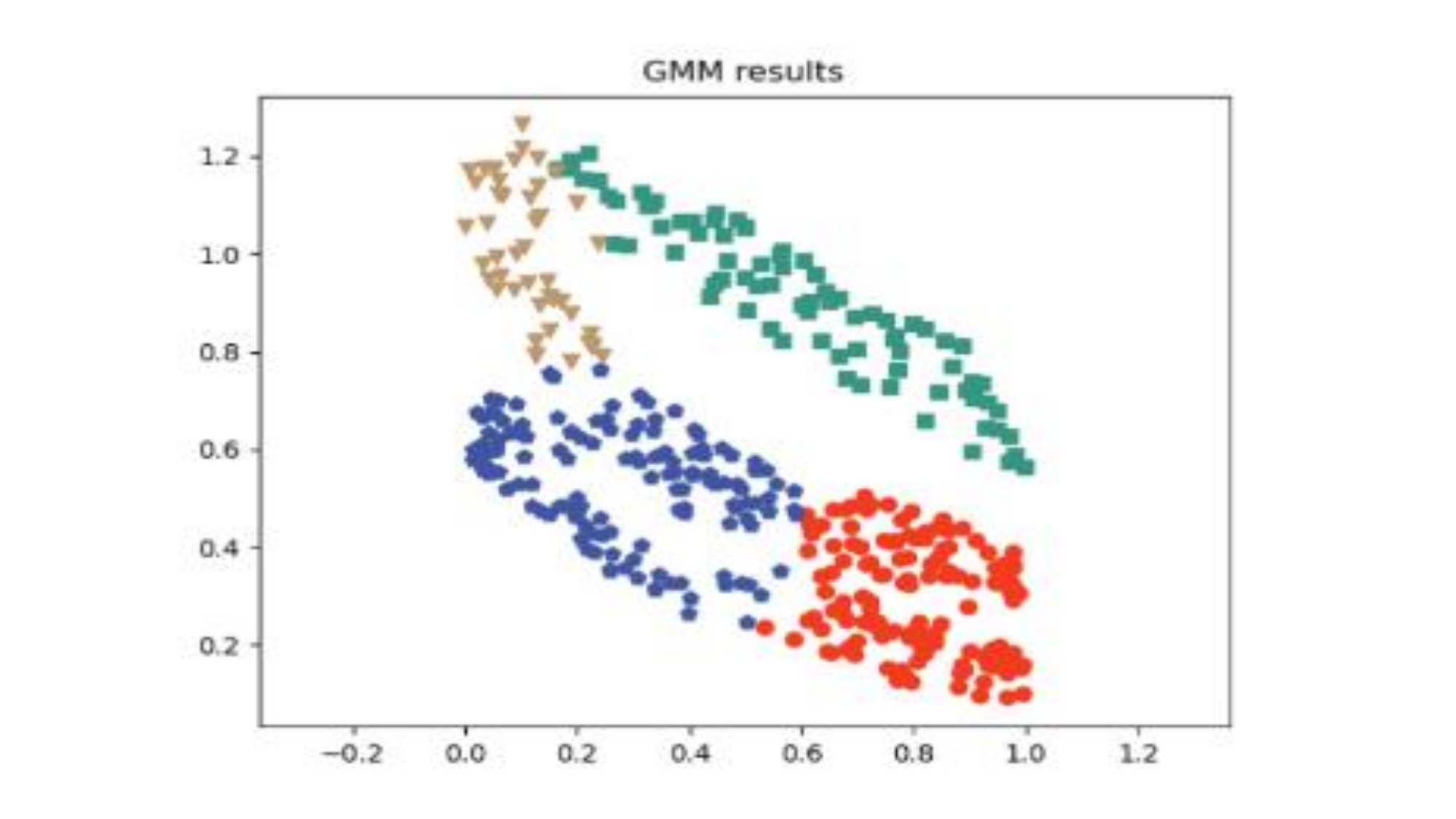}%
\label{fig_gmm}}
\subfloat[]{\includegraphics[width=1.8in]{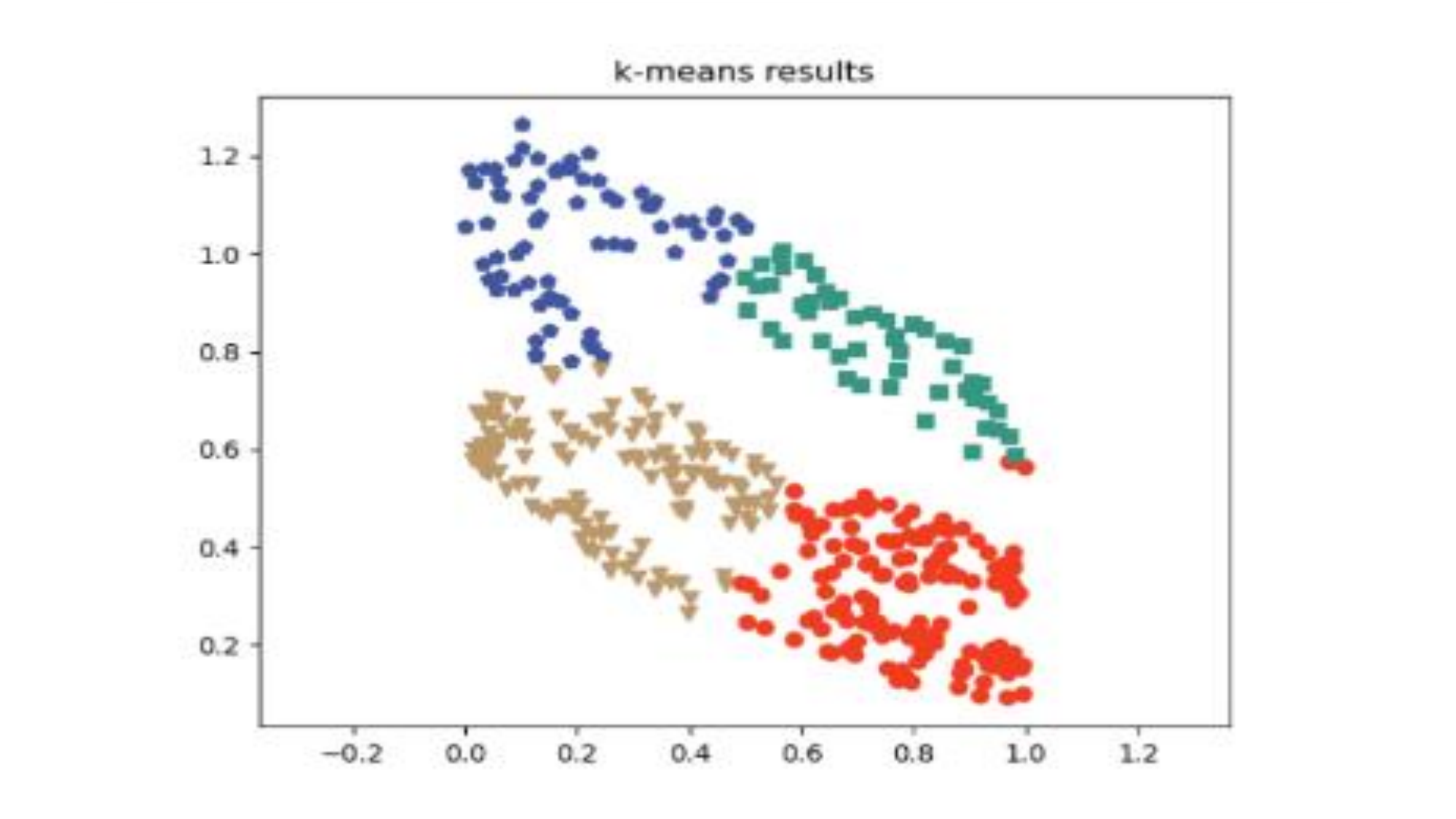}%
\label{fig_kmeans}}
\caption{The clustering results of MCVCC and comparison algorithms when $C=4$.}
\label{fig_comparison}
\end{figure*}

\begin{figure*}[!t]
\centering
\subfloat[]{\includegraphics[width=1.8in]{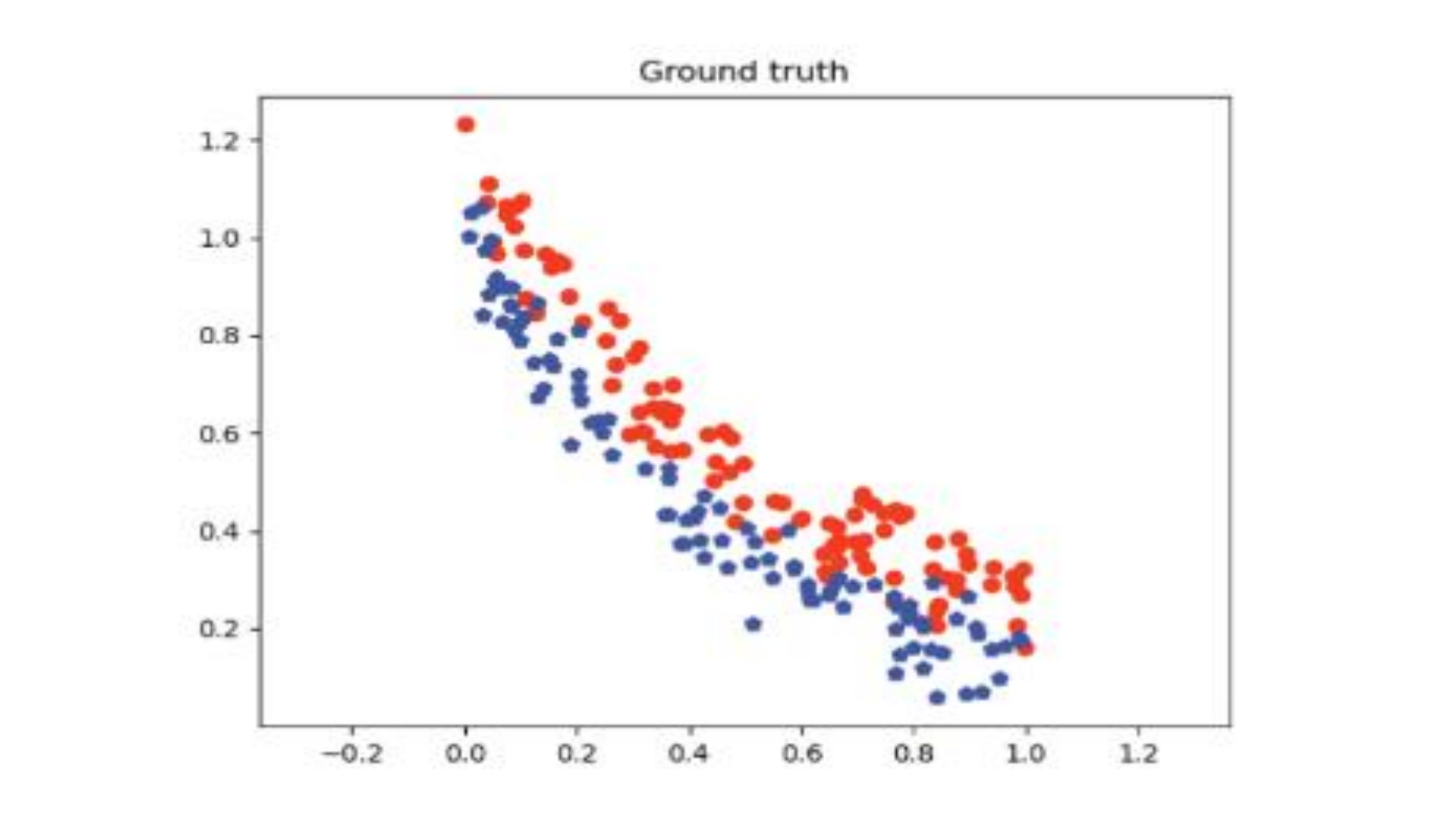}%
\label{fig_groundtruth}}
\subfloat[]{\includegraphics[width=1.8in]{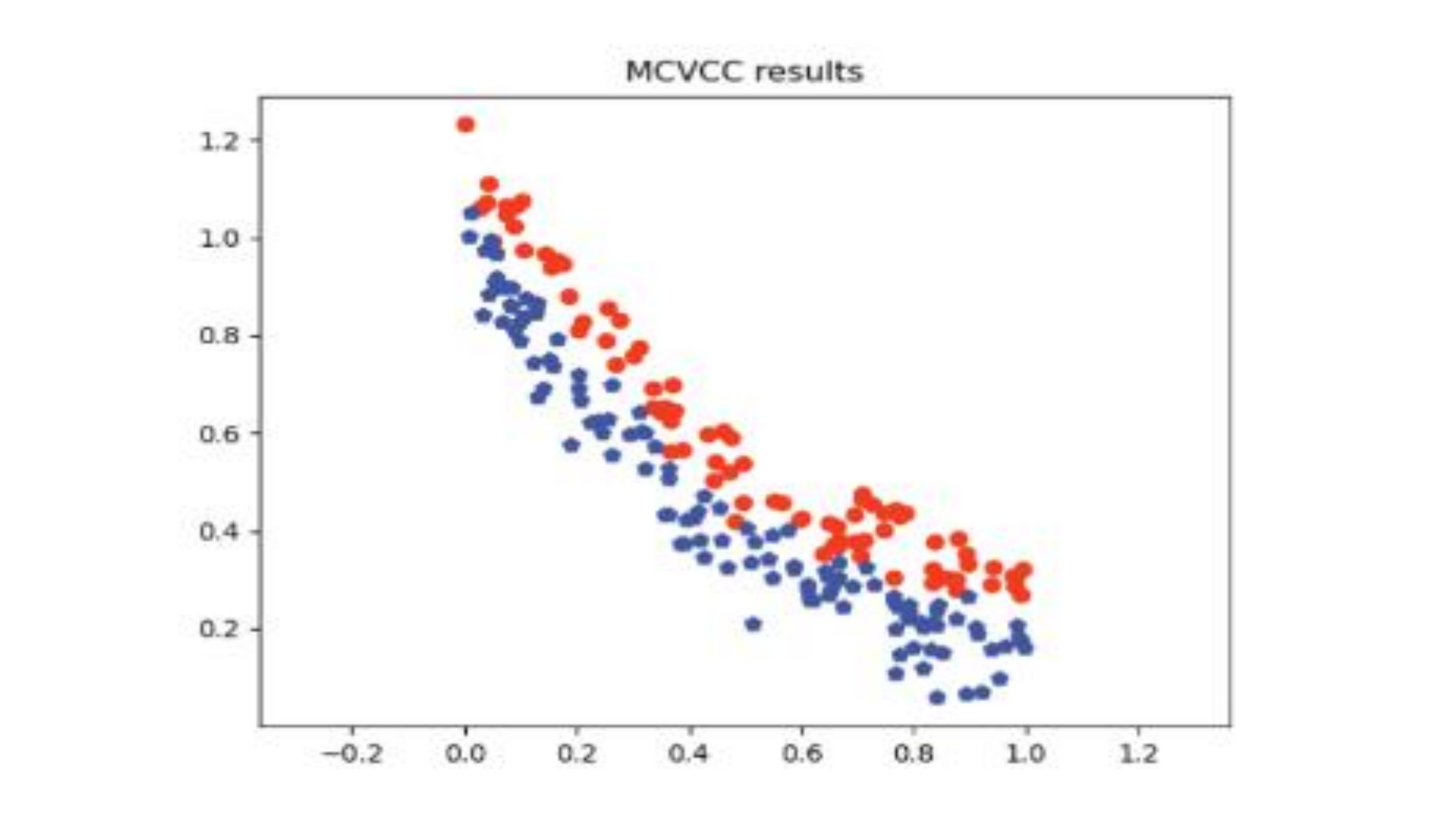}%
\label{fig_mcvcc}}
\subfloat[]{\includegraphics[width=1.8in]{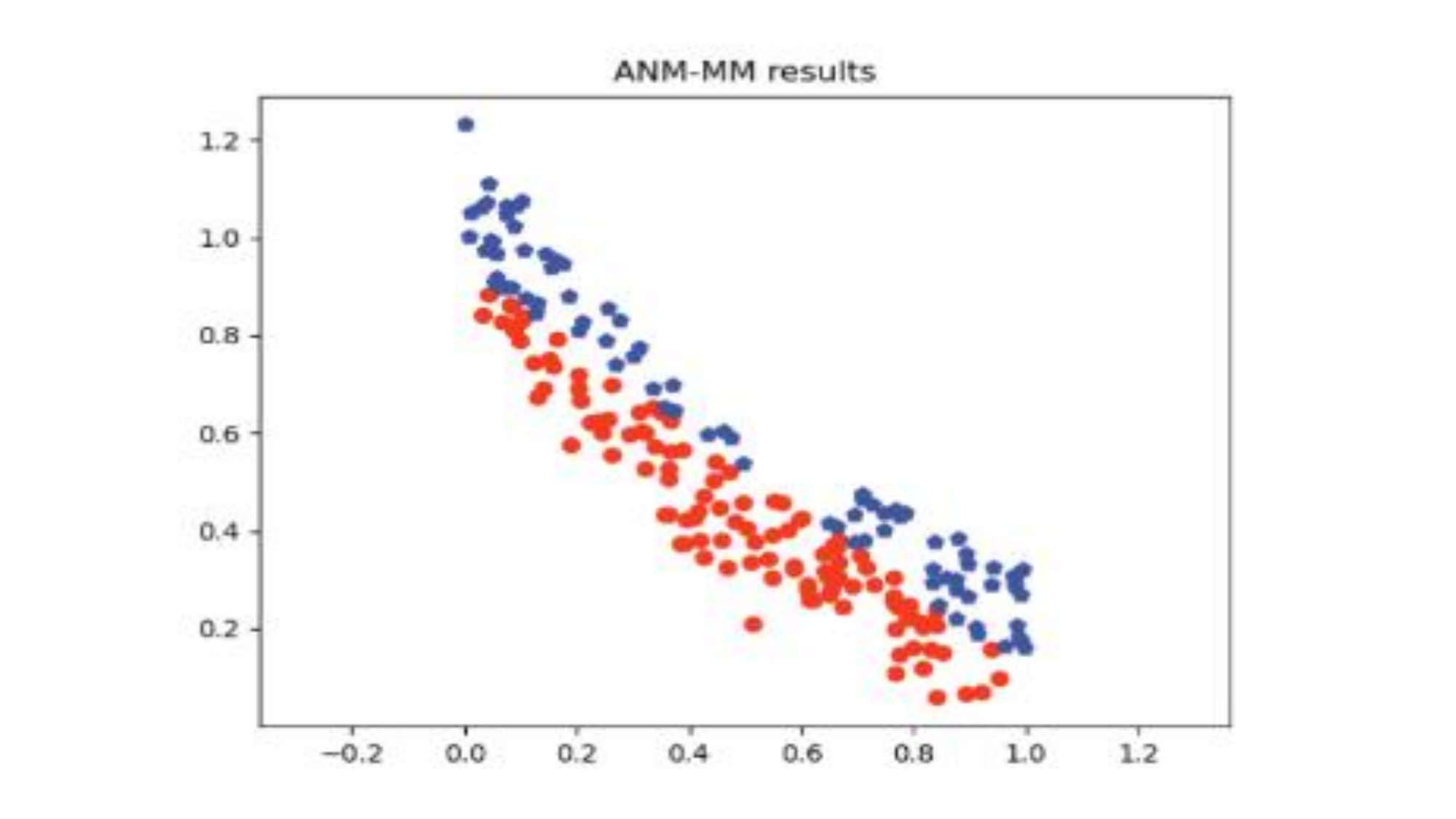}%
\label{fig_anm_mm}}
\subfloat[]{\includegraphics[width=1.8in]{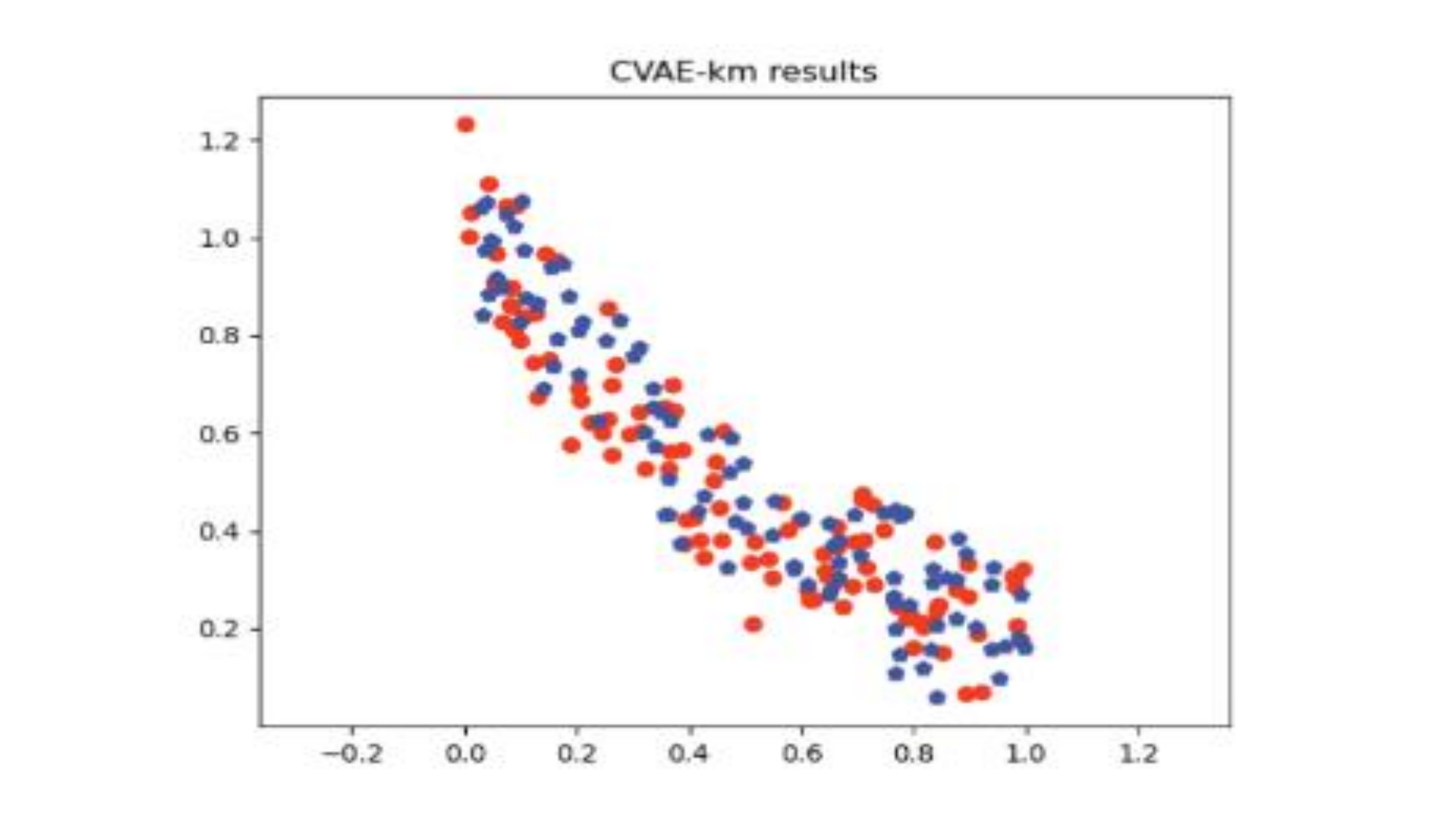}%
\label{fig_cvae_km}}
\hfil
\subfloat[]{\includegraphics[width=1.8in]{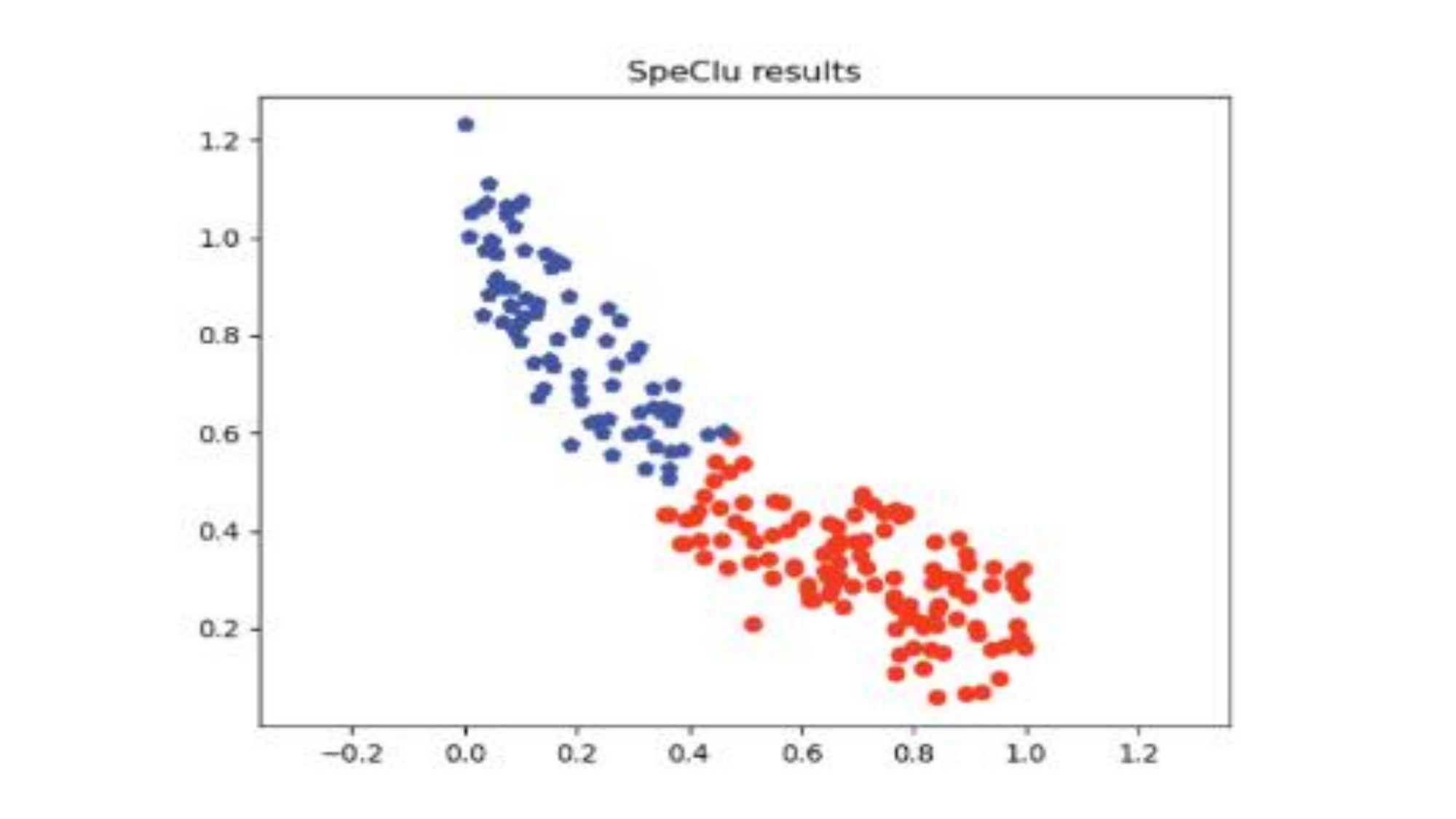}%
\label{fig_speclu}}
\subfloat[]{\includegraphics[width=1.8in]{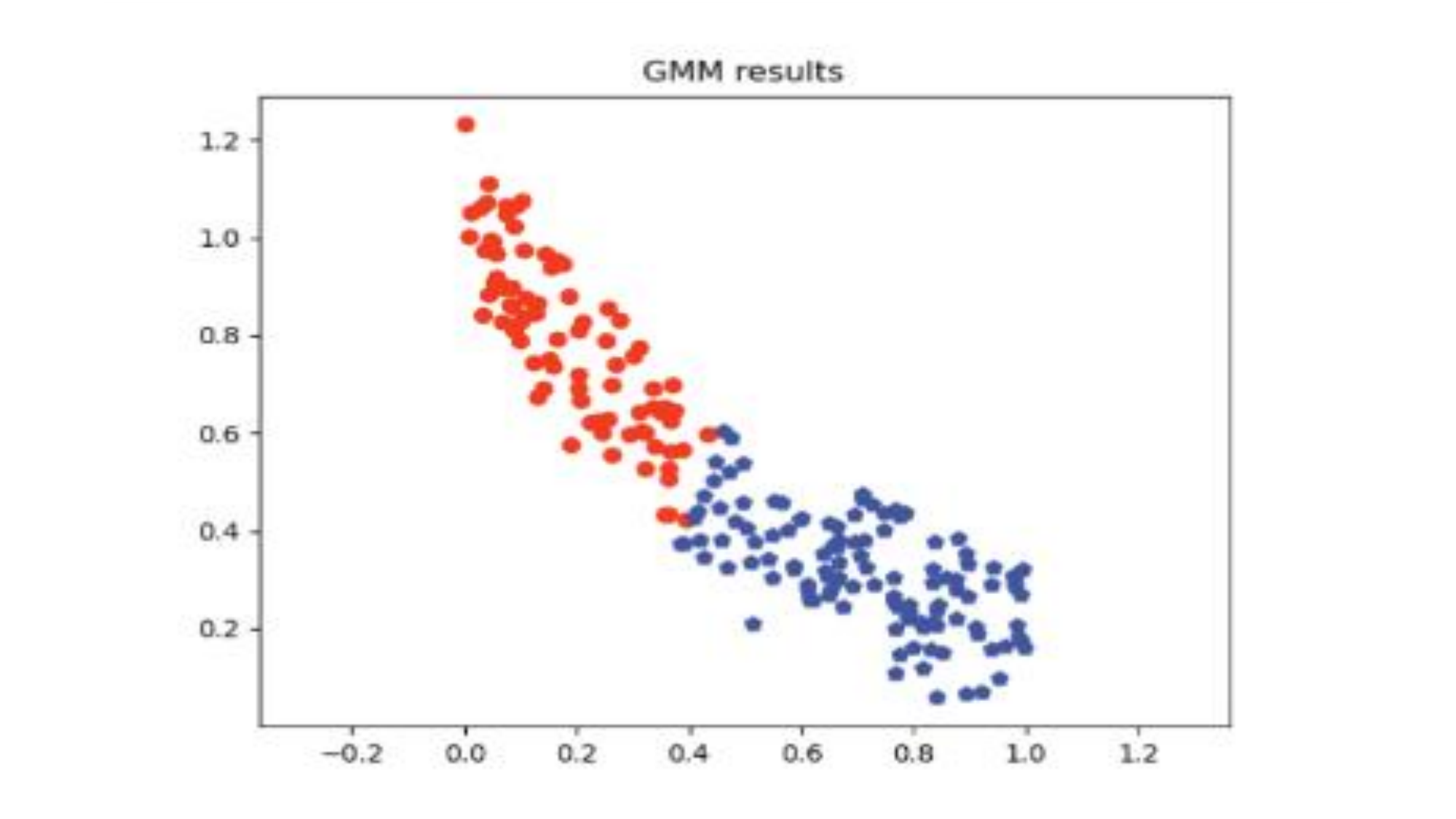}%
\label{fig_gmm}}
\subfloat[]{\includegraphics[width=1.8in]{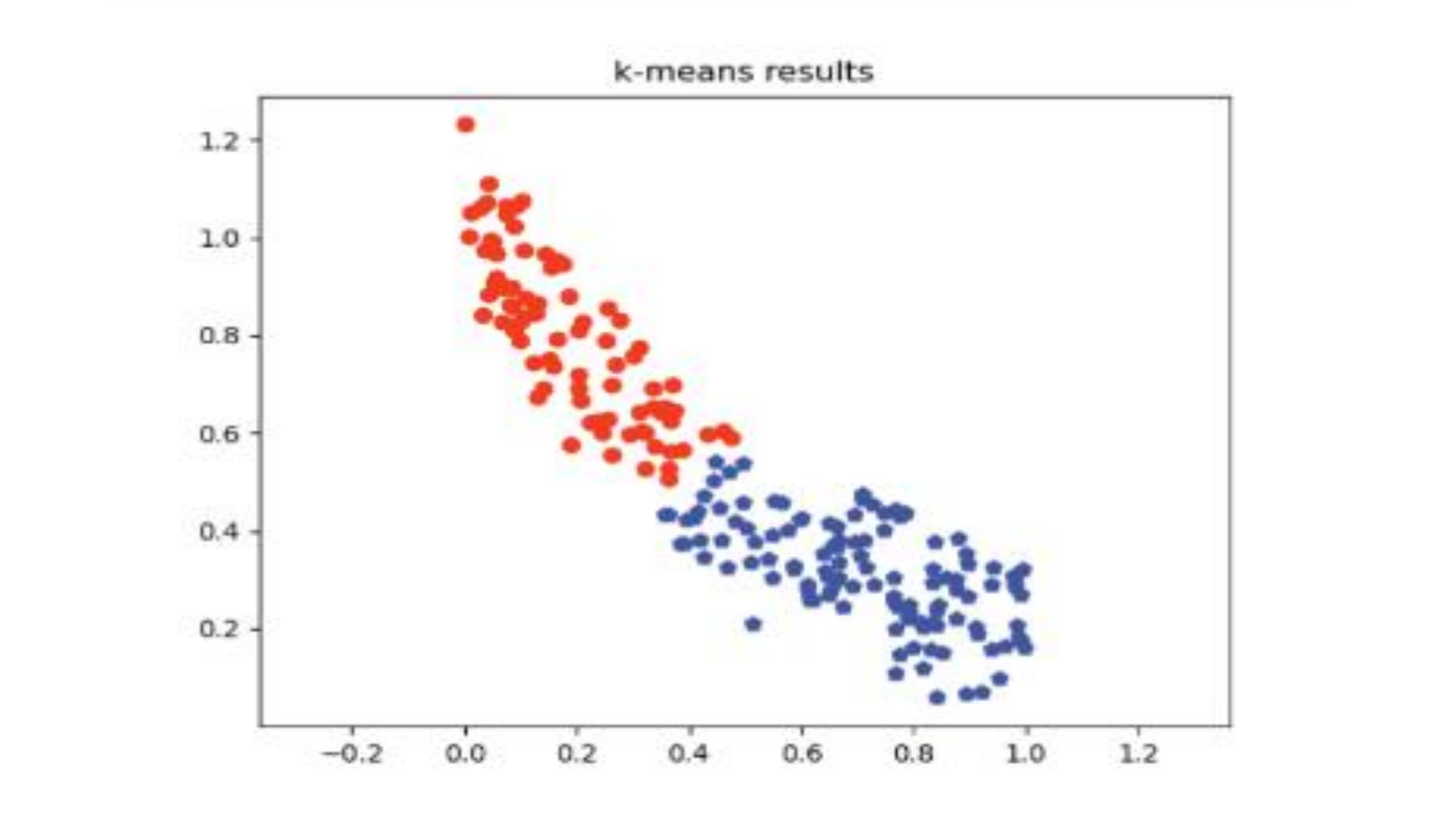}%
\label{fig_kmeans}}
\caption{The clustering results of MCVCC and comparison algorithms when $\sigma$=0.2 and 0.05 with $f=f_2$.}
\label{fig_comparison}
\end{figure*}

\begin{figure*}[!t]
\centering
\subfloat[]{\includegraphics[width=1.8in]{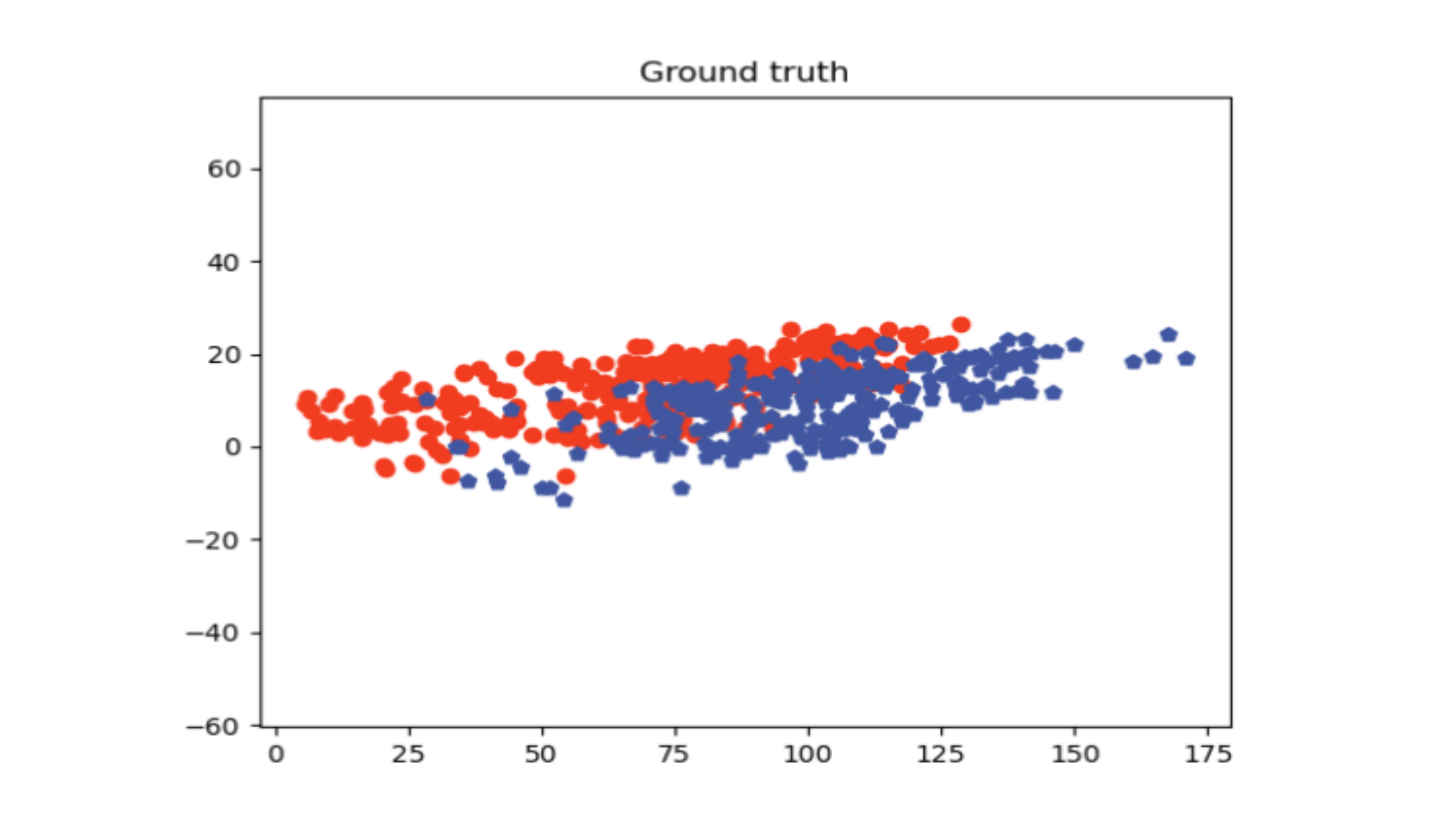}%
\label{fig_groundtruth}}
\subfloat[]{\includegraphics[width=1.8in]{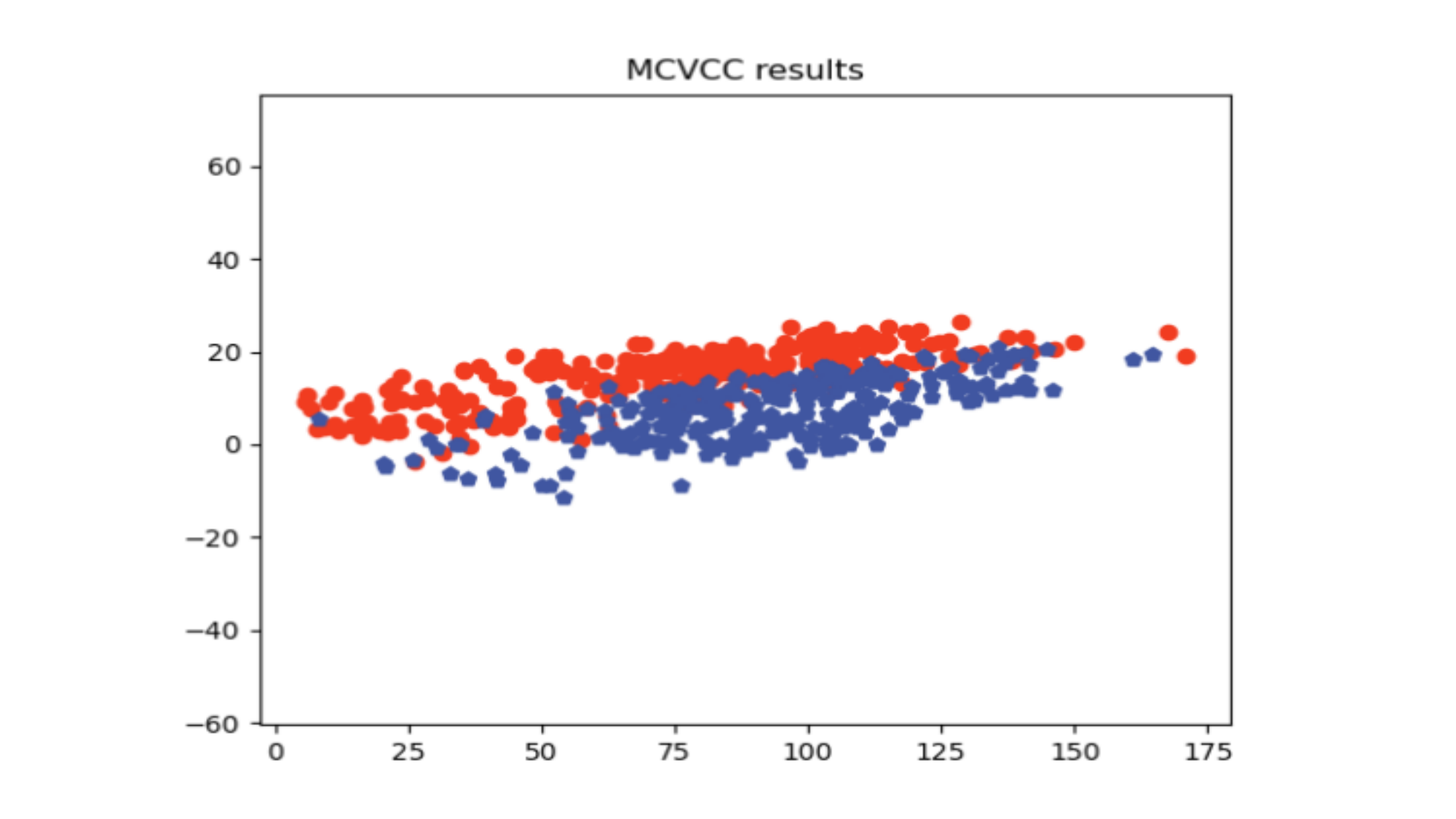}%
\label{fig_mcvcc}}
\subfloat[]{\includegraphics[width=1.8in]{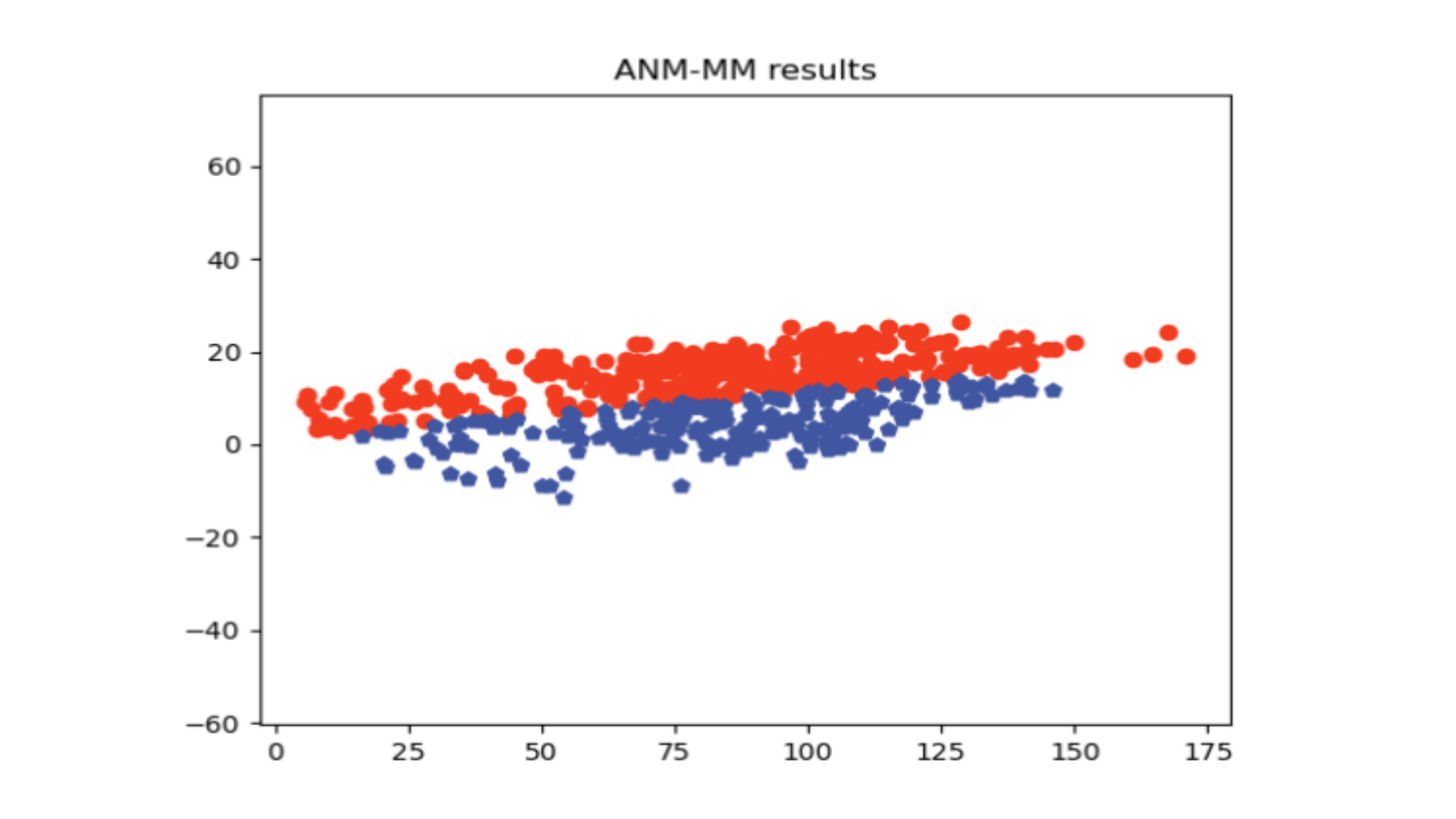}%
\label{fig_anm_mm}}
\subfloat[]{\includegraphics[width=1.8in]{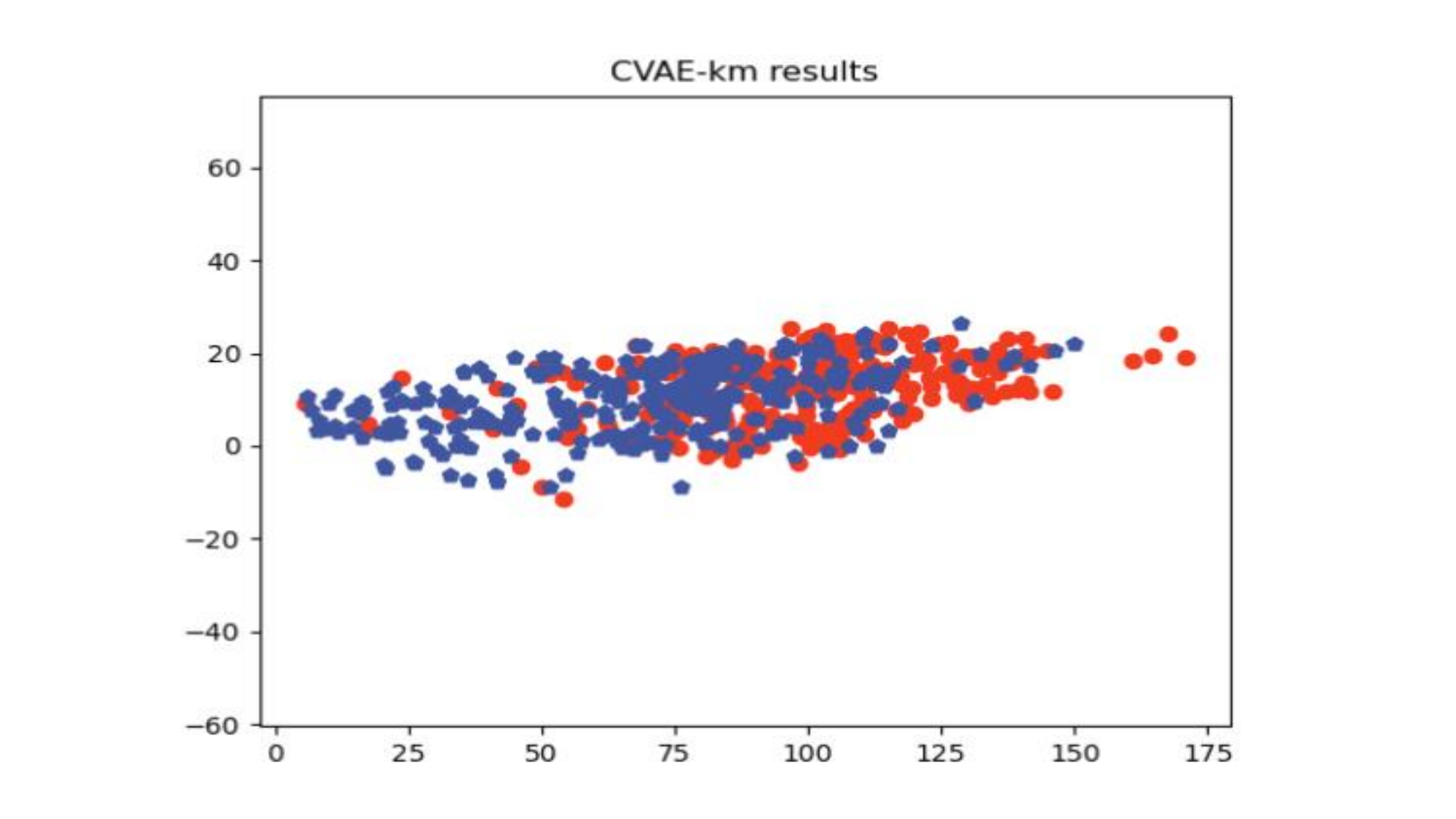}%
\label{fig_cvae_km}}
\hfil
\subfloat[]{\includegraphics[width=1.8in]{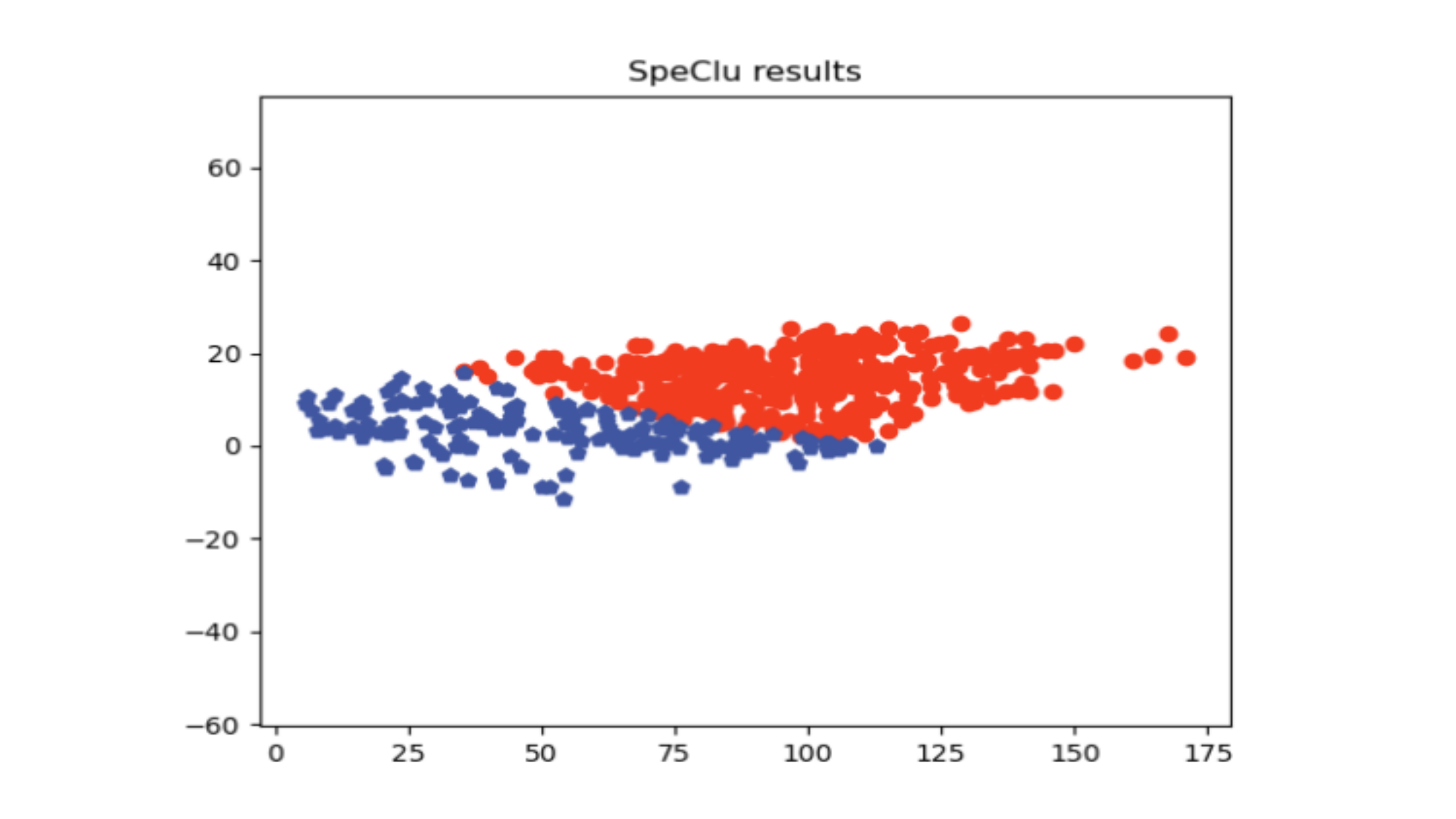}%
\label{fig_speclu}}
\subfloat[]{\includegraphics[width=1.8in]{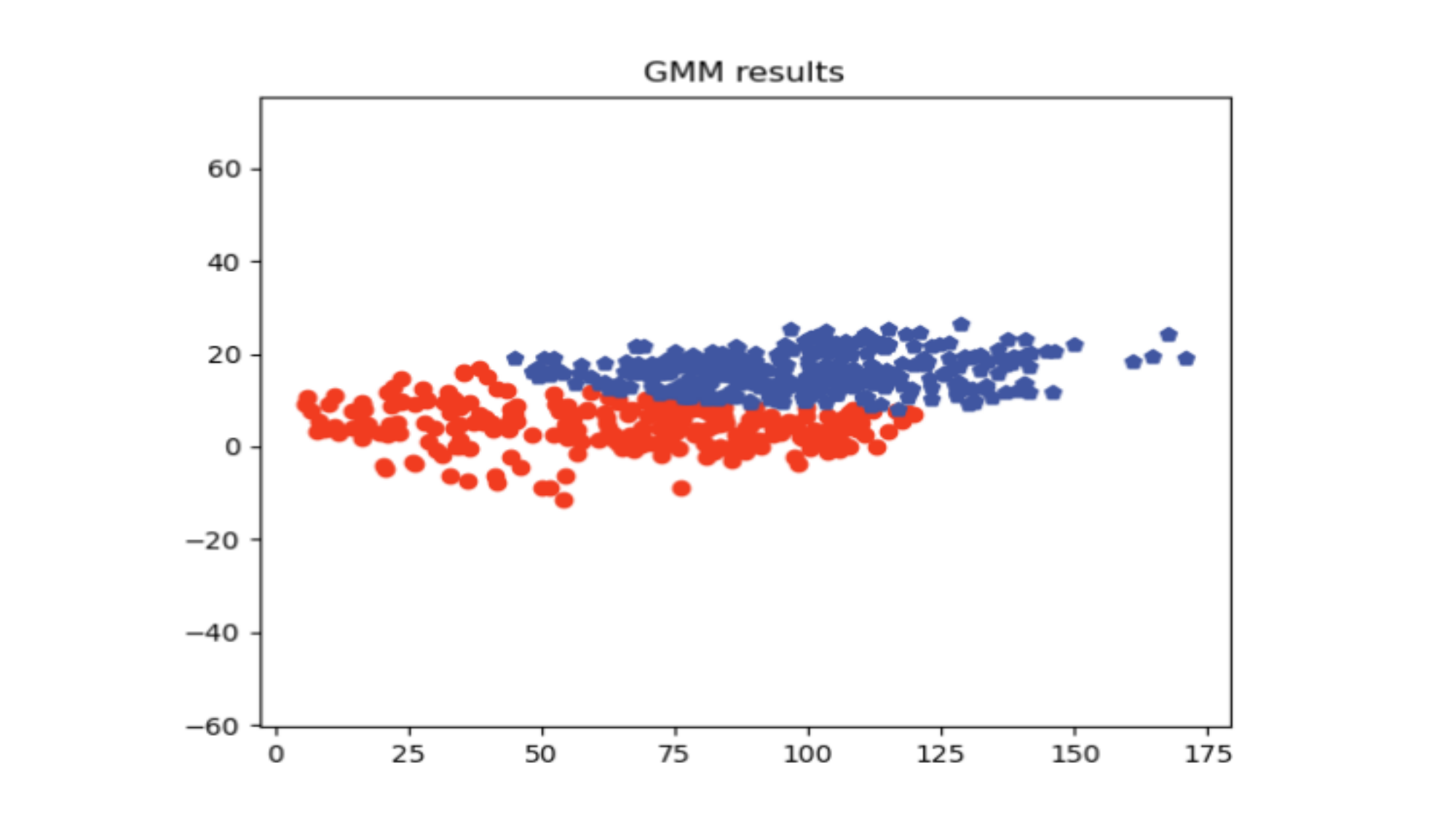}%
\label{fig_gmm}}
\subfloat[]{\includegraphics[width=1.8in]{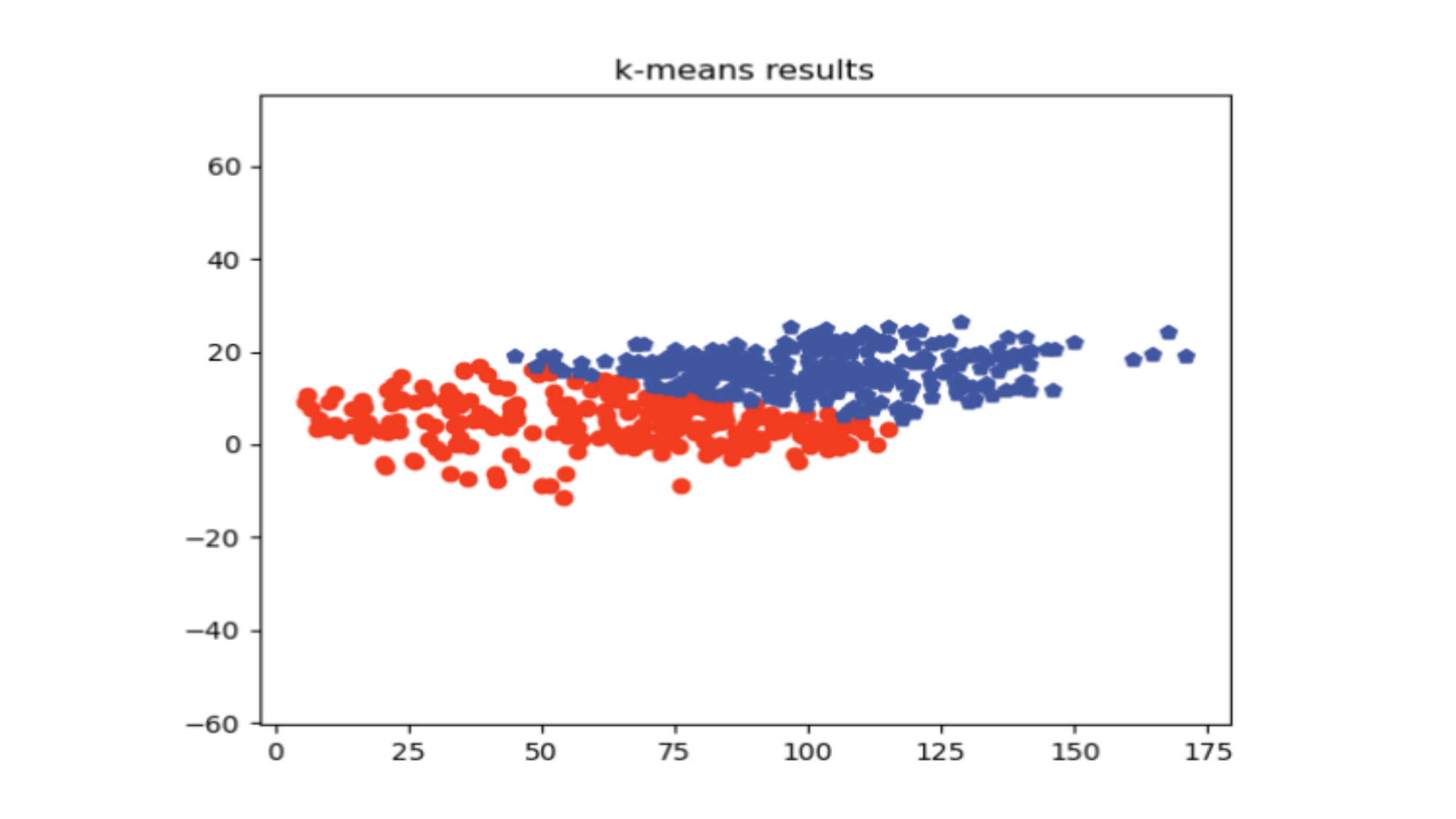}%
\label{fig_kmeans}}
\caption{Clustering results of comparison algorithms and MCVCC on BAFU air data.}
\label{fig_comparison}
\end{figure*}

\vspace{-4em}
\begin{IEEEbiography}[{\includegraphics[width=1in,height=1.25in,clip,keepaspectratio]{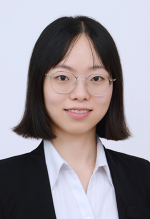}}]{Saixiong Liu}
Saixiong Liu is currently pursuing the Ph.D.degree at the School of Computer Science and Information Engineering and  working at the Institute of Big Data Science and Industry, Shaxi University, China. She received her B.S. in Network Engineering and M.S. degrees in Software Engineering from School of Computer Science and Communication Engineering in Jiangsu University in 2017 and 2020. Respectively, her mainresearch interests include causal discovery and unsupervised learning.
\end{IEEEbiography}

\vspace{-3em}
\begin{IEEEbiography}[{\includegraphics[width=1in,height=1.25in,clip,keepaspectratio]{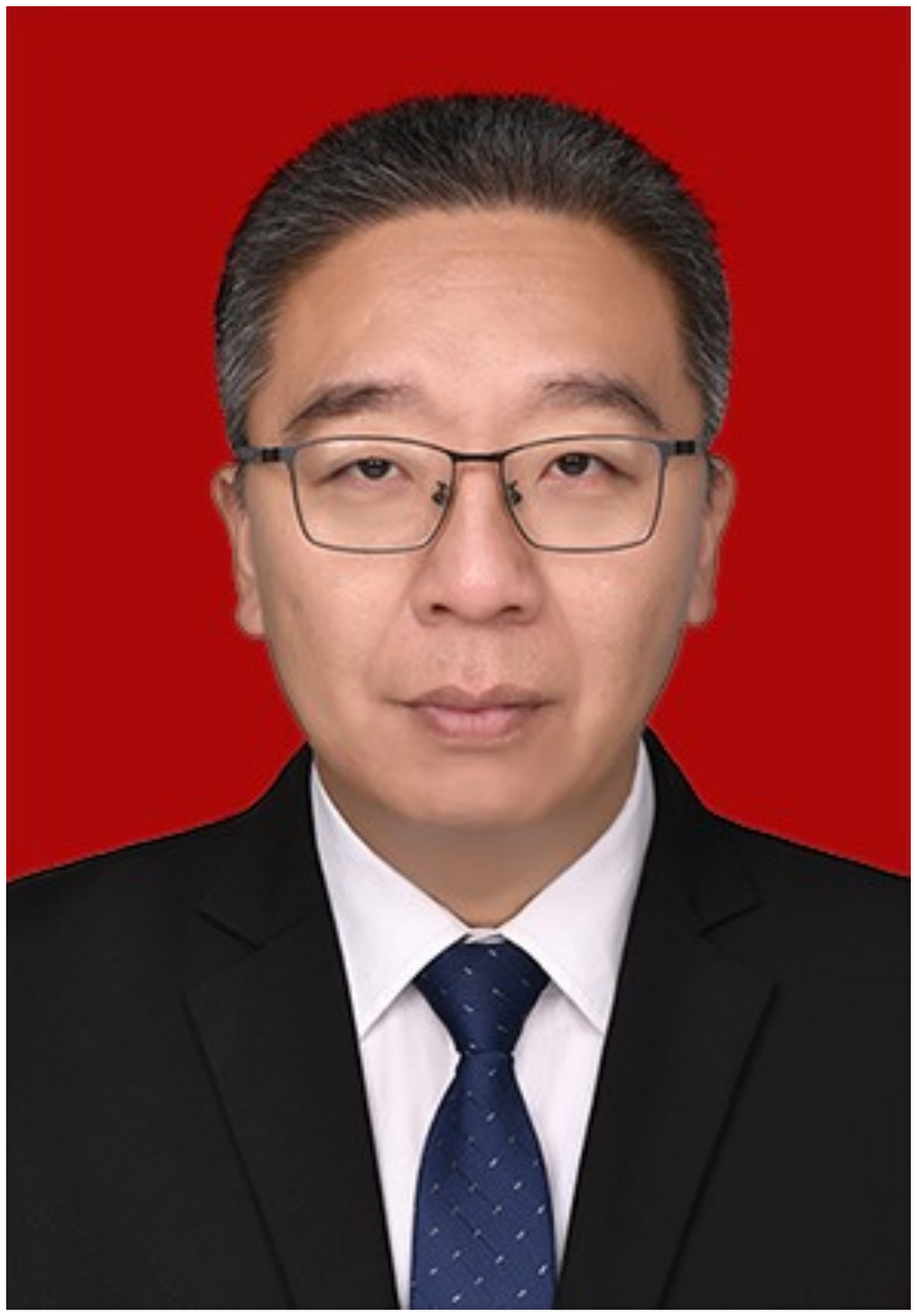}}]{Yuhua Qian}
(Member, IEEE) received the MS and PhD degrees in computers with applications from Shanxi University, Taiyuan, China, in 2005 and 2011, respectively. He is currently a director with the Institute of Big Data and Industry, Shanxi University, where he is also a professor with the Key Laboratory of Computational Intelligence and Chinese Information Processing, Ministry of Education. His research interests include artificial intelligence, data mining, machine learning, granular computing and machine vision. He has authored more than 100 articles on these topics in international journals.
\end{IEEEbiography}

\vspace{-2em}
\begin{IEEEbiography}[{\includegraphics[width=1in,height=1.25in,clip,keepaspectratio]{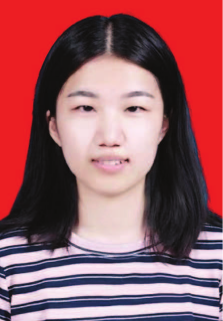}}]{Jue Li}
received the B.S. and M.S. degree in computer technology from Shanxi University, Taiyuan, China, in 2018 and 2021. She is currently pursuing the Ph.D degree with the School of Computer and Information
Technology, and working at the Institute of Big Data Science and Industry, Shanxi University. Her research interest includes machine learning and data mining.
\end{IEEEbiography}

\vspace{-2em}
\begin{IEEEbiography}[{\includegraphics[width=1in,height=1.25in,clip,keepaspectratio]{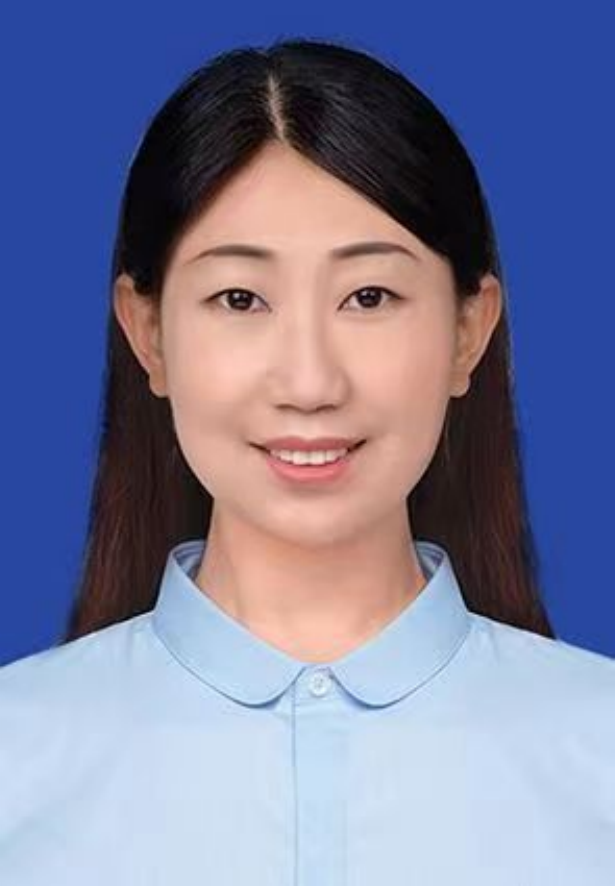}}]{Honghong Cheng}
(Member, IEEE) received the BS degree from the School of Mathematical Sciences, Shanxi University, Taiyuan, China, in 2012 and the PhD degree from the Institute of Big Data Science and Industry, Shanxi University, Taiyuan, China, in 2020. She is currently a teacher with the School of Information, Shanxi University of Finance and Economics. She was a visiting scholar with the City University of Hong Kong, Hong Kong, China, in 2019. Her research interests include associations mining, multimodal learning, data mining
\end{IEEEbiography}

\vspace{-2em}
\begin{IEEEbiography}[{\includegraphics[width=1in,height=1.25in,clip,keepaspectratio]{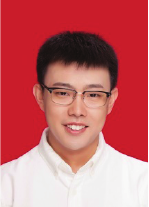}}]{Feijiang Li}
received the PhD degree in computers with applications from Shanxi University, Taiyuan, China, in 2020. He is currently an associate professor at the Institute of Big Data Science and Industry, Shanxi University. His research interest includes machine learning and knowledge discovery. In his research fields, he has published over 20 papers on international journals, including the Artificial Intelligence Journal, IEEE Transactions on Pattern Analysis and Machine Intelligence, IEEE Transactions on Neural Networks and Learning Systems, Machine Learning, and ACM Transactions on Knowledge Discovery from Data. 
\end{IEEEbiography}
\vfill

\end{document}